\documentclass[11pt]{article}

\usepackage{etoolbox}
\PassOptionsToPackage{final,eandd}{ext/neurips_2026}  %
\usepackage[venue=neurips, year=2026]{ext/mlconf}

\providetoggle{arxiv}
\providetoggle{neurips}
\providetoggle{icml}
\providetoggle{iclr}
\settoggle{neurips}{true}
\settoggle{arxiv}{false}
\settoggle{icml}{false}
\settoggle{iclr}{false}

\providetoggle{blind}
\settoggle{blind}{false}
\providetoggle{workingversion}
\settoggle{workingversion}{false}
\settoggle{showtodos}{false}
\settoggle{splitsections}{false}

\newcommand{\ourtitle}{Measuring What Matters:\\Synthetic Benchmarks for Concept Bottleneck Models}
\newcommand{\ourtitleshort}{Measuring What Matters: Synthetic Benchmarks for Concept Bottleneck Models}
\newcommand{\ourshorttitle}{\ourtitleshort}

\iftoggle{blind}{
\newcommand{\repourl}[0]{\href{https://anonymous.4open.science/r/concept-benchmark-84D2}{our anonymized repository}}
}{
\newcommand{\repourl}[0]{\href{https://github.com/ustunb/concept-benchmark}{on GitHub}}
}

\iftoggle{splitsections}{
\newcommand{\newsection}[1]{\clearpage\section{#1}}
}{
\newcommand{\newsection}[1]{\section{#1}}
}

\mlauthor{Julian Skirzy\'nski}{}{University of California, San Diego}{jskirzynski@ucsd.edu}
\mlauthor{Harry Cheon}{}{Harvard University}{hcheon@g.harvard.edu}
\mlauthor{Shreyas Kadekodi}{}{University of California, San Diego}{skadekodi@ucsd.edu}
\mlauthor{Meredith Stewart}{}{University of California, San Diego}{mstewart@ucsd.edu}
\mlauthor{Berk Ustun}{}{University of California, San Diego}{berk@ucsd.edu}

\title{\ourtitle{}}
\newcommand{\authorinfo}[4]{{#1}\\{\small #3}\\{\small\texttt{#4}}}
\author{
\authorinfo{Julian Skirzy\'nski}{CSE}{UCSD}{jskirzynski@ucsd.edu}%
\And
\authorinfo{Harry Cheon}{}{UCSD}{hcheon@ucsd.edu}%
\And
\authorinfo{Shreyas Kadekodi}{CSE}{UCSD}{skadekodi@ucsd.edu}%
\And
\authorinfo{Meredith Stewart}{CSE}{UCSD}{mstewart@ucsd.edu}%
\And
\authorinfo{Berk Ustun}{HDSI}{UCSD}{berk@ucsd.edu}
}

\usepackage[utf8]{inputenc} 
\usepackage[T1]{fontenc} 
\usepackage{environ,comment}
\usepackage{amsfonts,bbm,bm,courier,nicefrac,microtype}
\usepackage{amsthm,amssymb}
\usepackage{amsmath}
\usepackage{array,booktabs,tabularx,multirow,colortbl,hhline}
\usepackage{eqnarray}
\usepackage{enumitem}
\usepackage{xifthen}%
\usepackage{thmtools}
\usepackage{xspace}

\usepackage{graphicx,xcolor,placeins}
\usepackage{caption}
\usepackage[export]{adjustbox}%
\usepackage{wrapfig}
\usepackage{pifont}
\usepackage[most]{tcolorbox}

\definecolor{good}{HTML}{BAFFCD}
\definecolor{mid}{HTML}{FFEB9C}
\definecolor{bad}{HTML}{FFC8BA}
\tcbset{
  promptbox/.style={
    colback=gray!5,
    colframe=gray!40,
    boxrule=0.4pt,
    arc=1.5pt,
    left=6pt,
    right=6pt,
    top=6pt,
    bottom=6pt,
    width=0.9\linewidth,
    fontupper=\fontfamily{zi4}\selectfont\small,%
    before upper={\rightskip=0pt plus 3em\relax}%
  }
}

\PassOptionsToPackage{hyphens}{url}
\usepackage{hyperref}
\hypersetup{colorlinks=true, urlcolor=blue, linkcolor=blue, citecolor=blue, linkbordercolor=blue, pdfborderstyle={/S/U/W 1}}

\renewcommand{\arraystretch}{1.0}
\newcolumntype{H}{>{\setbox0=\hbox\bgroup}c<{\egroup}@{}}
\newcolumntype{R}[1]{>{\raggedright\arraybackslash}p{#1}}
\newcommand{\textheader}[1]{{\bfseries{#1}}}
\newcommand{\cell}[2]{\setlength{\tabcolsep}{0pt}\begin{tabular}{#1}#2 \end{tabular}}
\newcommand{\bfcell}[2]{\setlength{\tabcolsep}{0pt}\begin{tabular}{>\bfseries{#1}}#2 \end{tabular}}

\usepackage{natbib}
\setcitestyle{numbers,comma,square,sort}

\setitemize{leftmargin=1em,topsep=0.1em,parsep=0.5pt,}
\setlist[enumerate]{leftmargin=*, label= {\arabic*.}, itemsep=0.5em}

\usepackage[capitalise]{cleveref}
\Crefformat{figure}{Fig.~#2#1#3}
\AtBeginDocument{}
\Crefformat{definition}{Def.~#2#1#3}
\Crefformat{proposition}{Prop.~#2#1#3}
\Crefformat{theorem}{Thm.~#2#1#3}

\makeatletter
\@ifpackageloaded{algorithm}{}{\usepackage{algorithm}}
\@ifpackageloaded{algorithmic}{}{\usepackage{algpseudocode}}
\makeatother
\makeatletter\@ifpackageloaded{algpseudocode}{%

  \algnewcommand{\alginput}[2]{\Statex{Input:~#1}\Comment{#2}}
  \algnewcommand\algorithmicinput{\textbf{Input}}
  
  \algnewcommand\algorithmicinitialize{\textbf{Initialize}}
  \algnewcommand\algorithmicbigstep{\textbf{Step}}
  \algnewcommand\INPUT{\item[\algorithmicinput]}
  \algnewcommand\INITIALIZE{\item[\algorithmicinitialize]}
  \algnewcommand{\STEP}[1]{\item[\algorithmicbigstep]{\textbf{#1}}}
  \algnewcommand{\InputExplanation}[2][.6\linewidth]{\leavevmode\hfill\makebox[#1][r]{~{\footnotesize{#2}}}}
  \algnewcommand{\InitializationExplanation}[2][.6\linewidth]{\leavevmode\hfill\makebox[#1][r]{~{\footnotesize{#2}}}}
  \algnewcommand{\StateComment}[2]{\State{#1}\InputExplanation{#2}}
  \algnewcommand{\alginitialize}[2]{\Statex{#1}\InitializationExplanation{#2}}
  \algrenewcommand\algorithmiccomment[2][]{#1\hfill\textit{\scriptsize{#2}}}
  
}{%
}\makeatother

\usepackage{pgfplots}
\pgfplotsset{compat=1.17}
\usepgfplotslibrary{fillbetween}
\usetikzlibrary{patterns,decorations.pathreplacing,arrows.meta}
\definecolor{acabstain}{HTML}{2F6FB3}
\definecolor{acconfirm}{HTML}{10B981}
\definecolor{acgridline}{HTML}{DFE4E9}
\definecolor{acink}{HTML}{1F2933}
\definecolor{accheck}{HTML}{FFE699}
\definecolor{ours}{HTML}{3B6FB6}
\definecolor{annot}{HTML}{4D4D4D}
\definecolor{gridc}{HTML}{DFE4E9}
\definecolor{titlec}{HTML}{8C8C8C}
\definecolor{hair}{HTML}{B8B8B8}
\definecolor{figbad}{HTML}{E64B35}
\definecolor{ourslight}{HTML}{8FB2DB}
\definecolor{covgreen}{HTML}{4C9F70}

\newcommand{\prob}[1]{\textnormal{Pr}\left(#1\right)}
\renewcommand{\Pr}[1]{\textnormal{Pr}(#1)}
\DeclareMathOperator*{\E}{\mathbb{E}}
\newcommand{\indic}[1]{\mathbbm{I}[#1]}

\newcommand{\R}{\mathbb{R}}

\newcommand{\X}{\mathcal{X}}
\newcommand{\Y}{\mathcal{Y}}

\newcommand{\st}{\textnormal{s.t.}}
\newcommand{\optpar}[2]{\ifthenelse{\isempty{#2}}{#1}{#1({#2})}}

\newcommand{\mipwhat}[1]{\emph{\scriptsize #1}}

\newcommand{\intrange}[1]{[#1]}

\newcommand{\textds}[1]{{\small\texttt{#1}}}
\newcommand{\captionds}[1]{\texttt{#1}}

\newcommand{\textfn}[1]{{\footnotesize\texttt{#1}}}

\newcommand{\textcp}[1]{{\texttt{\footnotesize{#1}}}}
\newcommand{\captioncp}[1]{\texttt{#1}}
\newcommand{\textcpp}[1]{{\textmd{\footnotesize{#1}}}}

\newcommand{\features}{x}

\newcommand{\C}{\mathcal{C}}

\newcommand{\concepts}{c}
\newcommand{\concept}{c}

\newcommand{\outcome}{y}

\newcommand{\pipeline}{f}
\newcommand{\baseline}{f_0}
\newcommand{\detector}{g}
\newcommand{\frontend}{h}

\newcommand{\outcomepred}{\hat{\outcome}}
\newcommand{\conceptpred}{\hat{\concept}}
\newcommand{\conceptspred}{\hat{\concepts}}

\newcommand{\threshold}{\tau}
\newcommand{\abstain}{\perp}

\newcommand{\incacc}{\ensuremath{\Delta_{\textsf{Accuracy}}}}

\newcommand{\gain}{{\small\textsf{Gain}}}
\newcommand{\wrk}{{\small\textsf{WorkSaved}}}
\newcommand{\captionwrk}{\textsf{WorkSaved}}
\newcommand{\cov}{{\small\textsf{Coverage}}}
\newcommand{\captioncov}{\textsf{Coverage}}

\newcommand{\cErrPr}{\epsilon}

\newcommand{\nBoard}[0]{B}
\newcommand{\block}{\mathcal{B}}

\newcommand{\crow}[1]{c_{\textrm{row},{#1}}}
\newcommand{\ccol}[1]{c_{\textrm{col},{#1}}}
\newcommand{\cblk}[1]{c_{\textrm{blk},{#1}}}

\newcommand{\va}{\mathbf{a}}
\newcommand{\flipProb}{\text{resp}}
\newcommand{\flipThreshold}{T}

\newcommand{\CBM}{\textsf{CBM}}
\newcommand{\DNN}{\textsf{DNN}}
\newcommand{\CEM}{\textsf{CEM}}
\newcommand{\ProbCBM}{\textsf{ProbCBM}}
\newcommand{\ECBM}{\textsf{ECBM}}

\begin{document}

\doparttoc
\iftoggle{iclr}{\maketitle}{}
\iftoggle{neurips}{\maketitle}{}
\iftoggle{arxiv}{
\title{\ourtitle{}}
\author{\parbox{\textwidth}{\centering%
\textbf{Julian Skirzy\'nski, Harry Cheon, Shreyas Kadekodi, Meredith Stewart, Berk Ustun} \\[0.3em]
{\small\it University of California, San Diego} \\
{\small\sc \{jskirzynski, skadekodi, mstewart, berk\}@ucsd.edu, hcheon@g.harvard.edu}}%
}

\maketitle
}

\iftoggle{icml}{
\twocolumn[
\icmltitle{\ourtitle{}}
\icmltitlerunning{\ourtitleshort{}}
\begin{icmlauthorlist}
\icmlauthor{Julian Skirzy\'nski}{ucsd}
\icmlauthor{Harry Cheon}{ucsd}
\icmlauthor{Shreyas Kadekodi}{ucsd}
\icmlauthor{Meredith Stewart}{ucsd}
\icmlauthor{Berk Ustun}{ucsd}
\end{icmlauthorlist}
\icmlaffiliation{ucsd}{UCSD}
\icmlcorrespondingauthor{Berk Ustun}{berk@ucsd.edu}
\icmlkeywords{interpretability, explainability}
\vskip 0.42in
]
\printAffiliationsAndNotice{}
}{}

\begin{abstract}
Concept bottleneck models predict using high-level concepts detected in inputs. Although these models provide a simple way to reap benefits from interpretability, they are hard to train and evaluate because very few real-world datasets include concept labels. In this paper, we develop \emph{synthetic benchmarks} for concept bottleneck models.
With our benchmarks, researchers can generate labeled datasets and control properties that affect performance, such as data modality, the choice of concepts, or annotation quality. They can thus find problems that suit these models and isolate the factors behind their failures.
We demonstrate how our benchmarks can be used to evaluate major use cases for concept bottleneck models: \emph{decision support}, i.e., helping human experts make better decisions; and \emph{automation}, i.e., handling routine tasks without supervision. 
Our demonstrations show how the benchmarks can identify new failure modes and attribute them to either the architecture or the data.
\end{abstract}

\iftoggle{arxiv}{
\vspace{0.25em}
\begin{keywords}{Concept Bottleneck Models, Safety, Interpretability, Alignment, Benchmarks}
\end{keywords}
}

\faketableofcontents

\section{Introduction}
\label{Sec::Intro}
Many machine learning models are developed for settings in which humans require interpretability to verify, contest, and correct automated decisions. For example, many black-box models can accurately detect melanoma from dermoscopic images~\citep{esteva2017derm, kaur2022melanoma, sm2023classification, adepu2023melanoma, teja2024lightweight}, but clinicians hesitate to use them without being able to check how they reach their predictions~\citep{metta2024advancing, chanda2024dermatologist, tonekaboni2019clinicians}. A promising approach to introduce interpretability in these settings is to build models whose predictions are based on \emph{concepts}, i.e., meaningful, high-level attributes of the input. In melanoma detection, a concept-based model could first infer whether there are \textcp{dots} or \textcp{streaks} on the skin (diagnostic cues used by clinicians during screening~\citep{thomas1998semiological, ebell2008clinical}), and then use these concepts to predict the presence of melanoma. Clinicians could verify whether the detected concepts are correct, e.g., whether the image really shows streaks, and whether they influence the prediction in clinically sensible ways, e.g., whether streaks appropriately increase the probability of melanoma.

Concept bottleneck models (CBMs)~\citep{koh2020concept} are machine learning systems that implement this approach. They first predict which concepts are present in the input, and then use these concepts to predict the label. The same design now underlies typed decision models, in which a language model answers fixed questions about an input and a rule maps the answers to a decision~\citep{typesafe2026jev,firecrawl2026decisions}.

\begin{table*}[t!]
    \centering
    \resizebox{1.0\textwidth}{!}{
    \begin{tabular}{lllrrrcccc}
    & & & & & & \multicolumn{4}{c}{\bfseries Issues with Concepts} \\
    \cmidrule(lr){7-10}
    \bfcell{l}{\\Dataset} &
    \bfcell{l}{\\Task} &
    \bfcell{l}{\\Example Concepts} &
    \bfcell{c}{\\$n$} &
    \bfcell{c}{\\$m$} &
    \bfcell{l}{\\Papers} &
    \bfcell{c}{Noisy\\Annotations} &
    \bfcell{c}{Missing\\Annotations} &
    \bfcell{c}{Subjective\\Concepts} &
    \bfcell{c}{No Well-Defined\\Use Case} \\
    \midrule

    CUB-200~\citep{wah2011cub} & Classify bird species & \cell{l}{physical attributes\\(e.g., \textcp{has\_bill\_shape})} & 11,788 & 312 & 72 & $\blacksquare$ & $\blacksquare$ & $\blacksquare$ & $\blacksquare$ \\
    \midrule

    AwA2~\citep{xian2019zero} & Classify animal species & \cell{l}{physical attributes\\(e.g., \textcp{stripes})} & 37,322 & 85 & 22 & $\blacksquare$ & $\blacksquare$ &  & $\blacksquare$ \\
    \midrule

    CelebA~\citep{liu2015faceattributes} & Predict facial attributes & \cell{l}{appearance traits\\(e.g., \textcp{smiling})} & 202,599 & 40 & 21 & $\blacksquare$ & & $\blacksquare$ & $\blacksquare$ \\
    \midrule

    SkinCon~\citep{daneshjou2022skincon} & Classify skin lesion type & \cell{l}{pathologies\\(e.g., \textcp{plaque})} & 3,230 & 48 & 15 & $\blacksquare$ & & $\blacksquare$ & \\
    \midrule

    Derm7pt~\citep{kawahara2018derm} & Classify lesion as melanoma & \cell{l}{dermoscopic patterns\\(e.g., \textcp{pigment\_network})} & 2,022 & 7 & 10 & &  & $\blacksquare$ & \\
    \midrule

    CheXpert~\citep{irvin2019chexpert} & Detect chest abnormalities & \cell{l}{medical findings\\(e.g., \textcp{atelectasis})} & 224,316 & 14 & 7 & $\blacksquare$ & $\blacksquare$ & $\blacksquare$ & \\
    \midrule

    OAI~\citep{nevitt2006osteoarthritis} & Assess osteoarthritis severity & \cell{l}{joint pathology\\(e.g., \textcp{narrowing})} & 36,369 & 10 & 6 & N/A & $\blacksquare$ &  & \\
    \midrule

    PH2~\citep{mendonca2013ph2} & Classify lesion as melanoma & \cell{l}{dermoscopic patterns\\(e.g., \textcp{pigment\_network})} & 200 & 7 & 4 & &  & $\blacksquare$ & \\
    \midrule

    CEBaB~\citep{abraham2022cebab} & Predict restaurant rating & \cell{l}{sentiment in reviews\\(e.g., \textcp{food\_aspect})} & 15,089 & 4 & 4 & $\blacksquare$ & $\blacksquare$ & $\blacksquare$ & $\blacksquare$ \\
    \midrule

    DTD~\citep{cimpoi14describing} & Classify material texture & \cell{l}{texture properties\\(e.g., \textcp{blotchy})} & 5,640 & 47 & 4 & &  & $\blacksquare$ & $\blacksquare$ \\

    \bottomrule
    \end{tabular}%

    }
    \caption{Overview of datasets used to develop concept bottleneck models. We present the 10 most prevalent ones, based on 123 papers published since 2020. $n$ and $m$ are the size of the dataset and the number of concepts, respectively. Noisy Annotations marks concept labels with annotator errors or uncertainty; Missing Annotations marks concept values that are not recorded (e.g., ``not visible'' in CUB-200, ``not mentioned'' in CheXpert). $\blacksquare$ marks a dataset with the issue and N/A an issue we could not check; a dataset has no well-defined use case when correcting its concepts has no clear practical stake.}
    \label{Table::DatasetsInLiterature}
\end{table*}

The most significant barrier to adopting CBMs in practice is that their development requires datasets with concept annotations. 
There are few such datasets because developing them is costly: if a task has $m$ concepts, the final dataset will require $m$ times the annotation cost. This scarcity of datasets is a barrier to research progress on two fronts: 

\begin{enumerate}[]
\item \emph{Breadth}: Current evaluations are too narrow and may fail to reveal important failure modes. In standard machine learning, broad benchmarks reveal the limitations of a method early, while authors can still fix them or report them. For CBMs, evaluations rely on a few datasets (see~\cref{Table::DatasetsInLiterature}) whose concepts have issues that cannot be easily measured, so we do not know under which conditions the models are tested.

\item \emph{Use Cases}: Current evaluations do not always measure the practical benefit of CBMs. In many commonly used tasks (e.g., predicting attractiveness from face attributes~\citep{liu2015faceattributes}), correcting a concept does not have a clear practical purpose (e.g., improving a decision). In this regime, experiments provide limited guidance on how to improve the model.
\end{enumerate}

In this paper, we present new synthetic benchmarks for CBMs that help address these barriers. Our benchmarks target decision support and automation, and use synthetic data generation to control the concepts and their annotations without costly manual labeling. Our main contributions are as follows:

 \begin{enumerate}
    
     \item We introduce dataset generators for CBMs that produce labeled data from a known concept-to-label rule while varying modality, concept granularity, source, quality, and missingness.

     \item We formalize two use cases in which CBMs provide unique value --- \emph{decision support}, measured by gain over a black-box model, and \emph{automation}, measured by reduction in human work --- and build a benchmark for each on our generators.

     \item We demonstrate how to use our benchmarks to (1) identify limitations and determine the scope under which CBMs work; (2) test foundational assumptions about CBMs (e.g., what if we do not know the concepts); (3) evaluate how their performance changes under different conditions (e.g., automated concept labeling, alignment constraints).

 \end{enumerate}

\subsection*{Related Work}

\paragraph{\textmd{\textit{Concept Bottlenecks}}}
Our work is related to a stream of research on concept bottleneck models \citep{koh2020concept}. %
This body of work builds interpretable models that humans can correct through concept interventions~\citep[e.g.,][]{shin2023closer, mahinpei2021promises, shen2025adaptive, balloli2024they, espinosa2024learning, sheth2022learning}. %
The idea of adding bottlenecks to black-box models has inspired a diverse range of follow-up work on architectures~\citep[e.g., ][]{ismail2024concept, zintgraf025selective, jeon2025localitycbm, kim2024eqcbm, chowdhury2024adacbm, zhang2024decoupling, hu2024semisupervised, zarlenga2023tabcbm, steinmann2025object, wang2025mvp, hou2024concept} and training strategies~\citep[e.g.,][]{luyten2024theoretical, penaloza2025preference, selvaraj2024improving, yu2025language, qi2024vip, kim2023concept, sheth2023auxiliary, pang2024integrating}. Researchers have also studied properties CBMs exhibit with respect to concept specification, such as concept leakage \citep[e.g.,][]{havasi2022addressing, sun2024eliminating}, robustness \citep{hu2025stable, yan2023robust, park2025analysis, lai2024cat, raman2024concept}, or uncertainty quantification \citep{kim2023probabilistic, collins2023human, feng2024bayesian, gao2024evidential}. %

\paragraph{\textmd{\textit{Typed Decision Models}}}
Our work is also related to typed decision models, i.e., LLMs that pick one of a fixed set of options for a question and return it with a probability for each~\citep{typesafe2026jev,firecrawl2026decisions}. These models are designed to make decisions inside a software pipeline, such as whether an agent may take its next action. A developer specifies the questions and a rule that maps their answers to an action, and the model answers the questions. Such a system is a concept bottleneck model where the questions are its concepts, and the rule maps them to the label.

\paragraph{\textmd{\textit{Concept Datasets}}}
The literature evaluates CBMs on a range of concept-annotated datasets (see~\cref{Table::DatasetsInLiterature}), including \textds{CUB-200}~\citep{wah2011cub}, \textds{CelebA}~\citep{liu2015faceattributes}, and \textds{CheXpert}~\citep{irvin2019chexpert}. These datasets were not originally designed for concept supervision. For instance, it is unclear which of their concepts are relevant to the task, how noisy the concept labels are, and whether the concept set is complete. The high cost of building concept datasets motivated work in concept discovery~\citep[e.g.,][]{rao2024discover, schrodi2024concept, panousis2024coarse, xie2025discovering, zintgraf025selective}, and led many CBM papers to use datasets where concepts and labels are defined via large language models~\citep[as in, e.g.,][]{oikarinen2023label, yu2025language, yang2023language, zhang2025attribute, he2025v2c, jiang2025enhancing}. Both procedures can introduce noise into concept annotations that is difficult to quantify. As a result, the existing datasets do not provide the level of control to study CBMs under all relevant conditions, e.g., comparing how the models fare when not all relevant concepts are annotated. The evaluations may also confound shortcomings in the model architecture with imperfections in the data.

Our work seeks to complement research on CBMs with controlled synthetic benchmarks. Current synthetic datasets for concept-based models remain limited~\citep[e.g., ][]{pugnana2025deferring, shin2023closer, alukaev2023cross, barbiero2024relational, makonnen2025measuring, parisini2025leakage}, are rarely available for public use, and are often built ad hoc for specific questions. As shown in \cref{Table::Capability}, these datasets also expose few generation parameters, which limits the conditions under which CBMs can be studied. Like established ML benchmarks~\citep[see e.g.,][]{johnson2017clevr, ding2021retiring, schwettmann2023find, grinsztajn2022tree, koh2021wilds}, our benchmarks provide ground truth through a known labeling rule, reproducible results through a deterministic generator, and controlled variation through independent settings for each property of the data.

\begin{table}[t]
\centering
\small
\footnotesize
\setlength{\tabcolsep}{2.5pt}
\begin{tabular}{lcccccccc}
\bfcell{l}{\\Paper} & \bfcell{c}{\\Modality} & \bfcell{c}{Labeling\\Rule} & \bfcell{c}{Concept\\Granularity} & \bfcell{c}{Concept\\Count} & \bfcell{c}{Concept\\Source} & \bfcell{c}{Concept\\Noise} & \bfcell{c}{Concept\\Missingness} & \bfcell{c}{Intervention\\Source} \\
\midrule
Ours & \checkmark & \checkmark & \checkmark & \checkmark & \checkmark & \checkmark & \checkmark & \checkmark \\
\midrule
\citeauthor{alukaev2023cross}~(\citeyear{alukaev2023cross}) & ? & & & & ? & & & \\
\midrule[0.2pt]
\citeauthor{barbiero2024relational}~(\citeyear{barbiero2024relational}) & & & & & & ? & & \\
\midrule[0.2pt]
\citeauthor{makonnen2025measuring}~(\citeyear{makonnen2025measuring}) & & & & \checkmark & & \checkmark & & \\
\midrule[0.2pt]
\citeauthor{parisini2025leakage}~(\citeyear{parisini2025leakage}) & & ? & & \checkmark & & ? & & \\
\midrule[0.2pt]
\citeauthor{pugnana2025deferring}~(\citeyear{pugnana2025deferring}) & & & & \checkmark & & & & \checkmark \\
\midrule[0.2pt]
\citeauthor{shin2023closer}~(\citeyear{shin2023closer}) & & & & \checkmark & & \checkmark & ? & \\
\bottomrule
\end{tabular}

\caption{Overview of the data properties that our synthetic benchmarks can control, compared with the synthetic datasets used in other papers on concept-based models. Our benchmarks control many more properties, so the papers listed here could use them to drive algorithm development and stress-test new methods. \checkmark{} marks a property exposed as a control, ? a related variation that is not exposed as a control, and a blank a property that is absent.}
\label{Table::Capability}
\end{table}

\newsection{Use Cases}
\label{Sec::Preliminaries}

We start with a dataset of $n$ examples $\{(\features_i, \concepts_i, \outcome_i)\}_{i=1}^n$, where $\features_{i} \in \X \subseteq \R^d$ are features (e.g., pixels), $\concepts_{i} \in \C = \{0, 1\}^m$ are \emph{concepts} (e.g., $\concepts_{i,k} = 1$ if image $i$ shows a lesion with \textcp{streaks}), and $\outcome_{i} \in \Y = \{0, 1\}$ is a \emph{label} (e.g., patient $i$ has melanoma)\footnote{We focus on binary classification; the formalism extends to multi-class settings.}. 

We use the dataset to train:
\begin{itemize}[leftmargin =*]
    \item A set of $m$ \emph{concept detectors} $\detector_k: \X \to [0, 1]$ that predict the probability the concept is present $\Pr{\concept_{i, k} = 1 \mid \features_i}.$ 
    
    \item A \emph{front-end model} $\frontend: \C \to [0,1]$, which maps concepts $\concepts_i$ to the \emph{soft} label $\frontend(\concepts_i)$.
\end{itemize}
The full pipeline $\pipeline: \X \to \Y$ outputs the \emph{hard} label $\outcomepred_i$ by thresholding the soft label.

\paragraph{Interventions}

One practical benefit of training a concept bottleneck is that semantically meaningful concepts provide a path to improve models through human \emph{intervention}:
\begin{enumerate}
    
    \item \emph{Confirmation.} A user corrects the predicted concept. For example, if a detector predicts that concept $k = \textcp{dots}$ is present in image $i$ with unmarked skin, $\conceptspred_{i,k} = 1$, the user can correct this by setting $\conceptspred_{i,k} = 0$ and using the new value in the front-end model.
    
    \item \emph{Manipulation.} A user edits the raw input to toggle a concept. If the detector fails to detect \textcp{dots} in image $i$, the user can digitally enhance them so the detector picks them up.
    
\end{enumerate}

The value of interventions depends on the concepts and instances to which humans are asked to attend. 
In what follows, we consider a general strategy based on the responsiveness of the label to changes in concept values. We chose this approach over alternatives~\citep{shin2023closer,shen2025adaptive} because it targets the concepts most likely to change the prediction (see \cref{Appendix::InterventionPolicy}). 

An intervention induces a mapping $t: \C \times \X \to\C$ that, given input $\features$, modifies a subset of selected concepts $S(\features) \subseteq \{1,\ldots,m\}$ before they are passed to the front-end model. We bound the number of intervened concepts by a budget $b$, so that $|S(\features)| \leq b$; with the budget ``all'', we may intervene on every concept. We hold this protocol fixed across all architectures (only the propagation of corrected concepts to the label is architecture-specific).

We quantify the benefit of interventions as the improvement in expected performance with respect to a performance metric. For accuracy-based metrics, we write $\frontend(\cdot) = \outcome$ as shorthand for thresholding the soft output before checking equality. For accuracy, this improvement is
\begin{align}
\incacc(\pipeline) = \E[\frontend(t(\detector(\features), \features)) = \outcome] - \E[\frontend(\detector(\features)) = \outcome] 
\end{align}

\subsection{Decision Support}
Decision support encompasses tasks where models assist experts who make the final decision under uncertainty~\citep[see, e.g.,][]{kawahara2018derm,irvin2019chexpert}. Although black-box models can outperform experts~\citep{grove2000clinical, aegisdottir2006meta}, they are difficult to use in this role because experts cannot check their predictions. CBMs, instead, are well-suited for this role: experts can verify concept predictions, delegate the downstream task to the model, or encode domain knowledge as constraints on $\frontend$.

CBMs become viable to the extent that they outperform black-box alternatives with interventions. We benchmark CBMs by their post-intervention performance relative to a baseline predictor $\baseline$:
\begin{align}
\gain(\pipeline, \baseline) =  \E[\frontend(t(\detector(\features), \features)) = \outcome] - \E[\baseline(\features) = \outcome] .
\label{Def::Gain}
\end{align}

\subsection{Automation}
\begin{wrapfigure}[16]{R}{0.42\linewidth}
    \centering
    \vspace{-0.8em}
    \resizebox{0.94\linewidth}{!}{\def\figsf{\fontfamily{phv}\selectfont}%
\def\acaxistitle{\figsf\fontsize{10}{11}\selectfont}%
\def\actick{\figsf\fontsize{8}{9}\selectfont}%
\def\annot{\figsf\fontsize{10}{11}\selectfont}%
\def\acregion{\figsf\fontsize{8}{9}\selectfont}%
\begin{tikzpicture}[x=1.03pt,y=5.1500pt,>={Latex[length=4.2pt,width=3.6pt]}]%
\fill[accheck,fill opacity=0.55] (20,0) rectangle (40,20);%
\draw[acgridline,line width=0.6pt] (0,5) -- (100,5) (0,10) -- (100,10) (0,15) -- (100,15);%
\draw[acgridline,line width=0.6pt] (60,0) -- (60,20) (80,0) -- (80,20);%
\draw[acink,line width=0.8pt,dash pattern=on 2.2pt off 1.6pt] (0,19) -- (100,19);%
\draw[acconfirm,line width=1.2pt,dash pattern=on 2.8pt off 1.2pt,<->] (0.6,8) -- (19.4,8);%
\draw[acconfirm,line width=1.2pt,dash pattern=on 2.8pt off 1.2pt,<->] (20.6,8) -- (39.4,8);%
\draw[acconfirm,line width=1.2pt,dash pattern=on 2.8pt off 1.2pt,<->] (40.6,8) -- (99.4,8);%
\node[font=\acregion,acconfirm,anchor=base,inner sep=0pt] at (10,8.8) {auto};%
\node[font=\acregion,acconfirm,anchor=base,inner sep=0pt] at (30,8.8) {ask};%
\node[font=\acregion,acconfirm,anchor=base,inner sep=0pt] at (58,8.8) {human};%
\draw[acabstain,line width=1.2pt,dash pattern=on 2.8pt off 1.2pt,<->] (0.6,5) -- (19.4,5);%
\draw[acabstain,line width=1.2pt,dash pattern=on 2.8pt off 1.2pt,<->] (20.6,5) -- (99.4,5);%
\node[font=\acregion,acabstain,anchor=base,inner sep=0pt] at (10,5.8) {auto};%
\node[font=\acregion,acabstain,anchor=base,inner sep=0pt] at (58,5.8) {human};%
\begin{scope}%
\clip (0,0) rectangle (100,20);%
\draw[acabstain,line width=1.4pt] plot[smooth] coordinates {(0,20.000) (6,19.868) (12,19.577) (18,19.162) (20,19.000) (28,18.239) (36,17.311) (45,16.086) (55,14.514) (65,12.734) (75,10.756) (85,8.589) (95,6.240) (100,5.000)};%
\draw[acconfirm,line width=1.4pt] plot[smooth] coordinates {(0,20.000) (9,19.976) (18,19.866) (28,19.592) (40,19.000) (50,18.248) (60,17.230) (70,15.919) (80,14.292) (90,12.326) (100,10.000)};%
\end{scope}%
\draw[acink,line width=0.8pt] (0,0) rectangle (100,20);%
\fill[acabstain] (20,19) circle (3.1pt);%
\fill[acconfirm] (40,19) circle (3.1pt);%
\node[font=\annot,acabstain,anchor=west,inner sep=0pt] at (103,5) {CBM};%
\node[font=\annot,acconfirm,anchor=west,align=left,inner sep=0pt] at (103,10.4) {CBM +\\intervention};%
\node[font=\annot,acink,anchor=mid west,inner sep=0pt] at ([yshift=1pt] 103,19) {$\alpha_{\textrm{target}}$};%
\draw[acink,line width=0.6pt] (20,0) -- ([yshift=2pt] 20,0);%
\draw[acink,line width=0.6pt] (40,0) -- ([yshift=2pt] 40,0);%
\draw[acink,line width=0.6pt] (60,0) -- ([yshift=2pt] 60,0);%
\draw[acink,line width=0.6pt] (80,0) -- ([yshift=2pt] 80,0);%
\draw[acink,line width=0.6pt] (0,5) -- ([xshift=2pt] 0,5);%
\draw[acink,line width=0.6pt] (0,10) -- ([xshift=2pt] 0,10);%
\draw[acink,line width=0.6pt] (0,15) -- ([xshift=2pt] 0,15);%
\node[font=\actick,acink,anchor=north west,inner sep=0pt] at ([yshift=-4pt,xshift=-2pt] 0,0) {0\%};%
\node[font=\actick,acink,anchor=north,inner sep=0pt] at ([yshift=-4pt] 20,0) {20\%};%
\node[font=\actick,acink,anchor=north,inner sep=0pt] at ([yshift=-4pt] 40,0) {40\%};%
\node[font=\actick,acink,anchor=north,inner sep=0pt] at ([yshift=-4pt] 60,0) {60\%};%
\node[font=\actick,acink,anchor=north,inner sep=0pt] at ([yshift=-4pt] 80,0) {80\%};%
\node[font=\actick,acink,anchor=north,inner sep=0pt] at ([yshift=-4pt] 100,0) {100\%};%
\node[font=\actick,acink,anchor=east,inner sep=0pt] at ([xshift=-3pt] 0,0) {80\%};%
\node[font=\actick,acink,anchor=east,inner sep=0pt] at ([xshift=-3pt] 0,5) {85\%};%
\node[font=\actick,acink,anchor=east,inner sep=0pt] at ([xshift=-3pt] 0,10) {90\%};%
\node[font=\actick,acink,anchor=east,inner sep=0pt] at ([xshift=-3pt] 0,15) {95\%};%
\node[font=\actick,acink,anchor=east,inner sep=0pt] at ([xshift=-3pt] 0,20) {100\%};%
\node[font=\acaxistitle,acink,anchor=north,inner sep=0pt] at ([yshift=-15pt] 50,0) {Coverage};%
\node[font=\acaxistitle,acink,rotate=90,anchor=south,inner sep=0pt] at ([xshift=-26.8pt] 0,10) {Selective Accuracy};%
\end{tikzpicture}%
}
    \caption{Interventions increase the work a model automates at a fixed selective-accuracy target $\alpha_{\textrm{target}}$. A \CBM{} that abstains (\textcolor{acabstain}{blue}) automates 20\% of instances at the target and defers the rest to a human. When humans verify concepts where the model is uncertain (\colorbox{accheck}{shaded}), coverage grows to 40\% (\textcolor{acconfirm}{green}).}
    \label{Fig::AccuracyCoverage}
\end{wrapfigure}
One major use of concept-based models is automation~\citep{joren2024classification}. For example, government agencies process billions of non-digital documents annually~\citep{google2022documentai} (e.g., taxes, immigration, benefits) and use models to check if these documents contain responses for all mandatory sections~\citep[][]{pandey2023ai, di2024streamlining, gruzauskas2020robotic, shende2024enhancing}. These tasks require models to handle routine work and defer uncertain cases to humans.

Standard models for these tasks, called selective classifiers, automate the ``easy'' cases and defer uncertain instances to human experts~\citep[see][]{el2010foundations, geifman2017selective, gangrade2021selective}. These models abstain on uncertain predictions according to threshold $\threshold \in [0, 0.5]$, i.e., when $\frontend(\detector(\features)) \in [\threshold, 1-\threshold]$, the model outputs $\abstain$ (abstain) rather than a label. Their performance is evaluated on the subset of instances they choose to predict using \emph{selective accuracy}:
\begin{align}
    \textsf{SelectiveAccuracy}(\pipeline) = \E[\pipeline(\features) = \outcome \mid \pipeline(\features) \neq \abstain]
\end{align}
The proportion of instances where they predict is called \emph{coverage}.
\begin{align}
    \cov(\pipeline) = \E[\pipeline(\features) \neq \abstain]
\end{align}
Resolving abstentions is costly because humans must review each abstained instance in full. Concept-based models reduce that cost by asking humans to verify selected concepts~\citep{joren2024classification}. To ensure safe automation, we require models to meet a fixed selective-accuracy target $\alpha_{\textrm{target}}$ (see \cref{Fig::AccuracyCoverage} and \citep{geifman2017selective}).
We quantify automation using a net utility metric, which captures the fraction of work handled automatically after subtracting the cost of interventions, i.e., the average fraction of concepts per instance that a human intervenes on:
\begin{align}
    \wrk(\pipeline) = \underbrace{\E[\pipeline(\features) \neq \abstain]\vphantom{\E\!\left[\tfrac{|S(\features)|}{m}\right]}}_{\cov} - \underbrace{\E\!\left[\tfrac{|S(\features)|}{m}\right]}_{\textsf{Cost}}
\end{align}
High $\wrk$ means that a model automates most of the work after interventions; low $\wrk$ means that interventions fail to resolve abstentions.
We adopt an equal cost per intervention as a neutral default, so $\wrk{}$ reads as the fraction of the $m$ concepts a human does not need to verify. The metric also accepts other costs per intervention, e.g., when some concepts take longer to verify than others (see \cref{Appendix::AutomationSensitivity}).

\newsection{Decision Support}
\begin{figure}[t]
    \centering
    \begin{minipage}[t]{0.445\linewidth}
        \vspace{0pt}
        \centering
        \resizebox{\linewidth}{!}{\input{figures/tikz/robot_task}}
        \captionof{figure}{Instances from our synthetic classification task for decision support. We show three robots of each type, told apart by body parts (\cref{Fig::RobotFeatures}). We generate each robot programmatically and can control how its parts determine its type through a ground-truth function, controlling attainable accuracy and which concepts matter.}
        \label{Fig::RobotTask}
    \end{minipage}%
    \hfill
    \begin{minipage}[t]{0.535\linewidth}
        \vspace{0pt}
        \centering
        \resizebox{\linewidth}{!}{\input{figures/tikz/robot_features}}\par\vspace{4.4pt}
        \captionof{figure}{Concepts in the \captionds{robots} dataset. Each concept is a binary body part (e.g., \textcp{HeadShape} is square or round), and the ground-truth function decides which of them matter. \textcp{FootShape} has ten visible subtypes (right), five flat and five pointy, that can be labeled as separate concepts, e.g., \textcp{FootShapeIsRounded}.}
        \label{Fig::RobotFeatures}
    \end{minipage}
\end{figure}
\label{Sec::Robot}

Our decision-support benchmark asks models to classify fictional robots as Glorps and Drents (see \cref{Fig::RobotTask}). This task fits decision support because an expert can check every concept by looking at the robot, but cannot easily tell its type without the model. Our task is inspired by \citet{williams2010role}, who used these robots to study how explanations can help humans discover new concepts~\citep[see][for other variations]{skirzynski2024overreliance, skirzynski2025discrimination}. 
Each instance is an image of a robot with pixels $\features$, a label $\outcome \in \{0, 1\}$ for its type (Glorp or Drent), and concepts $\concepts$ for the shape or presence of its body parts:
\begin{align*}
\concept_k := \indic{\textcp{BodyPart}_k = \textcp{Value}}.
\end{align*}
As shown in \cref{Fig::RobotFeatures}, \textcp{FootShape} has ten subtypes. We assume that decision-makers could identify these variations as distinct concepts and use (sub)concepts of the form $\textcp{FootShape\_is\_Subtype}\in\{0,1\}$.

\paragraph{Dataset Generation}

We generate datasets from a finite set of 7{,}680 robots, each defined by 9 binary body parts (see \cref{Fig::RobotFeatures}), where \textcp{FootShape} comes in 10 subtypes (5 \textcp{Pointy} and 5 \textcp{Flat}) and \textcp{HandShape} in 6 (3 \textcp{Pointy} and 3 \textcp{Round}), each rendered in one of 71 color schemes. Because every robot is a known combination of concepts $\concepts$, we can control key elements of the data distribution that we would not be able to control otherwise:
\begin{itemize}[leftmargin=*]
\item \emph{Labels}: We can control how each robot maps to a Glorp or a Drent by setting $\prob{\outcome = 1 \mid \concepts}$, e.g., as a logistic function (see~\cref{Eq::RobotGT}). This sets which concepts are relevant and how they are encoded (e.g., \textcp{FootShape} as one concept or as ten subtypes, \cref{Fig::RobotFeatures}).

\item \emph{Prevalence}: We can control how often each robot appears by setting $\prob{\concepts}$, e.g., to balance classes, to make a concept rare, or to hold out a \textcp{FootShape} subtype from training.

\item \emph{Data Modality}: We can generate the \textds{robots} dataset as images, text, or tables. In images, we can control how visible each concept is through resolution: some features (e.g., \textcp{MouthType}) are machine-detectable but invisible to humans; others (e.g., \textcp{HasKnees}) become imperceptible to both. For text, we create descriptions at 3 complexity levels (simple templates, semi-structured, LLM-generated; see \cref{Appendix::RobotTextModality}). In tables, every concept is given by its true value, which removes errors in concept detection.
\end{itemize}
The model sees only the images and concept labels, so we can evaluate it under realistic conditions in which the true rule and the relevant concepts are hidden from it.

\newsection{Automation}
\label{Sec::Sudoku}
Our automation benchmark asks models to validate Sudoku boards (see \cref{Fig::Safeguard}). This task fits automation because the work a model saves is easy to count. A human checks a board in a fixed number of steps, and the model saves every step it takes over.

\begin{figure}[t]
    \centering
    \resizebox{\linewidth}{!}{\input{figures/tikz/sudoku_partial_work}}
    \caption{Partial work in the Sudoku automation benchmark. A concept-based model predicts the 27 constraint concepts of a board with handwritten digits and candidate marks, and combines them into a label probability. Here the probability falls in the abstention band, so the model abstains and asks the human to verify its two most informative concepts, block 1 ($\cblk{1}$) and row 5 ($\crow{5}$), instead of the whole board.}
    \label{Fig::Safeguard}
\end{figure}

A valid board contains digits 1--9 exactly once in each row, column, and block. In a standard Sudoku board, a $9 \times 9$ grid is divided into $\nBoard{} \times \nBoard{} = 3 \times 3$ non-overlapping sub-grids. We have to verify $3\nBoard{}^2 = 27$ rules: $9$ for each row, column, and block. We build a concept-based model with concepts \begin{align*}
\crow{j} &:= \indic{\text{$j$-th row contains all digits exactly once}}, \\
\ccol{j} &:= \indic{\text{$j$-th column contains all digits exactly once}}, \\
\cblk{j} &:= \indic{\text{$j$-th block contains all digits exactly once}}.
\end{align*}

The label is their logical AND:
\begin{align}
    y & = \mathrm{AND}(\ccol{1}, \ldots, \ccol{\nBoard{}^2},\ldots, \crow{\nBoard{}^2},\ldots, \cblk{\nBoard{}^2}) \notag \\
      &= \prod_{\substack{j=1,\ldots,\nBoard{}^2}} \crow{j}\ccol{j} \cblk{j}
\label{Eq::SudokuDomain}
\end{align}

\paragraph{Dataset Generation}

We generate datasets from boards whose validity we know by construction. We solve a mixed-integer program for a valid board (see~\cref{Appendix::Automation}) and derive further valid boards through symmetries, such as permuting digit labels (e.g., swapping all 1s and 2s) or reordering rows and columns within blocks; invalid boards come from corrupting a valid board, e.g., by swapping two numbers within a column. We then render each board as an image with at least 17 \emph{starter} values typeset in black~\citep{{mcguire2013cluesudoku}}, the solution handwritten in blue, and some cells carrying candidates, small digits in the corner that a person might have considered while solving the puzzle.
As in~\cref{Sec::Robot}, we can control elements of the data distribution that we could not control with a real dataset:

\begin{itemize}[leftmargin=*]

\item \emph{Labels}: We can flip a share of the training labels to mimic human error, without changing the rule that maps concepts to the label (see~\cref{Appendix::SudokuSampleBenchmarks}).

\item \emph{Detectability}: We can control how hard the concepts are to detect through the resolution of the image, the share of cells with candidates, the width of the grid lines, and the handwriting, without changing the boards or the rule.

\item \emph{Task Complexity}: 
We can control the difficulty of the task through the size of the board: an $\nBoard{}^2 \times \nBoard{}^2$ board has $\nBoard{}^4$ digits and $m = 3\nBoard{}^2$ concepts. On a valid board, a single concept error flips the label, so with independent detection errors at rate $\cErrPr$, the probability of mislabeling a valid board, $1 - (1-\cErrPr)^m$, grows with $m$. %

\item \emph{Out-of-Sample Examples}: 
We can generate atypical boards with missing entries or incorrect symbols (e.g., ``0'' or ``a'') to measure performance outside the training distribution.

\end{itemize}

\newsection{Demonstrations}
\label{Sec::Experiments}

In this section, we use our benchmarks to find and explain failure modes of CBMs that are hard to isolate on real data (details in~\cref{Appendix::Models}). The code to reproduce our results is available \repourl{}.

\paragraph{Setup}
\label{subsec::setup}

We generate data from two benchmarks:
\begin{enumerate}

    \item \emph{Decision Support}: We instantiate the \textds{robots} benchmark under a balanced rule (see \cref{Appendix::BalancedRule}; we study a rare-class rule in \cref{Appendix::RareClass}) that spreads weight over several features so that no single concept determines the label. To mimic a realistic annotation process, we train on an imbalanced sample that covers only six of the ten \textcp{FootShape} subtypes and test on all ten. We train models on five concept sets that differ in how the concepts are chosen and labeled: \textds{oracle}, the concepts of the labeling rule; \textds{discovered}, the concepts that annotators would list from a finite sample, which replace \textcp{FootShape} with the six subtypes seen in training and thus risk \emph{misspecification} (see \cref{Fig::RobotFeatures}); \textds{machine\_labeled}, the \textds{discovered} concepts labeled by CLIP~\citep{radford2021learning}; \textds{llm\_written}, concepts chosen by an LLM; and \textds{clip\_words}, single words ranked by CLIP. To intervene, we use three sources that differ in accuracy: the ground-truth concept labels of each dataset (\textds{perfect}), a simulated expert who is $80\%$ accurate on every concept (\textds{expert}), and an LLM that corrects each concept using the image (\textds{llm}; see~\cref{Appendix::AutomationDiagnostics}).

    \item \emph{Automation}: We generate 1{,}000 balanced $9\times 9$ Sudoku boards (50\% valid) with typeset starting values, handwritten entries in a different handwriting per board, and candidate marks. We render each board at two values of \textfn{pixels\_per\_cell}: $50$, where the handwritten entries are legible, and $10$, where they are hard to read. We calibrate the label probabilities of every selective classifier on the validation split~\citep{platt1999probabilistic, joren2024classification}.

\end{enumerate}

We test four established concept-based architectures, each designed to address a different failure mode of CBMs, and find when they fail and why. Our baseline concept-based model is the standard \CBM{}~\citep{koh2020concept}. We compare it with \CEM{}~\citep{zarlenga2022concept}, which replaces binary concept outputs with learned per-concept embeddings, \ProbCBM{}~\citep{kim2023probabilistic}, which models concepts as Gaussian latent variables, and \ECBM{}~\citep{xu2024energy}, which trains concept detection and label prediction jointly. For automation, we adapt all four architectures as selective classifiers~\citep{joren2024classification}. As a black-box baseline, we train a deep neural network (\DNN{}) end-to-end. We report all of our results as mean $\pm$ SE over $10$ seeds in \cref{Table::BigTable} (decision support) and in \cref{Table::sudoku_px} (automation), and mark in bold the highest mean in each block of a table. We call a difference significant when a paired $t$-test across seeds gives $p < 0.05$.

\begin{table*}[t]
\centering
\small
\resizebox{1.0\textwidth}{!}{\renewcommand{\arraystretch}{1.1}
\begin{tabular}{@{}ll r rrr rrr rrr@{}}
& & \multicolumn{10}{c}{\textbf{Interventions}} \\
\cmidrule(lr){3-12}
& & \multicolumn{1}{c}{none} & \multicolumn{3}{c}{$\leq 1$ concept} & \multicolumn{3}{c}{$\leq 3$ concepts} & \multicolumn{3}{c}{all concepts} \\
\cmidrule(lr){3-3} \cmidrule(lr){4-6} \cmidrule(lr){7-9} \cmidrule(lr){10-12}
\textbf{Dataset} & \textbf{Model} & & \textds{perfect} & \textds{expert} & \textds{llm} & \textds{perfect} & \textds{expert} & \textds{llm} & \textds{perfect} & \textds{expert} & \textds{llm} \\
\midrule
\multirow{4}{*}{\cell{l}{\textds{oracle}\\$m{=}7$}}
 & \CBM{}     & 84.5\%{\scriptsize$\pm$0.3} & 91.8\%{\scriptsize$\pm$0.5} & \textbf{88.6\%}{\scriptsize$\boldsymbol{\pm}$\textbf{0.4}} & 83.9\%{\scriptsize$\pm$0.5} & 91.8\%{\scriptsize$\pm$0.5} & 86.6\%{\scriptsize$\pm$0.5} & 83.5\%{\scriptsize$\pm$0.5} & \textbf{92.0\%}{\scriptsize$\boldsymbol{\pm}$\textbf{0.5}} & 86.0\%{\scriptsize$\pm$0.3} & 83.6\%{\scriptsize$\pm$0.3} \\
 & \CEM{}     & 84.2\%{\scriptsize$\pm$1.3} & 87.8\%{\scriptsize$\pm$1.1} & 85.4\%{\scriptsize$\pm$0.9} & 82.3\%{\scriptsize$\pm$0.9} & 89.3\%{\scriptsize$\pm$0.8} & 84.5\%{\scriptsize$\pm$0.5} & 81.3\%{\scriptsize$\pm$0.6} & 89.2\%{\scriptsize$\pm$0.7} & 84.6\%{\scriptsize$\pm$0.6} & 82.1\%{\scriptsize$\pm$0.6} \\
 & \ProbCBM{} & 86.8\%{\scriptsize$\pm$1.6} & 87.2\%{\scriptsize$\pm$1.3} & 85.4\%{\scriptsize$\pm$1.2} & 82.9\%{\scriptsize$\pm$1.2} & 88.2\%{\scriptsize$\pm$1.1} & 82.9\%{\scriptsize$\pm$0.6} & 80.2\%{\scriptsize$\pm$0.4} & 90.2\%{\scriptsize$\pm$1.1} & 84.3\%{\scriptsize$\pm$0.6} & 82.3\%{\scriptsize$\pm$0.5} \\
 & \ECBM{}    & \textbf{87.3\%}{\scriptsize$\boldsymbol{\pm}$\textbf{1.2}} & 85.0\%{\scriptsize$\pm$1.4} & 84.4\%{\scriptsize$\pm$1.5} & 84.0\%{\scriptsize$\pm$1.5} & 85.4\%{\scriptsize$\pm$1.3} & 81.4\%{\scriptsize$\pm$1.5} & 81.1\%{\scriptsize$\pm$1.5} & 91.0\%{\scriptsize$\pm$0.5} & 85.4\%{\scriptsize$\pm$1.0} & 84.7\%{\scriptsize$\pm$0.9} \\
\midrule
\multirow{4}{*}{\cell{l}{\textds{discovered}\\$m{=}12$}}
 & \CBM{}     & 77.6\%{\scriptsize$\pm$0.8} & 83.5\%{\scriptsize$\pm$0.9} & 81.8\%{\scriptsize$\pm$0.8} & 80.5\%{\scriptsize$\pm$0.7} & 84.5\%{\scriptsize$\pm$0.6} & 81.8\%{\scriptsize$\pm$0.6} & 80.7\%{\scriptsize$\pm$0.6} & 85.7\%{\scriptsize$\pm$0.4} & 80.5\%{\scriptsize$\pm$0.6} & 80.3\%{\scriptsize$\pm$0.6} \\
 & \CEM{}     & 85.8\%{\scriptsize$\pm$0.7} & 88.0\%{\scriptsize$\pm$0.8} & \textbf{86.1\%}{\scriptsize$\boldsymbol{\pm}$\textbf{0.8}} & 83.2\%{\scriptsize$\pm$0.8} & \textbf{88.1\%}{\scriptsize$\boldsymbol{\pm}$\textbf{0.7}} & 84.6\%{\scriptsize$\pm$0.7} & 82.9\%{\scriptsize$\pm$0.8} & 87.6\%{\scriptsize$\pm$0.6} & 84.5\%{\scriptsize$\pm$0.6} & 84.8\%{\scriptsize$\pm$0.5} \\
 & \ProbCBM{} & \textbf{86.1\%}{\scriptsize$\boldsymbol{\pm}$\textbf{2.1}} & 84.7\%{\scriptsize$\pm$2.0} & 84.3\%{\scriptsize$\pm$2.0} & 83.8\%{\scriptsize$\pm$1.7} & 83.3\%{\scriptsize$\pm$1.3} & 81.5\%{\scriptsize$\pm$1.0} & 81.3\%{\scriptsize$\pm$0.6} & 86.1\%{\scriptsize$\pm$0.6} & 83.2\%{\scriptsize$\pm$0.7} & 82.4\%{\scriptsize$\pm$0.5} \\
 & \ECBM{}    & 74.0\%{\scriptsize$\pm$3.2} & 72.5\%{\scriptsize$\pm$3.2} & 71.8\%{\scriptsize$\pm$3.2} & 72.0\%{\scriptsize$\pm$3.3} & 74.5\%{\scriptsize$\pm$2.0} & 68.5\%{\scriptsize$\pm$2.1} & 69.1\%{\scriptsize$\pm$2.6} & 84.3\%{\scriptsize$\pm$0.8} & 74.2\%{\scriptsize$\pm$1.4} & 75.7\%{\scriptsize$\pm$1.3} \\
\midrule
\multirow{4}{*}{\cell{l}{\textds{machine\_}\\\textds{labeled}\\$m{=}12$}}
 & \CBM{}     & 71.9\%{\scriptsize$\pm$0.1} & 60.8\%{\scriptsize$\pm$0.7} & 57.6\%{\scriptsize$\pm$0.4} & 59.1\%{\scriptsize$\pm$0.5} & 57.3\%{\scriptsize$\pm$0.9} & 53.2\%{\scriptsize$\pm$0.7} & 56.2\%{\scriptsize$\pm$0.8} & 53.6\%{\scriptsize$\pm$0.3} & 52.3\%{\scriptsize$\pm$0.2} & 53.3\%{\scriptsize$\pm$0.3} \\
 & \CEM{}     & \textbf{84.8\%}{\scriptsize$\boldsymbol{\pm}$\textbf{0.6}} & 84.8\%{\scriptsize$\pm$0.6} & 84.7\%{\scriptsize$\pm$0.6} & 84.8\%{\scriptsize$\pm$0.6} & 84.7\%{\scriptsize$\pm$0.6} & 84.6\%{\scriptsize$\pm$0.6} & 84.7\%{\scriptsize$\pm$0.6} & 84.8\%{\scriptsize$\pm$0.6} & 84.8\%{\scriptsize$\pm$0.6} & \textbf{84.9\%}{\scriptsize$\boldsymbol{\pm}$\textbf{0.6}} \\
 & \ProbCBM{} & 82.1\%{\scriptsize$\pm$1.4} & 81.8\%{\scriptsize$\pm$1.4} & 81.9\%{\scriptsize$\pm$1.4} & 82.0\%{\scriptsize$\pm$1.4} & 80.1\%{\scriptsize$\pm$1.3} & 80.4\%{\scriptsize$\pm$1.2} & 80.6\%{\scriptsize$\pm$1.3} & 73.8\%{\scriptsize$\pm$0.7} & 73.2\%{\scriptsize$\pm$0.6} & 73.8\%{\scriptsize$\pm$0.7} \\
 & \ECBM{}    & 83.8\%{\scriptsize$\pm$0.4} & 66.2\%{\scriptsize$\pm$0.2} & 66.2\%{\scriptsize$\pm$0.2} & 66.2\%{\scriptsize$\pm$0.2} & 63.1\%{\scriptsize$\pm$0.6} & 62.9\%{\scriptsize$\pm$0.6} & 63.1\%{\scriptsize$\pm$0.6} & 72.9\%{\scriptsize$\pm$0.7} & 73.2\%{\scriptsize$\pm$0.6} & 73.1\%{\scriptsize$\pm$0.7} \\
\midrule
\multirow{4}{*}{\cell{l}{\textds{llm\_written}\\$m{=}12$}}
 & \CBM{}     & 72.7\%{\scriptsize$\pm$0.1} & 66.6\%{\scriptsize$\pm$0.2} & 61.9\%{\scriptsize$\pm$0.2} & 52.8\%{\scriptsize$\pm$0.3} & 64.3\%{\scriptsize$\pm$0.2} & 58.2\%{\scriptsize$\pm$0.3} & 44.6\%{\scriptsize$\pm$0.1} & 64.1\%{\scriptsize$\pm$0.2} & 55.4\%{\scriptsize$\pm$0.2} & 41.1\%{\scriptsize$\pm$0.2} \\
 & \CEM{}     & \textbf{85.3\%}{\scriptsize$\boldsymbol{\pm}$\textbf{0.7}} & 85.2\%{\scriptsize$\pm$0.7} & 85.2\%{\scriptsize$\pm$0.7} & 85.1\%{\scriptsize$\pm$0.7} & 85.1\%{\scriptsize$\pm$0.7} & 85.0\%{\scriptsize$\pm$0.7} & 84.9\%{\scriptsize$\pm$0.7} & 85.2\%{\scriptsize$\pm$0.7} & 85.2\%{\scriptsize$\pm$0.7} & 85.2\%{\scriptsize$\pm$0.7} \\
 & \ProbCBM{} & 83.0\%{\scriptsize$\pm$1.7} & 83.1\%{\scriptsize$\pm$1.4} & 83.0\%{\scriptsize$\pm$1.4} & 83.1\%{\scriptsize$\pm$1.5} & 78.7\%{\scriptsize$\pm$0.9} & 79.2\%{\scriptsize$\pm$0.9} & 80.7\%{\scriptsize$\pm$1.2} & 68.7\%{\scriptsize$\pm$0.4} & 70.8\%{\scriptsize$\pm$0.5} & 74.6\%{\scriptsize$\pm$0.9} \\
 & \ECBM{}    & 84.5\%{\scriptsize$\pm$0.6} & 67.6\%{\scriptsize$\pm$0.2} & 67.5\%{\scriptsize$\pm$0.2} & 67.3\%{\scriptsize$\pm$0.2} & 66.9\%{\scriptsize$\pm$0.2} & 65.7\%{\scriptsize$\pm$0.3} & 64.8\%{\scriptsize$\pm$0.4} & 70.5\%{\scriptsize$\pm$0.2} & 71.7\%{\scriptsize$\pm$0.2} & 74.0\%{\scriptsize$\pm$0.5} \\
\midrule
\multirow{4}{*}{\cell{l}{\textds{clip\_words}\\$m{=}12$}}
 & \CBM{}     & 74.2\%{\scriptsize$\pm$0.1} & 71.1\%{\scriptsize$\pm$0.2} & 68.3\%{\scriptsize$\pm$0.3} & 64.8\%{\scriptsize$\pm$0.4} & 68.6\%{\scriptsize$\pm$0.2} & 65.5\%{\scriptsize$\pm$0.3} & 53.9\%{\scriptsize$\pm$0.3} & 68.4\%{\scriptsize$\pm$0.2} & 58.7\%{\scriptsize$\pm$0.3} & 49.9\%{\scriptsize$\pm$0.1} \\
 & \CEM{}     & 83.8\%{\scriptsize$\pm$0.8} & 83.7\%{\scriptsize$\pm$0.8} & 83.7\%{\scriptsize$\pm$0.8} & 83.6\%{\scriptsize$\pm$0.8} & 83.5\%{\scriptsize$\pm$0.8} & 83.6\%{\scriptsize$\pm$0.8} & 83.5\%{\scriptsize$\pm$0.8} & 83.7\%{\scriptsize$\pm$0.8} & 83.7\%{\scriptsize$\pm$0.8} & 83.7\%{\scriptsize$\pm$0.8} \\
 & \ProbCBM{} & 84.9\%{\scriptsize$\pm$1.0} & 84.6\%{\scriptsize$\pm$0.9} & 84.6\%{\scriptsize$\pm$0.9} & 84.6\%{\scriptsize$\pm$0.9} & 82.5\%{\scriptsize$\pm$1.0} & 82.7\%{\scriptsize$\pm$1.0} & 83.2\%{\scriptsize$\pm$1.1} & 71.2\%{\scriptsize$\pm$0.3} & 72.6\%{\scriptsize$\pm$0.3} & 74.7\%{\scriptsize$\pm$1.4} \\
 & \ECBM{}    & \textbf{85.4\%}{\scriptsize$\boldsymbol{\pm}$\textbf{0.6}} & 70.7\%{\scriptsize$\pm$0.2} & 70.6\%{\scriptsize$\pm$0.2} & 70.3\%{\scriptsize$\pm$0.2} & 69.8\%{\scriptsize$\pm$0.2} & 68.4\%{\scriptsize$\pm$0.2} & 67.2\%{\scriptsize$\pm$0.3} & 72.7\%{\scriptsize$\pm$0.2} & 73.8\%{\scriptsize$\pm$0.3} & 76.2\%{\scriptsize$\pm$0.4} \\
\bottomrule
\end{tabular}%
}
\caption{Accuracy of concept-based models on the decision-support benchmark from the \captionds{robots} task described in \cref{Sec::Robot} (see settings in \cref{Appendix::RobotsSetupParams}).
We report accuracy without interventions (none) and after correcting up to $1$, $3$, or all uncertain concepts per robot, for each concept set and intervention source defined in \cref{Sec::Experiments}. Most models are more accurate after interventions on \captionds{oracle} and \captionds{discovered} concepts, and can be less accurate on automated concept sets.}
\label{Table::BigTable}
\end{table*}

\paragraph{Decision Support with Discovered Concepts}
\begin{wrapfigure}{R}{0.55\linewidth}
    \centering
    \resizebox{\linewidth}{!}{\begin{tikzpicture}
\newcommand{\dsCM}[2]{\fontencoding{OT1}\fontfamily{#1}\fontsize{#2}{#2}\selectfont}
\definecolor{dsBlue}{HTML}{3B6FB6}
\definecolor{dsCBM}{HTML}{C44E52}
\definecolor{dsCEM}{HTML}{3E8E5E}
\definecolor{dsECBM}{HTML}{8E5BA8}
\definecolor{dsProb}{HTML}{D9822B}
\definecolor{dsLabel}{HTML}{4D4D4D}
\definecolor{dsGrid}{HTML}{DFE4E9}
\colorlet{dsDNN}{black!60}
\begin{axis}[
  set layers, width=5.6cm, height=4.8cm, scale only axis, clip=false,
  xmin=-0.35, xmax=3.35, ymin=68, ymax=94,
  axis x line*=bottom, axis y line*=left,
  axis line style={-, black!55, line width=0.5pt},
  every tick/.style={black!55, line width=0.4pt}, tick align=inside, tick pos=left,
  xtick={0,1,2,3}, xticklabels={none,{$\leq$\,1},{$\leq$\,3},all},
  ytick={70,75,80,85,90}, yticklabel={\pgfmathprintnumber[assume math mode=true]{\tick}\%},
  ticklabel style={font=\dsCM{phv}{8}, text=black!75},
  label style={font=\dsCM{phv}{8}, text=dsLabel}, xlabel style={yshift=2pt}, ylabel style={yshift=-3.36pt},
  xlabel={Interventions}, ylabel={Accuracy},
  ymajorgrids, grid style={dsGrid, line width=0.4pt},
  mark size=1.8pt,
  dsLine/.style={#1, line width=1.1pt, mark options={solid, fill=#1}},
  dsBand/.style={draw=none, fill=#1!18, forget plot},
]
\addplot[draw=none, fill=black!18, forget plot] coordinates {(-0.35,87.71)(3.35,87.71)(3.35,88.29)(-0.35,88.29)} -- cycle;
\addplot[dsBand=dsBlue] coordinates {(0,84.24)(1,91.25)(2,91.36)(3,91.54)(3,92.44)(2,92.34)(1,92.27)(0,84.76)} -- cycle;
\addplot[dsBand=dsProb] coordinates {(0,83.97)(1,82.69)(2,81.99)(3,85.51)(3,86.65)(2,84.53)(1,86.69)(0,88.26)} -- cycle;
\addplot[dsBand=dsCBM]  coordinates {(0,76.81)(1,82.53)(2,83.87)(3,85.32)(3,86.05)(2,85.13)(1,84.42)(0,78.45)} -- cycle;
\addplot[dsBand=dsCEM]  coordinates {(0,85.09)(1,87.18)(2,87.31)(3,86.94)(3,88.21)(2,88.80)(1,88.84)(0,86.43)} -- cycle;
\addplot[dsBand=dsECBM] coordinates {(0,70.73)(1,69.30)(2,72.47)(3,83.52)(3,85.13)(2,76.57)(1,75.66)(0,77.20)} -- cycle;
\addplot[dsLine=dsBlue, mark=*]          coordinates {(0,84.5)(1,91.76)(2,91.85)(3,91.99)};
\addplot[dsLine=dsCBM, mark=*]           coordinates {(0,77.63)(1,83.47)(2,84.50)(3,85.69)};
\addplot[dsLine=dsCEM, mark=square*]     coordinates {(0,85.76)(1,88.01)(2,88.05)(3,87.58)};
\addplot[dsLine=dsECBM, mark=diamond*]   coordinates {(0,73.97)(1,72.48)(2,74.52)(3,84.33)};
\addplot[dsLine=dsProb, mark=triangle*]  coordinates {(0,86.12)(1,84.69)(2,83.26)(3,86.08)};
\addplot[black, line width=1.1pt, dash pattern=on 3pt off 2.5pt, forget plot, on layer=axis foreground] coordinates {(-0.35,88) (3.35,88)};
\draw[dsLabel, line width=0.6pt, <->, >=latex] (axis cs:3.2,88) -- (axis cs:3.2,91.99);
\node[anchor=east, inner sep=0pt, font=\fontsize{8}{9}\selectfont] at (axis cs:3.14,90) {\fontfamily{phv}\selectfont Gain = 4.0 $\pm$ 0.5\%};
\tikzset{dsEnd/.style={anchor=west, inner sep=0pt, font=\fontsize{8}{9}\fontfamily{phv}\selectfont}}
\node[dsEnd, text=dsBlue] at ([xshift=4pt]axis cs:3.35,92.6) {CBM (oracle)};
\node[dsEnd, text=black!85] at ([xshift=4pt]axis cs:3.35,90.4) {DNN};
\node[dsEnd, text=dsCEM] at ([xshift=4pt]axis cs:3.35,88.2) {CEM};
\node[dsEnd, text=dsProb] at ([xshift=4pt]axis cs:3.35,86.0) {ProbCBM};
\node[dsEnd, text=dsCBM] at ([xshift=4pt]axis cs:3.35,83.8) {CBM};
\node[dsEnd, text=dsECBM] at ([xshift=4pt]axis cs:3.35,81.6) {ECBM};
\end{axis}
\end{tikzpicture}%
}
    \caption{Our benchmarks show how the choice of concepts and the architecture determine the value of interventions. With \captionds{oracle} concepts, interventions lift the \CBM{} past the \DNN{}. With \captionds{discovered} concepts, which are finer-grained and do not cover all the variability in the data, no architecture outperforms the \DNN{}, even those that use information beyond the concepts.}
    \label{Fig::MisspecifiedConcepts}
\end{wrapfigure}
\label{Sec::MisspecifiedConcepts}
We test our decision-support setup on the \textds{oracle} and \textds{discovered} concepts. As shown in \cref{Fig::MisspecifiedConcepts}, the \DNN{} reaches $88\%$ accuracy on this task, whereas the performance of the \CBM{} depends on the concept set:

\textbf{With oracle concepts, the \CBM{} outperforms the \DNN{} after interventions.} Without interventions, the \CBM{} is less accurate than the \DNN{} because its detectors mispredict some concepts, e.g., \textcp{HasKnees}, which they detect at chance. After interventions, the \CBM{} is significantly more accurate than the \DNN{}, with a final \gain{} of over $4$ percentage points.

\textbf{With discovered concepts, the \CBM{} remains less accurate than the \DNN{} after interventions.} At every intervention budget, the \CBM{} is significantly less accurate than the \DNN{}, by about $2$ points even after correcting every concept. 

This is because no intervention can correct a concept that is not in the concept set. In the test set, $40\%$ of robots have a foot subtype unseen in training. The existing foot-related concepts are all false for these robots and cannot represent whether their feet are pointy. With every concept corrected, the \CBM{} on \textds{discovered} concepts is therefore correct on $77\%$ of such robots and on $92\%$ of the rest, whereas on \textds{oracle} concepts it is correct on $91\%$ and $93\%$, respectively.

\paragraph{Failure Modes Across Architectures}
\label{Sec::Architectures}
One of the benefits of our approach is that we can test different algorithms under the same controlled conditions. Here, the comparison shows that no architecture can make up for a missing concept. \CEM{}, \ProbCBM{}, and \ECBM{} pass information from the image to their label predictors beyond the listed concepts and could in principle recover the missing foot shape. However, with perfect interventions on \textds{discovered} concepts, none of them outperforms the \DNN{} (see \cref{Fig::MisspecifiedConcepts}). Since we know which concept is missing, we can trace each failure to a different cause.

\textbf{\CEM{} encodes the missing concept but makes little use of it.} A probe trained on the embeddings of robots with unseen foot types recovers their foot shape with $99\%$ accuracy, whereas a probe trained on robots with seen foot types is between $30\%$ and $90\%$ accurate on them (see~\cref{Appendix::ArchitectureDiagnostics}). The embeddings thus carry the foot shape of unseen types, but in a form that a model fit on seen types cannot read.

\textbf{\ProbCBM{} gains little from correcting concepts that its detector cannot recognize.} An intervention replaces the concept's image-based embedding with a learned vector for ``present'' or ``absent'', which \ProbCBM{} trains to match its concept detector. For a concept the detector cannot recognize, the two vectors are nearly identical (see~\cref{Appendix::RareClass}). \ProbCBM{} thus loses the image information and gains little about the label, so it is significantly less accurate after a few interventions and stays below the \DNN{} even with every concept corrected.

\textbf{\ECBM{} loses image information by learning incomplete concepts.} \ECBM{} uses one encoding of the image to detect the concepts and, before any intervention, to predict the label. To test whether learning the concepts harms the encoding, we retrain \ECBM{} without the concept terms in its loss. Without learning the concepts, \ECBM{} is $8$ points more accurate (see \cref{Appendix::ArchitectureDiagnostics,Appendix::Leakage}). We conclude that learning incomplete concepts makes the encoding less informative about the label.

\paragraph{Decision Support When Concepts and Interventions Are Automated}
\label{Sec::Noise}

\begin{figure}[t]
    \centering
    \resizebox{\linewidth}{!}{\begin{tikzpicture}
\def\csrm{\fontsize{9pt}{10pt}\usefont{OT1}{phv}{m}{n}}%
\def\cstt{\fontsize{8.5pt}{10pt}\usefont{OT1}{cmtt}{m}{n}}%
\def\cssm{\fontsize{8pt}{9pt}\usefont{OT1}{phv}{m}{n}}%
\def\csbf{\fontsize{9pt}{10pt}\usefont{OT1}{phv}{m}{n}}%
\def\cssf{\fontsize{9.5pt}{11pt}\usefont{OT1}{phv}{m}{n}}%
\colorlet{csrule}{black!55}%
\def\csgap{0.25}%
\pgfplotstableread{
x y lab sd
0 3 7.4 0.5
1 3 2.6 0.4
2 3 -0.8 0.4
3 3 6.9 0.7
4 3 3.8 0.6
5 3 2.9 0.7
6 3 -14.7 0.6
7 3 -17.5 0.5
8 3 -15.7 0.5
9 3 -7.7 0.2
10 3 -14.2 0.2
11 3 -26.6 0.2
12 3 -4.8 0.1
13 3 -10.0 0.2
14 3 -18.0 0.3
0 2 4.6 0.5
1 2 0.6 0.6
2 2 -2.2 0.6
3 2 2.1 0.1
4 2 -0.7 0.2
5 2 -2.1 0.3
6 2 -0.0 0.0
7 2 -0.1 0.0
8 2 -0.0 0.1
9 2 -0.2 0.0
10 2 -0.2 0.0
11 2 -0.2 0.0
12 2 -0.2 0.0
13 2 -0.2 0.0
14 2 -0.2 0.1
0 1 -0.2 0.7
1 1 -3.5 0.8
2 1 -4.0 0.9
3 1 3.1 1.8
4 1 -2.5 1.5
5 1 -1.7 1.5
6 1 -16.3 0.4
7 1 -16.3 0.4
8 1 -16.3 0.4
9 1 -16.2 0.6
10 1 -16.3 0.6
11 1 -15.9 0.6
12 1 -14.3 0.5
13 1 -14.4 0.5
14 1 -14.1 0.5
0 0 1.7 1.4
1 0 -2.6 1.1
2 0 -5.1 1.2
3 0 -1.4 1.2
4 0 -3.1 1.0
5 0 -3.6 1.3
6 0 -3.5 0.5
7 0 -3.6 0.5
8 0 -3.3 0.5
9 0 -6.2 1.1
10 0 -5.3 0.9
11 0 -3.5 0.6
12 0 -5.5 0.4
13 0 -4.9 0.4
14 0 -4.1 0.4
}\csdatatable
\begin{axis}[
  name=csaxis,
  scale only axis, width=15.147cm, height=3.2cm,
  xmin=-0.5, xmax=15.5, ymin=-0.5, ymax=3.5,
  axis line style={draw=none},
  major tick style={csrule, thin},
  xtick={0,1,2,3.25,4.25,5.25,6.5,7.5,8.5,9.75,10.75,11.75,13,14,15}, ytick={0,...,3},
  xticklabel pos=upper,
  xticklabels={perfect,expert,llm,perfect,expert,llm,perfect,expert,llm,perfect,expert,llm,perfect,expert,llm},
  yticklabels={ProbCBM,ECBM,CEM,CBM},
  xticklabel style={font=\cssm, anchor=base, inner sep=0pt, yshift=5.5bp},
  yticklabel style={font=\cssf, text=black!75, anchor=east, inner sep=0pt, xshift=-3.321bp},
  colormap={csrdbu}{rgb255=(178,24,43) rgb255=(244,165,130) rgb255=(247,247,247) rgb255=(146,197,222) rgb255=(33,102,172)},
  point meta min=-30, point meta max=30,
  colorbar,
  colorbar shift/.style={xshift=0.32cm},
  colorbar style={width=0.25cm, ytick={-30,0,30}, yticklabel={\pgfmathprintnumber[assume math mode=true]{\tick}\%}, yticklabel style={font=\csrm},
    axis line style={csrule}, major tick style={csrule, thin},
    title={$\Delta_{\textsf{Accuracy}}$}, title style={font=\fontsize{10pt}{12pt}\usefont{OT1}{phv}{m}{n}, xshift=8bp, yshift=2bp}},
  clip=false,
]
\addplot[matrix plot*, mesh/cols=15, point meta=explicit, x filter/.expression={x+\csgap*floor((x+0.5)/3)}] table [meta=lab] {\csdatatable};
\foreach \xa/\xb in {2.5/2.75, 5.75/6, 9/9.25, 12.25/12.5}
  {\edef\csnode{\noexpand\fill[white] (axis cs:\xa,-0.6) rectangle (axis cs:\xb,3.6);}\csnode}
\addplot[draw=none, mark=none, point meta=explicit symbolic,
  visualization depends on={\thisrow{sd} \as \cssd},
  nodes near coords={\pgfplotspointmeta\%\\[-1.5pt]{\fontsize{7pt}{8pt}\usefont{OT1}{phv}{m}{n}\color{.!60}$\pm$\pgfmathprintnumber[fixed, fixed zerofill, precision=1, assume math mode=true]{\cssd}}},
  every node near coord/.style={anchor=center, align=center, font=\cssm, inner sep=0pt, black},
  every node near coord/.append code={\ifdim\pgfplotspointmeta pt<-20pt\pgfkeysalso{text=white}\fi},
  x filter/.expression={x+\csgap*floor((x+0.5)/3)},
] table [meta=lab] {\csdatatable};
\foreach \i/\name in {1/oracle, 4.25/discovered, 7.5/machine\_labeled, 10.75/llm\_written, 14/clip\_words}
  {\edef\csnode{\noexpand\node[font=\noexpand\cstt, anchor=base, inner sep=0pt] at ([yshift=19bp]axis cs:\i,3.5) {\unexpanded\expandafter{\name}};}\csnode}
\foreach \xa/\xb in {-0.4/2.4, 2.85/5.65, 6.1/8.9, 9.35/12.15, 12.6/15.4}
  {\edef\csnode{\noexpand\draw[csrule, line width=0.4pt] ([yshift=15bp]axis cs:\xa,3.5) -- ([yshift=15bp]axis cs:\xb,3.5);}\csnode}
\foreach \y/\txt in {5.5/Intervention, 19/Dataset}
  {\edef\csnode{\noexpand\node[font=\noexpand\csbf, anchor=base east, inner sep=0pt] at ([xshift=-4.663bp, yshift=\y bp]rel axis cs:0,1) {\txt};}\csnode}
\end{axis}
\pgfresetboundingbox
\path ([shift={(-54.67bp,-3bp)}]csaxis.south west) rectangle ([shift={(471bp,122bp)}]csaxis.south west);
\end{tikzpicture}%
}
    \caption{Our benchmarks show how automating concepts and interventions changes the value of interventions. We report mean $\incacc$ $\pm$ SE over correcting up to $1$, $3$, and all concepts per robot for each model, concept set, and intervention source. On every automated concept set, every model is less accurate after interventions, and almost always when an LLM intervenes.}
    \label{Fig::ConceptSource}
\end{figure}

Annotating concepts and intervening on them are both expensive, so practitioners often turn to automated pipelines~\citep{oikarinen2023label, oikarinen2022clip}. We test both approaches in our benchmark and show the average effect of interventions in \cref{Fig::ConceptSource}. We compare the \textds{discovered} concepts to the three automated sets, with the label-free CBM of \citet{oikarinen2023label} as the \CBM{}. For interventions, we compare the simulated expert to the LLM.

\textbf{Models trained on automated concepts are less accurate after interventions because they rely on leakage}. On every automated concept set, correcting the concepts with their ground-truth labels makes every model significantly less accurate (besides the \CEM{} on \textds{machine\_labeled}; \cref{Fig::ConceptSource}). To understand why, we intervene on each model with its own answers, i.e., its concept predictions thresholded at 0.5. Although these answers change no concept value, accuracy drops by a similar amount (see~\cref{Appendix::AutomationDiagnostics}). We conclude that the models rely on information in the detector's output beyond the concept values~\citep[i.e., on leakage, see][]{mahinpei2021promises,havasi2022addressing}, which is lost with any intervention. On \textds{oracle} and \textds{discovered} concepts, in contrast, every architecture gains more from the ground truth values than from its own answers. The CEM loses less accuracy because it needs few interventions (about 5\% of robots). We observe the same drop in accuracy on two real datasets (see~\cref{Appendix::RealData}).

\textbf{An LLM as accurate as an expert can be far less useful as an intervener.} On \textds{oracle} and \textds{discovered} concepts, the LLM and the simulated expert each answer about 80\% of the concepts correctly. Yet with the expert, the CBM recovers 42\% of the gain from perfect interventions, and with the LLM significantly less, only 11\%. The reason is that the LLM's errors concentrate on \textcp{HasKnees}, the concept that needs correction most often, where it predicts at chance level. In contrast, the expert is $80\%$ accurate on every concept. When we swap their answers on \textcp{HasKnees} alone, the CBM's gains switch places (see \cref{Table::KneeSwap}).

\paragraph{Robustness to Alignment Constraints}
\label{Sec::Alignment}

\begin{wrapfigure}{R}{0.55\linewidth}
    \centering
    \resizebox{\linewidth}{!}{\begin{tikzpicture}
\definecolor{raBlue}{HTML}{3B6FB6}
\definecolor{raRed}{HTML}{C44E52}
\definecolor{raText}{HTML}{4D4D4D}
\definecolor{raGrid}{HTML}{DFE4E9}
\colorlet{raBlueL}{raBlue!55!white}
\colorlet{raRedL}{raRed!55!white}
\newcommand{\raRm}[1]{\fontfamily{phv}\fontsize{#1}{#1}\selectfont}
\begin{axis}[
  scale only axis, width=6cm, height=3.7cm, clip=false,
  xmin=75, xmax=96, ymin=0.3, ymax=5.3,
  axis x line*=bottom, axis y line*=left,
  axis line style={-, black!55, line width=0.5pt},
  xtick={75,80,85,90,95}, xticklabel={\pgfmathprintnumber[assume math mode=true]{\tick}\%},
  ytick={1,2,3.5,4.5},
  yticklabels={Constrained CBM,CBM,Constrained CBM,CBM},
  y tick style={draw=none}, yticklabel style={anchor=east, xshift=-2pt, font=\fontfamily{phv}\fontsize{8.5}{9.5}\selectfont, text=black!75},
  every tick/.style={black!55, line width=0.4pt}, tick align=inside, tick pos=left,
  xmajorgrids, grid style={raGrid, line width=0.4pt},
  ticklabel style={font=\raRm{9}, text=black!75},
  xlabel={Accuracy}, xlabel style={font=\raRm{9}, text=raText, yshift=2pt},
]
\draw[raBlue, opacity=0.25, line width=7pt, line cap=round] (axis cs:84.24,4.5) -- (axis cs:84.76,4.5);
\draw[raBlue, opacity=0.25, line width=7pt, line cap=round] (axis cs:91.54,4.5) -- (axis cs:92.44,4.5);
\draw[raBlueL, opacity=0.25, line width=7pt, line cap=round] (axis cs:89.11,3.5) -- (axis cs:90.51,3.5);
\draw[raBlueL, opacity=0.25, line width=7pt, line cap=round] (axis cs:90.25,3.5) -- (axis cs:91.23,3.5);
\draw[raRed, opacity=0.25, line width=7pt, line cap=round] (axis cs:76.81,2) -- (axis cs:78.45,2);
\draw[raRed, opacity=0.25, line width=7pt, line cap=round] (axis cs:85.32,2) -- (axis cs:86.06,2);
\draw[raRedL, opacity=0.25, line width=7pt, line cap=round] (axis cs:79.94,1) -- (axis cs:82.16,1);
\draw[raRedL, opacity=0.25, line width=7pt, line cap=round] (axis cs:82.18,1) -- (axis cs:84.24,1);
\draw[raBlue, line width=2.4pt, -{Stealth[length=7pt,width=6pt]}] (axis cs:84.5,4.5) -- (axis cs:91.54,4.5);
\filldraw[fill=white, draw=raBlue, line width=1.3pt] (axis cs:84.5,4.5) circle [radius=2.7pt];
\filldraw[fill=raBlue, draw=raBlue, line width=1.3pt] (axis cs:91.99,4.5) circle [radius=2.7pt];
\node[anchor=base west, inner sep=0, font=\raRm{9}, text=raText] at (axis cs:92.79,4.38) {+7.5\%{\raRm{7}$\pm$0.5}};
\draw[raBlueL, line width=2.4pt, -{Stealth[length=7pt,width=6pt]}] (axis cs:89.81,3.5) -- (axis cs:90.29,3.5);
\filldraw[fill=white, draw=raBlueL, line width=1.3pt] (axis cs:89.81,3.5) circle [radius=2.7pt];
\filldraw[fill=raBlueL, draw=raBlueL, line width=1.3pt] (axis cs:90.74,3.5) circle [radius=2.7pt];
\node[anchor=base west, inner sep=0, font=\raRm{9}, text=raText] at (axis cs:91.54,3.38) {+0.9\%{\raRm{7}$\pm$0.5}};
\draw[raRed, line width=2.4pt, -{Stealth[length=7pt,width=6pt]}] (axis cs:77.63,2) -- (axis cs:85.24,2);
\filldraw[fill=white, draw=raRed, line width=1.3pt] (axis cs:77.63,2) circle [radius=2.7pt];
\filldraw[fill=raRed, draw=raRed, line width=1.3pt] (axis cs:85.69,2) circle [radius=2.7pt];
\node[anchor=base west, inner sep=0, font=\raRm{9}, text=raText] at (axis cs:86.49,1.88) {+8.1\%{\raRm{7}$\pm$0.8}};
\draw[raRedL, line width=2.4pt, -{Stealth[length=7pt,width=6pt]}] (axis cs:81.05,1) -- (axis cs:82.76,1);
\filldraw[fill=white, draw=raRedL, line width=1.3pt] (axis cs:81.05,1) circle [radius=2.7pt];
\filldraw[fill=raRedL, draw=raRedL, line width=1.3pt] (axis cs:83.21,1) circle [radius=2.7pt];
\node[anchor=base west, inner sep=0, font=\raRm{9}, text=raText] at (axis cs:84.01,0.88) {+2.2\%{\raRm{7}$\pm$0.5}};
\draw[raBlue, line width=0.6pt] ([xshift=-74pt]axis cs:75,4.75) -- ([xshift=-77pt]axis cs:75,4.75) -- ([xshift=-77pt]axis cs:75,3.25) -- ([xshift=-74pt]axis cs:75,3.25);
\node[rotate=90, anchor=south, font=\fontfamily{cmtt}\fontsize{8.5}{9.5}\selectfont, text=raBlue] at ([xshift=-80pt]axis cs:75,4) {oracle};
\draw[raRed, line width=0.6pt] ([xshift=-74pt]axis cs:75,2.25) -- ([xshift=-77pt]axis cs:75,2.25) -- ([xshift=-77pt]axis cs:75,0.75) -- ([xshift=-74pt]axis cs:75,0.75);
\node[rotate=90, anchor=south, font=\fontfamily{cmtt}\fontsize{8.5}{9.5}\selectfont, text=raRed] at ([xshift=-80pt]axis cs:75,1.5) {discovered};
\end{axis}
\end{tikzpicture}%
}
    \caption{Our benchmarks show how an alignment constraint changes what a \CBM{} gains from interventions. Each arrow runs from the accuracy before interventions (open marker) to the accuracy with all concepts corrected (filled marker; shaded regions show standard errors); its length is $\incacc$, printed at the tip. Constraining the weight on \captioncp{HasKnees} raises accuracy before interventions but removes most of the gain.}
    \label{Fig::robot_alignment}
\end{wrapfigure}

Concept-based models allow practitioners to encode domain knowledge as \emph{alignment constraints} on front-end weights, for example by fixing a feature sign to match clinical expectations~\citep{joseph2025cognitive, adler2022meeting}. Because we know the true concept weights, we can measure the benefit of interventions before and after a constraint is applied.

We constrain the weight on \textcp{HasKnees} to be non-negative, which sets it to zero. The true weight is negative, yet a practitioner could plausibly choose this constraint, because knees appear in both classes and the detector cannot recognize them. The constrained \CBM{} is only slightly less accurate in training ($91.5\%$ vs.\ $93.0\%$), so a standard accuracy check would accept the constraint.

\textbf{An alignment constraint can remove the benefit of interventions.} The constrained \CBM{} no longer relies on \textcp{HasKnees}, a concept that its detector predicts at chance, so before interventions it is significantly more accurate than the unconstrained \CBM{} on \textds{oracle} and \textds{discovered} concepts. \textcp{HasKnees}, however, is also the concept that needs correction most often: the constrained \CBM{} gains only $1$ to $2$ points from correcting every concept and the unconstrained \CBM{} about $8$, so after interventions the unconstrained \CBM{} is significantly more accurate.

\paragraph{When the Goal Is Automation}
\label{Sec::AutomationValidation}

\begin{table}[t]
    \centering
    \resizebox{\linewidth}{!}{
    \renewcommand{\arraystretch}{1.05}
\setlength{\tabcolsep}{3pt}
\begin{tabular}{@{}ll rrrr rrrr@{}}
& & \multicolumn{8}{c}{\textbf{Interventions}} \\
\cmidrule(lr){3-10}
& & \multicolumn{4}{c}{$\wrk$} & \multicolumn{4}{c}{$\cov$} \\
\cmidrule(lr){3-6} \cmidrule(lr){7-10}
\textbf{Resolution} & \textbf{Model} & \multicolumn{1}{c}{none} & \multicolumn{1}{c}{$\leq 1$ concept} & \multicolumn{1}{c}{$\leq 3$ concepts} & \multicolumn{1}{c}{all concepts} & \multicolumn{1}{c}{none} & \multicolumn{1}{c}{$\leq 1$ concept} & \multicolumn{1}{c}{$\leq 3$ concepts} & \multicolumn{1}{c}{all concepts} \\
\midrule
\multirow{5}{*}{\textds{50\,px}}
 & \DNN{}     & 2.6\%{\scriptsize$\pm$0.8} & -- & -- & -- & 2.6\%{\scriptsize$\pm$0.8} & -- & -- & -- \\
 & \CBM{}     & 79.8\%{\scriptsize$\pm$5.9} & 81.0\%{\scriptsize$\pm$6.2} & 81.9\%{\scriptsize$\pm$6.6} & 74.5\%{\scriptsize$\pm$11.9} & 79.8\%{\scriptsize$\pm$5.9} & 81.8\%{\scriptsize$\pm$6.0} & 84.0\%{\scriptsize$\pm$6.0} & 88.5\%{\scriptsize$\pm$6.0} \\
 & \CEM{}     & 72.5\%{\scriptsize$\pm$9.5} & 73.5\%{\scriptsize$\pm$9.7} & 80.5\%{\scriptsize$\pm$9.3} & \textbf{86.1\%}{\scriptsize$\boldsymbol{\pm}$\textbf{9.6}} & 72.5\%{\scriptsize$\pm$9.5} & 74.2\%{\scriptsize$\pm$9.7} & 82.0\%{\scriptsize$\pm$9.3} & 89.5\%{\scriptsize$\pm$10.0} \\
 & \ProbCBM{} & 79.5\%{\scriptsize$\pm$2.6} & 80.2\%{\scriptsize$\pm$2.3} & 80.3\%{\scriptsize$\pm$2.4} & 80.5\%{\scriptsize$\pm$2.7} & 79.5\%{\scriptsize$\pm$2.6} & 81.0\%{\scriptsize$\pm$2.2} & 82.5\%{\scriptsize$\pm$2.1} & \textbf{92.9\%}{\scriptsize$\boldsymbol{\pm}$\textbf{1.1}} \\
 & \ECBM{}    & 53.1\%{\scriptsize$\pm$11.8} & 66.7\%{\scriptsize$\pm$12.3} & 67.1\%{\scriptsize$\pm$12.5} & 58.0\%{\scriptsize$\pm$13.8} & 53.1\%{\scriptsize$\pm$11.8} & 67.7\%{\scriptsize$\pm$12.4} & 68.9\%{\scriptsize$\pm$12.5} & 69.3\%{\scriptsize$\pm$12.5} \\
\midrule
\multirow{5}{*}{\textds{10\,px}}
 & \DNN{}     & 4.3\%{\scriptsize$\pm$1.3} & -- & -- & -- & 4.3\%{\scriptsize$\pm$1.3} & -- & -- & -- \\
 & \CBM{}     & 58.8\%{\scriptsize$\pm$3.8} & 63.2\%{\scriptsize$\pm$4.0} & \textbf{66.6\%}{\scriptsize$\boldsymbol{\pm}$\textbf{3.8}} & 66.6\%{\scriptsize$\pm$2.8} & 58.8\%{\scriptsize$\pm$3.8} & 64.7\%{\scriptsize$\pm$3.9} & 70.6\%{\scriptsize$\pm$3.4} & \textbf{88.5\%}{\scriptsize$\boldsymbol{\pm}$\textbf{1.0}} \\
 & \CEM{}     & 36.4\%{\scriptsize$\pm$8.6} & 38.3\%{\scriptsize$\pm$9.4} & 41.6\%{\scriptsize$\pm$10.6} & 53.4\%{\scriptsize$\pm$11.8} & 36.4\%{\scriptsize$\pm$8.6} & 39.5\%{\scriptsize$\pm$9.5} & 44.9\%{\scriptsize$\pm$10.8} & 64.7\%{\scriptsize$\pm$12.6} \\
 & \ProbCBM{} & 39.6\%{\scriptsize$\pm$6.9} & 40.7\%{\scriptsize$\pm$7.6} & 40.5\%{\scriptsize$\pm$8.4} & 26.5\%{\scriptsize$\pm$14.7} & 39.6\%{\scriptsize$\pm$6.9} & 42.8\%{\scriptsize$\pm$7.4} & 46.5\%{\scriptsize$\pm$7.6} & 66.3\%{\scriptsize$\pm$7.4} \\
 & \ECBM{}    & 34.5\%{\scriptsize$\pm$11.1} & 52.7\%{\scriptsize$\pm$11.1} & 54.0\%{\scriptsize$\pm$11.9} & 43.4\%{\scriptsize$\pm$15.0} & 34.5\%{\scriptsize$\pm$11.1} & 54.0\%{\scriptsize$\pm$11.2} & 56.6\%{\scriptsize$\pm$11.9} & 58.5\%{\scriptsize$\pm$12.3} \\
\bottomrule
\end{tabular}

    }
    \caption{Our benchmarks show how interventions and the quality of concept detection change the work a model automates. We report \captionwrk{} and \captioncov{} without interventions (none) and after verifying up to $1$, $3$, or all concepts per deferred board, at a selective-accuracy target of $\alpha_{\textrm{target}}{=}0.99$, at two image resolutions (other targets in \cref{Appendix::AutomationSensitivity}). Every model automates less work at 10\,px than at 50\,px, with or without interventions.}
    \label{Table::sudoku_px}
\end{table}

We test our automation setup with all four architectures and the \DNN{} as selective classifiers that must reach a selective-accuracy target of $\alpha_{\textrm{target}}{=}0.99$ (we report $0.90$ and $0.95$ in \cref{Appendix::AutomationSensitivity}). We report \wrk{} and \cov{} of each model in \cref{Table::sudoku_px}.

\textbf{Concept-based models automate far more work than the \DNN{}.} At 50\,px, every concept-based model labels at least half of the boards at the target without a single intervention, whereas the \DNN{} labels almost none. A likely reason is that a concept-based model splits the task in two: it learns to detect each constraint and, separately, the rule that combines them. The \DNN{} instead has to learn the whole validity rule at once, and it is rarely confident enough at $\alpha_{\textrm{target}}{=}0.99$ to provide a prediction.

\textbf{Increasing the number of interventions may not lead to more saved work.} At 50\,px, every model automates more work with up to three concepts verified per deferred board, and this gain is significant for the \ProbCBM{} and the \ECBM{}. Beyond three concepts, the effect depends on the front-end model. The front-end models of the \CBM{}, \ProbCBM{}, and \ECBM{} remain uncertain about invalid boards that violate only a few constraints, so $38\%$ to $49\%$ of their deferred boards remain deferred even with every concept verified. The front-end model of the \CEM{} is more certain about these boards, and only the \CEM{} automates significantly more work with every concept verified (see \cref{Appendix::AutomationCeiling}).

\textbf{Small errors in concept detection may become a large loss in automated work.} At 10\,px per cell instead of 50\,px, each concept detector is only a few points less accurate. A board depends on 27 concepts, however, so the share of boards with at least one wrong concept grows from a quarter to more than half. With these errors, every architecture is less certain about the label, so it defers more boards and automates less work (see \cref{Table::sudoku_px}). No architecture recovers all of this work with interventions.

\newsection{Concluding Remarks}
\label{Sec::ConcludingRemarks}

We introduce synthetic benchmarks for concept-based models, which generate benchmark datasets for decision support and automation from a known concept-to-label rule. Our benchmarks address the scarcity of concept-annotated datasets and let researchers train models quickly on data whose concepts they can control and measure. Our benchmarks are meant to diagnose models rather than rank them. Because researchers can change one factor at a time, each finding can be traced to the mechanism that produces it. When a real task shares that mechanism, we expect the finding to transfer. For instance, our automation results are driven by the AND structure of the task; we therefore expect them to hold for document checking, where a valid decision likewise requires many conditions to hold at once. The main limitation of our benchmarks is that they simplify real-world tasks, so the absolute accuracies and savings we report may overestimate performance on real data. We also test only some mechanisms that drive CBM performance, and leave the rest to future work.

\paragraph{Broader Impacts}
Our benchmarks can help researchers find brittle designs early, before spending effort on concept labels or deploying a model. The main risk is that good performance on our benchmarks is read as evidence that a model is safe to deploy. To mitigate this risk, results on our benchmarks should complement validation on real data, not replace it.

\clearpage
\nottoggle{blind}{%
\subsection*{Contribution Statement}
Julian Skirzy\'nski developed the robots benchmark, contributed to extending both benchmarks to \CEM{}, \ProbCBM{}, and \ECBM{}, ran the experiments reported in the paper, and led the writing of the manuscript. Harry Cheon developed the Sudoku benchmark and ran the initial experiments on automation, concept set specification, and missingness. Shreyas Kadekodi developed the text-robot benchmark, ran the initial experiments on concept noise, and contributed to extending both benchmarks to \CEM{}, \ProbCBM{}, and \ECBM{}. Meredith Stewart contributed to the development of the Sudoku benchmark. Berk Ustun conceived the idea and supervised the project. All authors provided feedback and revisions throughout the writing process.
}{}

\mlacknowledgments{%
This work was supported by funding from the National Science Foundation IIS 2313105 and Cooperative Agreement 2433429 through the AI Research Institute on Interaction for AI Assistants (ARIA), and the NIH Bridge2AI Center Grant U54HG012510. This work used Bridges-2 at Pittsburgh Supercomputing Center through allocation CIS260305 from the Advanced Cyberinfrastructure Coordination Ecosystem: Services \& Support (ACCESS) program, which is supported by U.S. National Science Foundation grants 2138259, 2138286, 2138307, 2137603, and 2138296~\citep{boerner2023access}.}

\mlbibliographystyle
{\small\bibliography{concept_benchmark}}

\clearpage
\appendix
\crefalias{section}{appendix}
\crefalias{subsection}{appendix}
\crefname{appendix}{Appendix}{Appendices}
\Crefname{appendix}{Appendix}{Appendices}
\iftoggle{icml}{\onecolumn}{}

\makeatletter
\@ifpackageloaded{lineno}{%
  \setlength{\linenumbersep}{10pt}
  \linenumbers
  \setlength{\linenumberwidth}{2.5em} %
}{}
\makeatother

\mtcaddpart[Supplementary Materials]
\thispagestyle{empty}
{%
\bfseries\Large\vspace{-2.0em}%
\begin{center}
{%
\ourtitleshort{}\\[0.25em]%
Supplementary Materials\\[0.25em]%
}%
\end{center}%
\vspace{-1.0em}
}
\renewcommand\ptctitle{}
\mtcsetfeature{parttoc}{open}{}%
\setlength{\ptcindent}{0pt}
\noptcrule
\parttoc[c]
\clearpage

\section{Notation}
\label{Appendix::Notation}
We list recurring notation in \cref{Table::Notation}. We use $i$ to index examples, $j$ or $k$ to index concepts, and $b$ for the intervention budget.

\begin{table}[h]
    \centering
    \resizebox{0.95\textwidth}{!}{
    \begin{tabular}{ll}
        \textbf{Symbol} & \textbf{Meaning} \\
        \midrule
        $n$ & Number of examples in dataset. \\
        $m$ & Number of concepts. \\
        $\features_i \in \X \subseteq \R^d$ & Feature vector for example $i$ (e.g., pixels). \\
        $\concepts_i \in \C = \{0,1\}^m$ & Ground-truth concept vector for example $i$. \\
        $\outcome_i \in \Y = \{0,1\}$ & Ground-truth label for example $i$. \\
        $\detector_k: \X \to [0,1]$ & Concept detector for concept $k$; predicts $\Pr{c_{i,k} = 1 \mid \features_i}$. \\
        $\frontend: \C \to [0,1]$ & Front-end model mapping concepts to soft labels. \\
        $\pipeline: \X \to \Y$ & Full CBM pipeline outputting hard labels. \\
        $\baseline: \X \to \Y$ & Baseline model used for comparison (e.g., a black-box \DNN). \\
        $\outcomepred_i = \pipeline(\features_i)$ & Hard label prediction for example $i$. \\
        $\threshold \in [0, 0.5]$ & Abstention threshold for selective classification. \\
        $\alpha_{\textrm{target}}$ & Selective-accuracy target that sets the abstention threshold $\threshold$. \\
        $\abstain$ & Abstention symbol (model refuses to predict). \\
        \midrule
        \multicolumn{2}{l}{\textbf{Interventions and Metrics}} \\
        \midrule
        $t: \C \times \X \to \C$ & Intervention mapping modifying concept values given input $\features$. \\
        $S(\features) \subseteq \{1,\ldots,m\}$ & Subset of concepts selected for intervention for input $\features$. \\
        $|S(\features)|$ & Number of concepts requiring intervention for input $\features$. \\
        $b$ & Intervention budget: the largest number of concepts intervened on per input. \\
        $\flipProb(S)$ & Responsiveness: probability that intervening on concepts $S$ changes the prediction. \\
        $\flipThreshold$ & Responsiveness threshold above which an input is intervened on ($\flipThreshold = 0.2$). \\
        $\gain(\pipeline, \baseline)$ & Accuracy gain of CBM $\pipeline$ over baseline $\baseline$ after interventions. \\
        $\textsf{SelectiveAccuracy}(\pipeline)$ & Accuracy on non-abstained examples. \\
        $\cov(\pipeline)$ & Proportion of examples where model does not abstain. \\
        $\wrk(\pipeline)$ & Work saved: coverage minus the share of concepts verified. \\
        $\incacc$ & Improvement in accuracy due to interventions (same model, before vs.\ after). \\

        \midrule
        \multicolumn{2}{l}{\textbf{Task-Specific Notation}} \\
        \midrule
        $\nBoard$ & Block size of a Sudoku board: a $\nBoard^2 \times \nBoard^2$ board with $\nBoard \times \nBoard$ blocks (e.g., $\nBoard=3$ for $9 \times 9$ boards). \\
        $\rho$ & Share of training labels flipped in the Sudoku label-noise experiment. \\
        $\crow{j}, \ccol{j}, \cblk{j}$ & Sudoku constraint concepts for row $j$, column $j$, block $j$. \\
        $\textcp{concept\_name}$ & Specific concept names (e.g., \textcp{FootShape}, \textcp{HasKnees}). \\
        $\outcome$ & True robot type in robots benchmark ($\textcp{G}$ for Glorp, $\textcp{D}$ for Drent). \\
        \bottomrule
    \end{tabular}

    }
    \caption{Notation used throughout the paper.}
    \label{Table::Notation}
\end{table}

\section{Decision-Support Benchmark}
\label{Appendix::DecisionSupport}

\subsection{Generation Parameters}
\label{Appendix::RobotsImageParams}
We list the parameters used to generate the \textds{robots} datasets in our experiments.
\begin{itemize}
    \item \textfn{concepts}:
    \textcp{HeadShape} $\in \{\text{square}, \text{round}\}$; \\
    \textcp{BodyShape} $\in \{\text{square}, \text{round}\}$; \\
    \textcp{HasKnees} $\in \{\text{false}, \text{true}\}$; \\
    \textcp{HasElbows} $\in \{\text{false}, \text{true}\}$; \\
    \textcp{HasAntennae} $\in \{\text{false}, \text{true}\}$; \\
    \textcp{EarShape} $\in \{\text{square}, \text{triangle}\}$; \\
    \textcp{MouthType} $\in \{\text{closed}, \text{open}\}$; \\
    \textcp{HandShape} $\in \{\text{round circle}, \text{round oval}, \text{round oval variant}, \text{edgy triangle}, \text{edgy square}, \\ \text{edgy trapezoid}\}$; \\
    \textcp{FootShape} $\in \{\text{flat trapezoid}, \text{flat rounded}, \text{flat square}, \text{flat five-sided}, \text{flat L-shaped}, \text{pointy trapezoid}, \\ \text{pointy rounded}, \text{pointy square}, \text{pointy three-sided}, \text{pointy four-sided}\}$.
    \item \textfn{drop\_concepts}: \textcp{HasElbows} and \textcp{HandShape} are rendered but excluded from the concept set. The \textds{oracle} set keeps $7$ binary concepts, with \textcp{FootShape} binarized to flat vs.\ pointy; the \textds{discovered} set replaces \textcp{FootShape} with one binary concept for each of the six subtypes seen in training ($12$ concepts).
    \item \textfn{skew}: the share of each \textcp{FootShape} subtype in the training sample, which we set for each labeling rule (see \cref{Appendix::BalancedRule}).
    \item \textfn{drawing\_parameters}: $32 \times 32$\,px images in two-tone color.
    \item \textfn{samples\_per\_instance}: four color renderings per robot; the colors carry no information about the label.
    \item \textfn{split\_procedure}: from the pool of rendered robots (unique robots $\times$ color renderings) we draw 10{,}000 test robots, then 3{,}800 training robots so that each foot subtype has at least the share given by \textfn{skew}, and keep the rest for validation. The foot subtypes in each split depend on the labeling rule (see \cref{Appendix::BalancedRule}).
\end{itemize}

\subsection{Labeling Rules}
\label{Appendix::BalancedRule}
Each robot's label is drawn from a probability given by its features. Every rule computes a score from the features of a robot and labels it a Glorp with a probability that is a logistic function of $4.2$ times this score. The factor $4.2$ sets the label noise: a robot with a score of $\pm 0.5$ receives the label that its score favors with probability $89\%$, and a robot with a score of $\pm 1.5$ or more with probability above $99\%$. We use two rules, which differ in the score.

\paragraph{Balanced rule}
The main text uses a rule under which no single concept decides the label. Five visible features add the same amount to the score, pointy feet add more, and knees and elbows subtract:
\begin{align}
    & \prob{\outcome=\textcp{G}\mid \concepts} = \sigma\Bigl(4.2 \cdot \bigl(6 \cdot \indic{\textcp{MouthType} = \textcpp{Closed}} + 6 \cdot \indic{\textcp{BodyShape} = \textcpp{Round}} \nonumber \\
    & \qquad + 6 \cdot \indic{\textcp{HeadShape} = \textcpp{Round}} + 6 \cdot \indic{\textcp{HasAntennae} = \textcpp{True}} + 6 \cdot \indic{\textcp{EarShape} = \textcpp{Triangular}} \nonumber \\
    & \qquad + 8 \cdot \indic{\textcp{FootShape} = \textcpp{Pointy}} - 3 \cdot \indic{\textcp{HasKnees} = \textcpp{True}} - 2 \cdot \indic{\textcp{HasElbows} = \textcpp{True}} - 16.5\bigr)\Bigr).
    \label{Eq::BalancedRule}
\end{align}
The intercept makes both classes equally likely. Only $8\%$ of robots have a score of $\pm 0.5$; every other robot receives the label its score favors with probability above $99\%$, so a model that knows every feature reaches $99\%$ accuracy. \textcp{HasElbows} enters the rule but is not a concept, so it cannot be intervened on. The training sample contains six of the ten \textcp{FootShape} subtypes (pointy four-sided and flat five-sided at $30\%$ each; pointy square, pointy rounded, flat square, and flat trapezoid at $10\%$ each), and the test sample contains all ten in equal proportion.

\paragraph{Sparse rule}
\label{Appendix::RobotsSetupParams}
In \cref{Appendix::RareClass}, we use a rule that depends on three concepts:
\begin{align}
    \prob{\outcome=\textcp{G}\mid \concepts} = \sigma\Bigl(4.2 \cdot \bigl(& 5 \cdot \indic{\textcp{MouthType} = \textcpp{Closed}} + 8 \cdot \indic{\textcp{FootShape} = \textcpp{Pointy}} \nonumber \\
    & - 5 \cdot \indic{\textcp{HasKnees} = \textcpp{True}} + 2\bigr)\Bigr).
    \label{Eq::RobotGT}
\end{align}
The score is negative only for robots with an open mouth, flat feet, and knees, so $87.5\%$ of robots are Glorps. Every score is at least $2$ away from zero, so the labels are effectively deterministic. The training sample under this rule contains pointy four-sided and flat five-sided feet at $49\%$ each, and pointy square, pointy rounded, flat square, and flat trapezoid feet at $0.5\%$ each. The remaining four subtypes do not appear in the generated data.

\subsection{Text Modality and Concept Missingness}
\label{Appendix::RobotTextModality}
Robots can also be described in text. We prompted an LLM (GPT-4o-mini) with each robot's attributes and asked for a one- to three-sentence description in plain language, using the prompts below, and fixed the generations as a template corpus so that every run draws the same captions. The \textfn{template\_complexity} option selects one of three corpora: simple templates (low), semi-structured templates (medium), and these LLM-generated captions (high, the default). The \textfn{corpus\_file} option selects a corpus; the released corpora include variants in which a concept is never stated.

\paragraph{Concept noise via masking/omission}
A caption can hide a concept. We mask an attribute, so that the LLM describes it only generically (``it has feet''), or leave it out, so that the caption says nothing about it; the other attributes are described as usual. The corpora \textfn{foot\_generic}, \textfn{ears\_generic}, and \textfn{no\_antenna} are built this way.

\paragraph{Unstructured description prompt}
\begin{center}
\begin{tcolorbox}[promptbox]
System prompt: You are concise and concrete. Use everyday language. Do not invent
locations or scenarios. \\

User prompt: Using the provided attributes, write a description in plain, spoken
language that sounds like a person describing an image they saw. Keep it
concise and natural, between 1 and 3 sentences. Avoid list-like phrasing,
avoid locations/situations, and focus only on what the attributes imply. \\

Attributes (JSON): \{...\}
\end{tcolorbox}
\end{center}

\paragraph{Masked/omitted prompt for concept noise}
\begin{center}
\begin{tcolorbox}[promptbox]
System prompt: You describe simple robots from attribute lists. \\

User prompt: You are given attributes of a robot as JSON.
Write a single English sentence describing the robot.
Use natural language, not attribute names.
For attributes in 'masked', describe them only generically without revealing
their exact value.
For attributes in 'omitted', do not include any information about them.
All other attributes should be reflected in the sentence.
Do not add extra attributes that are not implied by the JSON. \\

JSON: \{...\}
\end{tcolorbox}
\end{center}

\cref{Table::RobotModalities} shows a Glorp and a Drent with a caption from each corpus.
\begin{table}[H]
\centering

\resizebox{\textwidth}{!}{%

\small
\renewcommand{\arraystretch}{1.2}
\begin{tabular}{l >{\centering\arraybackslash}m{1.3in} >{\raggedright\arraybackslash}m{1.25in} >{\raggedright\arraybackslash}m{1.35in} >{\raggedright\arraybackslash}m{2.2in}}
& & \multicolumn{3}{c}{\textbf{Caption Format}} \\
\cmidrule(lr){3-5}
\multicolumn{2}{@{}l}{\textbf{Robot}} & \textbf{Simple templates} & \textbf{Semi-structured} & \textbf{LLM-generated} \\
\midrule
Glorp & \includegraphics[width=1.25in]{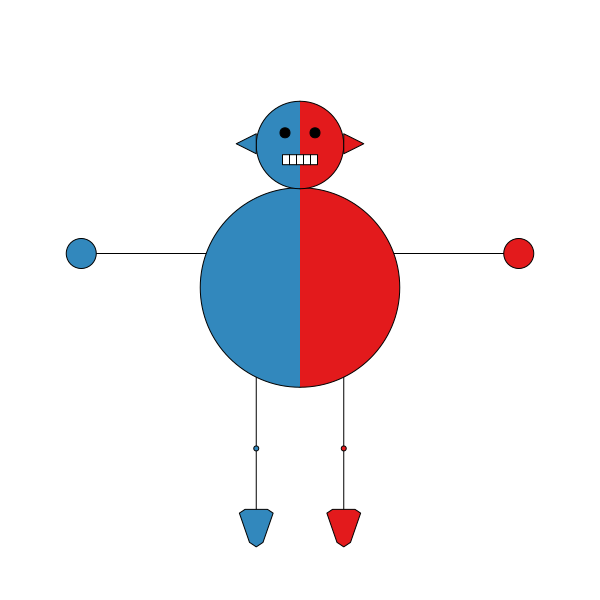} & ``Head is round, body is round, ears triangle, mouth open, hands round\_circle, feet pointy\_trapezoid, knees true, elbows false, antennae false.'' & ``The head isn’t square and the body isn’t square. Knees true, elbows false, antennae false, ears triangle, mouth open, hands round\_circle, feet pointy\_trapezoid.'' & ``The mouth sits open, like it means it, the head keeps its edges—rounded and deliberate, and the feet settle into a tapered trapezoid feet stance. There are no antennae if you’re looking, it has knees, and the hands read round hands, ready and sure. The ears come to a neat point—three-angled, sharp enough to catch the light, its arms have no elbows, and the body holds to a barrel-shaped frame without bending.'' \\
\midrule
Drent & \includegraphics[width=1.25in]{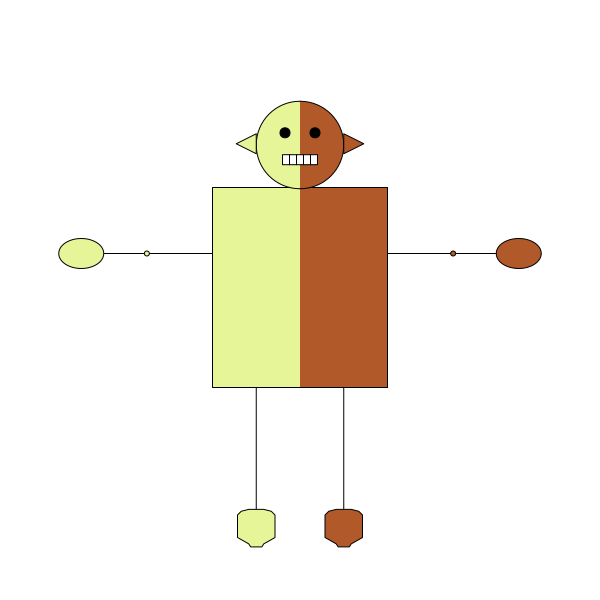} & ``Head = round. Body = square. Knees = false. Elbows = true. Antennae = false. Ears = triangle. Mouth = open. Hands = round\_oval. Feet = flat\_square.'' & ``It’s got a round head on a square body. Knees false, elbows true, antennae false, ears triangle, mouth open, hands round\_oval, and feet flat\_square.'' & ``The hands read wide-oval hands, ready and sure, its arms bend at the elbows, and the head keeps its edges—circular and deliberate. The ears come to a neat point—pointed, sharp enough to catch the light, it has no knees, and there are no antennae if you’re looking. The feet settle into a flat quad pads stance, the mouth sits open, like it means it, and the body holds to a angular frame without bending.'' \\
\bottomrule
\end{tabular}
}
\caption{A Glorp and a Drent from the \captionds{robots} benchmark, each with a caption from the three text corpora and its image.}
\label{Table::RobotModalities}
\end{table}

\paragraph{Concept missingness}
We implement missingness at the concept level and mask missing labels during training. Under MCAR, each concept entry is independently missing with probability $p$. Under MNAR, the probability depends on the value: $1.5p$ when the concept is present and $0.5p$ when it is absent (clipped to $[0, 1]$). Missing labels are ignored by the default loss using an observation mask, so only observed concept labels contribute to training. We do not impute missing concepts unless explicitly stated.

\subsection{Example Studies}
\label{Appendix::RobotSampleBenchmarks}

\begin{itemize}[leftmargin=0pt,label={}]

    \item \emph{Concept Discovery}: One of the key benefits of benchmarking with a synthetic task is that we can test assumptions we would be unable to explore in practice. In this case, we can evaluate how the performance and functionality of CBMs depend on our ability to identify and label concepts. In real-world decision-support tasks, the ground-truth concepts are unknown. It is not clear which concepts best predict, e.g., knee osteoarthritis from X-rays~\citep{koh2020concept}, and such determinations are subject to new scientific findings and our better understanding of the world~\citep[e.g., see the paradigm changes in][]{ahmed200523}. To model this scenario, we can set up two datasets: one that uses the \textcp{FootShape\_is\_Subtype} concepts, and another that uses the ground-truth \textcp{FootShape} concept. We can compare the performance of the CBMs trained on both datasets to see how the models handle concept misspecification. We perform this evaluation in~\cref{Sec::MisspecifiedConcepts}, where these are the \textds{discovered} and \textds{oracle} concepts.

    \item \emph{Human Assistance}: By using low-resolution images, we can make automatic concept detection for \textcp{HasKnees} and \textcp{HasElbows} unreliable and simulate two deployment scenarios. In practice, this setting would reflect computational constraints that require training the model on low-resolution images, while experts still have access to high-resolution versions for interventions. We can also assume that both experts and models have access only to low-resolution images, and that experts must rely on the model for certain concepts. In this way, we can measure the human-AI team's $\gain$ as a function of the accuracy with which experts provide concept labels to determine when interventions provide benefit. We perform a version of this evaluation in \cref{Sec::Noise}, where an expert and an LLM intervene on concepts that the detector cannot recognize.

\end{itemize}

\section{Automation Benchmark}
\label{Appendix::Automation}

\subsection{Board Generation}
The Sudoku generator solves the following MIP for valid boards. The block size is $\nBoard{} \in \mathbb{N}$. Row, column, and value indices range over $\{1,\dots,\nBoard{}^2\}$. Random objective coefficients $w_{i,j,k} \sim \mathcal{U}(0,1)$ break ties across feasible boards.
\begin{samepage}
\begin{subequations}
\label{MIP::Sudoku}
\footnotesize
\renewcommand{\arraystretch}{1.5}
\begin{equationarray}{@{}c@{}r@{\,}c@{\,}l>{\,}l>{\,}l@{\;}}
\max_{x_{ijk}}\qquad & \sum_{i=1}^{\nBoard{}^2}\sum_{j=1}^{\nBoard{}^2}\sum_{k=1}^{\nBoard{}^2} w_{ijk}\, x_{ijk} \nonumber \\[1.5em]
\st
& \sum_{k=1}^{\nBoard{}^2} x_{ijk} & = & 1 & \textrm{for } i, j \in \intrange{\nBoard{}^2} & \mipwhat{one value per cell} \label{Con::Sudoku::Cell} \\
& \sum_{j=1}^{\nBoard{}^2} x_{ijk} & = & 1 & \textrm{for } i, k \in \intrange{\nBoard{}^2} & \mipwhat{each value once per row} \label{Con::Sudoku::Row} \\
& \sum_{i=1}^{\nBoard{}^2} x_{ijk} & = & 1 & \textrm{for } j, k \in \intrange{\nBoard{}^2} & \mipwhat{each value once per column} \label{Con::Sudoku::Col} \\
& \sum_{(i,j) \in \block_{p,q}} x_{ijk} & = & 1 & \textrm{for } p, q \in \intrange{\nBoard{}},\, k \in \intrange{\nBoard{}^2} & \mipwhat{each value once per $\nBoard{}\!\times\!\nBoard{}$ block} \label{Con::Sudoku::Block} \\
& x_{ijk} & \in & \{0,1\} & \textrm{for } i, j, k \in \intrange{\nBoard{}^2} & \mipwhat{set to 1 if cell in row $i$ and column $j$ has value $k$} \label{Con::Sudoku::Binary}
\end{equationarray}
\end{subequations}
\end{samepage}
where $\block_{p,q} = \{(i,j) \mid i \in \{(p{-}1)\nBoard{}+1,\ldots, p\nBoard{}\},\, j \in \{(q{-}1)\nBoard{}+1,\ldots, q\nBoard{}\}\}$ denotes the set of cells in block $(p,q)$.

Invalid boards start from a valid board and apply between $1$ and \textfn{max\_corrupt} corruption actions. These actions swap entries within selected rows or columns, swap whole rows or columns across bands or stacks, or duplicate a value within one row or column. We then recompute the row, column, and block concepts from the corrupted board.

\subsection{Generation Parameters}
The \textds{sudoku} benchmark uses the following parameters. We give the default of each and the value used in our experiments where it differs.

\begin{itemize}
    \item $\nBoard{}$: block size. The default is $3$, which yields a $9 \times 9$ board with values in $\{1,\ldots,\nBoard{}^2\}$.
    \item \textfn{valid\_ratio}: fraction of valid boards. The default is $0.5$.
    \item \textfn{max\_corrupt}: maximum number of corruption actions used to make an invalid board. The default is $9$.
    \item \textfn{n\_samples}: number of boards to generate. The default is $1000$, which we split into $600$/$200$/$200$ boards for training, validation, and testing.
    \item \textfn{target\_accuracy}: minimum accuracy required on kept predictions for selective classification. The default is $0.9$; we use $0.99$ in the main text and $0.90$ and $0.95$ in \cref{Appendix::AutomationSensitivity}.
    \item \textfn{data\_type}: representation type, either \textfn{tabular} or \textfn{image}. This determines the CBM embedding model. We use \textfn{image}.
    \item \textfn{pixels\_per\_cell}: number of pixels per cell in image datasets. We use $50$ and $10$.
    \item \textfn{handwriting}: for image datasets, handwriting variation is generated by varying font, font size, and font color, with a different handwriting for every board. We use handwritten digits.
    \begin{itemize}
        \item \textfn{starter\_percent}: fraction of cells shown as starter values. These use the default non-handwriting font and black color. We use $0.35$.
        \item \textfn{candidates}: candidate-value rendering settings, including the fraction of cells with candidates, the maximum number of candidates per cell, and the candidate color. We use candidates in a fraction $0.22$ of cells.
    \end{itemize}
\end{itemize}

\subsection{Example Studies}
\label{Appendix::SudokuSampleBenchmarks}

\begin{itemize}[leftmargin=0pt,label={}]

\item \emph{Domain Knowledge}: In automation tasks the labels can be noisy (human disagreement or error), but the rule that maps concepts to the label is known, so the front-end model of a CBM can be specified instead of learned~\citep{dash2022review, guo2023di}. We train on the digits of the boards with a share $\rho$ of the training labels flipped and compare a CBM with a learned front-end model, a CBM whose front-end model is the rule $y_i = \prod_j c_{i,j}$, and the DNN, for $\rho = 0, 0.05, \ldots, 0.3$. \cref{Fig::FrontEndModel} reports this comparison.

\item \emph{Intervention Sensitivity}: The quality of concept detection is a parameter of the boards, not of the model: \textfn{pixels\_per\_cell} (the digits are harder to read at fewer pixels per cell), the share of cells with candidate marks (small pencil digits next to the entry), the line widths between cells and blocks, and the handwriting, which differs for every board. Each makes the constraints harder to detect without changing the boards or the rule, so the reduction in work can be compared across architectures at the same difficulty. \cref{Sec::AutomationValidation} varies \textfn{pixels\_per\_cell}; the other three are left for users of the benchmark.

\end{itemize}

\begin{figure}[h]
    \centering
    \resizebox{0.5\linewidth}{!}{\begin{tikzpicture}
\def\femFont{\usefont{T1}{DejaVuSans-TLF}{m}{n}}
\def\femPx{0.72bp}
\foreach \femCorner in {(-89*\femPx,-60*\femPx),(300*\femPx-0.5bp,228*\femPx-0.5bp)}
  {\path \femCorner rectangle ++(0.5bp,0.5bp);}
\begin{axis}[
  scale only axis, width=287*\femPx, height=212*\femPx,
  xmin=-1.5, xmax=31.5, ymin=50, ymax=100,
  xtick={0,5,...,30}, ytick={50,60,...,100},
  xticklabels={0\%,5\%,10\%,15\%,20\%,25\%,30\%},
  yticklabels={50\%,60\%,70\%,80\%,90\%,100\%},
  ticklabel style={font=\femFont\fontsize{11bp}{13bp}\selectfont},
  axis line style={line width=0.8bp}, every tick/.style={line width=0.8bp, black},
  tick align=outside, major tick length=3.5bp, tick pos=left,
  xmajorgrids, ymajorgrids,
  major grid style={draw={rgb,255:red,207;green,207;blue,207}, line width=0.8bp, dash pattern=on 2.96bp off 1.28bp},
  clip=false,
]
\addplot[black, line width=1.5bp, mark=*, mark size=3.35bp, mark options={line width=0.4pt}] coordinates {(0,94.9)(5,94.9)(10,94.9)(15,94.9)(20,94.9)(25,94.9)(30,94.9)};
\addplot[only marks, blue, mark=*, mark size=4.35bp] coordinates {(0,92.42)(5,88.68)(10,88.43)(15,88.43)(20,87.97)(25,87.75)(30,87.51)};
\addplot[only marks, red, mark=square*, mark size=4.1bp] coordinates {(0,88.68)(5,79.21)(10,57.77)(15,72.42)(20,73.84)(25,69.04)(30,69.32)};
\end{axis}
\begin{scope}[font=\femFont\fontsize{16bp}{19bp}\selectfont]
\node[anchor=base] at (142*\femPx,-45.5*\femPx) {Noise Level};
\node[anchor=base, rotate=90] at (-65*\femPx,104.5*\femPx) {Accuracy};
\end{scope}
\end{tikzpicture}%
}
    \caption{Accuracy under label noise of a \CBM{} with a learned front-end model (dots), a \CBM{} whose front-end model is the validity rule (top line), and the DNN (squares). Specifying the front-end model with the known rule makes the \CBM{} immune to label noise, whereas a learned front-end model loses about five points and the \DNN{} about twenty points at $\rho = 0.3$.}
    \label{Fig::FrontEndModel}
\end{figure}

\section{Models and Training}
\label{Appendix::Models}

\subsection{Architectures}
\label{Appendix::Architectures}

\begin{wraptable}{R}{0.42\linewidth}
\centering
\vspace{-1em}
\small
\resizebox{\linewidth}{!}{%
\begin{tabular}{llr}
\textbf{Architecture} & \textbf{Parameter} & \textbf{Value} \\
\midrule
\multirow{4}{*}{\CEM{}} & Embedding size & 16 \\
& Training interv.\ prob. & 0.25 \\
& Concept / task loss wt. & 1.0 / 1.0 \\
& Weight decay & $4\!\times\!10^{-5}$ \\
\midrule
\multirow{5}{*}{\ProbCBM{}} & Hidden dim & 8 \\
& Class hidden dim & 64 \\
& Latent dim & 8 \\
& Inference samples & 50 \\
& Training interv.\ prob. & 0.25 \\
\midrule
\multirow{7}{*}{\ECBM{}} & Hidden size & 64 \\
& Training $\lambda_{xy}/\lambda_{xc}/\lambda_{cy}$ & 3/1/1 \\
& Inference $\lambda$ & 1/1/0.01 \\
& Intervention $\lambda$ & 0/0/3 \\
& Inference lr / patience & 0.1 / 10 \\
& Flipped concepts / examples & 20\% / 20\% \\
& Weight decay & $10^{-4}$ \\
\bottomrule
\end{tabular}%

}
\caption{Architecture-specific hyperparameters of the \CEM{}, \ProbCBM{}, and \ECBM{}.}
\label{Table::ArchHyperparams}
\vspace{-2\baselineskip}
\end{wraptable}

For the \textds{robots} image baseline, we use a three-block DNN. Each block includes ReLU and 2\texttimes2 max-pooling, and uses 16, 32, and 64 channels, respectively. The backbone is followed by a 128-unit fully connected layer, dropout at 0.5, and a two-class head. For \textds{robots} concept detection, we use a DNN backbone with one linear head per concept.

For \textds{sudoku}, we first train an OCR model to infer digit values from cell images, then train every model on the true digits and evaluate it on the inferred ones. The \textds{sudoku} DNN baseline encodes each digit as a one-hot vector mapped through a linear layer (embedding dimension 16), reshapes the board into a $9\times9$ spatial grid, and applies three convolutional layers (64, 128, 128 channels) followed by adaptive average pooling and a two-layer MLP with dropout 0.2. The \textds{sudoku} CBM uses a group-pooling architecture: it computes mean and max pooling over rows, columns, and blocks to produce 27 group features, each passed through a per-group MLP (hidden dimension 64) to predict one binary concept. The CBM front-end is logistic regression with an L2 penalty (C$=$1.0, max 1000 iterations).

For CEM, ProbCBM, and ECBM, we use the same CNN backbone as the standard CBM on both benchmarks. For ECBM, we use the authors' energy network, losses, gradient inference, and intervention procedure~\citep{xu2024energy}, replacing their pretrained ResNet backbone with this CNN and training with Adam and early stopping. As in the authors' code, we flip $20\%$ of the concepts in $20\%$ of the training examples before they enter the term that predicts the label from the concepts. We list the architecture-specific hyperparameters in \cref{Table::ArchHyperparams}.

\subsection{Training}
We run every experiment on ten seeds: $1014$--$1023$ for \textds{robots} and $171$--$180$ for \textds{sudoku}. For \textds{robots}, we train the concept detectors and the DNN with Adam at a learning rate of $10^{-3}$ for up to 50 epochs, and stop early with a patience of 10 epochs. Early stopping uses the mean validation F1 across concepts for the detectors and the validation loss for the DNN. For \textds{sudoku}, we train the concept detectors with Adam at a learning rate of $10^{-3}$ for up to 100 epochs with a patience of 20, and the DNN for up to 20 epochs with a patience of 5. The \CBM{} front-end is a logistic regression on the concept labels, trained to convergence with at most 1000 iterations. The digit recognizer that feeds the \textds{sudoku} concept detectors trains for up to five epochs with early stopping on held-out cells of the training boards, so the handwriting of the validation and test boards is never seen.

\paragraph{Compute and licenses}
We ran the experiments on university and national GPU clusters with a single GPU per run. A model trains in minutes, and generating a \textds{sudoku} dataset takes about five minutes per seed. Our implementations of the \CEM{} and \ProbCBM{} use the code of \citet{zarlenga2022concept}, and our implementation of \ECBM{} follows the code of \citet{xu2024energy}; both are released under the MIT license. We release our code and benchmark datasets under the MIT license.

\subsection{Intervention Policy}
\label{Appendix::InterventionPolicy}
We select the concepts to intervene on by the responsiveness of the label to changes in their values. The two use cases measure responsiveness differently because they act on different outputs: decision support acts on the predicted label, and automation on whether the label probability leaves the abstention band.

\paragraph{Decision support}
We measure the likelihood of prediction $i$ changing upon interventions on concepts $S_i$ using \emph{responsiveness} $\flipProb(S_i)$:
\begin{align*}
\flipProb(S_i) = \sum_{a \in \{0,1\}^{|S_i|}} p(\va\mid S_i) \cdot \mathbbm{1}[\outcomepred_i \neq \outcomepred_i^a]
\end{align*}
where:
\begin{itemize}
\item $p(\va\mid S_i) = \prod_{j \in S_i} \conceptpred_{i,j}^{a_j} (1-\conceptpred_{i,j})^{1-a_j}$ is the probability that human intervention would set values of concepts $S_i$ to $\va$
\item $\outcomepred_i^a$ is the prediction when concepts $S_i$ are set to $\va$.
\end{itemize}
We intervene on an instance only if its maximum responsiveness exceeds $\flipThreshold = 0.2$, on the concepts that maximize it, $S_i^* = \arg\max_{|S_i| \leq b} \flipProb(S_i)$. With the budget ``all'', we correct every concept of the selected instances.

\paragraph{Automation}
We verify the concepts of a deferred board one at a time. Following \citet{joren2024classification}, we score each unverified concept $k$ by the expected change in the label probability per unit of cost, $\hat{q}_k(1-\hat{q}_k)\,\lvert \frontend(\hat{q}; \hat{q}_k{\leftarrow}1) - \frontend(\hat{q}; \hat{q}_k{\leftarrow}0)\rvert / c_k$, where $\hat{q}_k$ is the predicted probability of concept $k$, $\frontend(\hat{q}; \hat{q}_k{\leftarrow}v)$ is the label probability with concept $k$ set to $v$, and $c_k$ is the cost of verifying it. We verify the concept with the highest score, set it to its true value, and recompute the label probability. We stop when the board leaves the abstention band, when no unverified concept changes the label probability, or after $b$ interventions.

\section{Supporting Material for Decision Support}
\label{Appendix::DecisionSupportResults}

In what follows, we describe the tests behind our findings on decision support and present additional results. We report results over ten seeds unless stated otherwise.

\subsection{Architectures on Misspecified Concepts}
\label{Appendix::ArchitectureDiagnostics}

We describe the two tests behind our analysis of the \CEM{} and \ECBM{} in \cref{Sec::Architectures}. Both use models trained on \textds{discovered} concepts on four seeds.

\begin{table}[t]
\begin{minipage}[t]{0.49\linewidth}
\centering
\resizebox{\linewidth}{!}{%
\begin{tabular}{@{}lrrrrrr@{}}
& \multicolumn{2}{c}{\textheader{Probe (unseen)}} & \multicolumn{2}{c}{\textheader{Logit Gap}} & \multicolumn{2}{c}{\textheader{Accuracy}} \\
\cmidrule(lr){2-3} \cmidrule(lr){4-5} \cmidrule(lr){6-7}
\textheader{Seed} & within & from training & seen & unseen & seen & unseen \\
\midrule
1 & 100.0\% & 89.9\% & 11.3 & 7.7 & 90.4\% & 83.1\% \\
2 & 99.4\% & 57.6\% & 7.6 & 1.6 & 90.6\% & 74.1\% \\
3 & 100.0\% & 29.6\% & 8.4 & 0.1 & 89.9\% & 73.9\% \\
4 & 100.0\% & 47.5\% & 13.3 & 0.9 & 91.1\% & 71.2\% \\
\bottomrule
\end{tabular}%

}
\caption{We report the accuracy of the two probes on robots with an unseen foot type, the logit gap between pointy and flat feet, and the accuracy of the \CEM{} before any intervention. The \CEM{} encodes the foot shape of unseen foot types, but its label predictor makes little use of it.}
\label{Table::CEMProbes}
\end{minipage}\hfill
\begin{minipage}[t]{0.49\linewidth}
\centering
\resizebox{\linewidth}{!}{%
\begin{tabular}{@{}lrrrrrr@{}}
& \multicolumn{3}{c}{\textheader{With concepts}} & \multicolumn{3}{c}{\textheader{Without concepts}} \\
\cmidrule(lr){2-4} \cmidrule(lr){5-7}
\textheader{Seed} & all & seen & unseen & all & seen & unseen \\
\midrule
1 & 78.0\% & 85.4\% & 67.0\% & 84.9\% & 89.2\% & 78.4\% \\
2 & 72.1\% & 74.4\% & 68.8\% & 85.6\% & 88.5\% & 81.4\% \\
3 & 74.3\% & 79.3\% & 66.8\% & 79.4\% & 83.0\% & 74.2\% \\
4 & 73.8\% & 81.4\% & 62.5\% & 79.9\% & 83.6\% & 74.3\% \\
\bottomrule
\end{tabular}%

}
\caption{We report the accuracy of \ECBM{} before any intervention when it is trained with and without learning the concepts. \ECBM{} is more accurate on every seed when it does not learn the concepts.}
\label{Table::ECBMAblation}
\end{minipage}
\end{table}

\paragraph{Probes on the \CEM{}}
Robots with an unseen foot type have every \textcp{FootShape} concept equal to zero, and the training sample contains none of them. For each robot, we take the six \textcp{FootShape} embeddings that the \CEM{} passes to its label predictor before any intervention. We fit a logistic regression that predicts whether the foot is pointy from these embeddings in two ways: on robots with an unseen foot type, evaluated by 5-fold cross-validation (\emph{within}), and on the training robots, evaluated on robots with an unseen foot type (\emph{from training}). We also compute the \emph{logit gap}, i.e., the mean difference in the label predictor's logit for a Glorp between robots with pointy and flat feet. We report the results in \cref{Table::CEMProbes}. The probe within unseen foot types is nearly perfect, whereas the probe from training is not, and the logit gap is smaller for unseen foot types on every seed. The \CEM{} thus encodes the foot shape of unseen types, but its label predictor makes little use of it.

\paragraph{\ECBM{} without learning the concepts}
\ECBM{} encodes the image once and uses the encoding to detect the concepts, to predict the label from the image, and to predict the label from the concepts. Before any intervention, it predicts the label from the image. We retrain \ECBM{} with the weights of the two concept terms in its loss set to zero, so that the encoding is trained only to predict the label, and compare it to the default \ECBM{} (see \cref{Table::ECBMAblation}). Without learning the concepts, \ECBM{} is more accurate on every seed, on robots with seen and unseen foot types alike.

\subsection{Leakage Measures}
\label{Appendix::Leakage}

\begin{table}[h!]
\centering
\small
\begin{tabular}{lcc}
\textheader{Model} & \textheader{CTL} & \textheader{ICL} \\
\midrule
\CBM{}     & 0.0523 & 0.0124 \\
\CEM{}     & 0.0982 & 0.0240 \\
\ProbCBM{} & 0.1084 & 0.0414 \\
\ECBM{}    & 0.0017 & 0.0000 \\
\bottomrule
\end{tabular}

\caption{We report the leakage of each architecture trained on \captionds{discovered} concepts. \ProbCBM{} has the highest leakage on both measures.}
\label{Table::Leakage}
\end{table}

We compute two published measures of leakage for each architecture trained on \textds{discovered} concepts on one seed~\citep{parisini2025leakage}. Concept-task leakage (\textsf{CTL}) measures how much more information a predicted concept carries about the label than its true value does, averaged over concepts. Interconcept leakage (\textsf{ICL}) measures how much more information two predicted concepts carry about each other than their true values do, averaged over pairs of concepts. Both are zero for a model whose concept predictions carry no information beyond the concept values. We report them in \cref{Table::Leakage}. \ProbCBM{} has the highest leakage on both measures. \ECBM{} scores near zero because it predicts the label from the image rather than from its concepts.

\subsection{Automated Concepts and Interventions}
\label{Appendix::AutomationDiagnostics}

\paragraph{Models} The \textds{llm\_written} concepts were written by Gemini 2.5 Pro. The LLM that intervenes is Gemini 2.5 Flash-Lite, which answers each concept query from a $224$-pixel image of the robot, whereas the concept-based models see the robots at $32$ pixels.

\begin{table}[h!]
\centering
\small
\resizebox{0.6\linewidth}{!}{%
\begin{tabular}{@{}llrr@{}}
& & \multicolumn{2}{c}{\textheader{Change in accuracy upon interventions}} \\
\cmidrule(lr){3-4}
\textheader{Dataset} & \textheader{Model} & \textheader{Own concept predictions} & \textheader{Ground-truth labels} \\
\midrule
\multirow{4}{*}{\textds{oracle}} & \CBM{} & $0.0\%$ & $+25.4\%$ \\
 & \CEM{} & ${-}16.4\%$ & $+18.0\%$ \\
 & \ProbCBM{} & ${-}15.3\%$ & $+3.2\%$ \\
 & \ECBM{} & ${-}11.2\%$ & ${-}1.6\%$ \\
\midrule
\multirow{4}{*}{\textds{discovered}} & \CBM{} & $0.0\%$ & $+27.7\%$ \\
 & \CEM{} & ${-}18.0\%$ & $+9.3\%$ \\
 & \ProbCBM{} & ${-}17.7\%$ & ${-}6.7\%$ \\
 & \ECBM{} & ${-}6.0\%$ & $+3.5\%$ \\
\midrule
\multirow{4}{*}{\textds{machine\_labeled}} & \CBM{} & ${-}11.3\%$ & ${-}14.7\%$ \\
 & \CEM{} & ${-}3.6\%$ & ${-}1.0\%$ \\
 & \ProbCBM{} & ${-}13.8\%$ & ${-}9.7\%$ \\
 & \ECBM{} & ${-}44.5\%$ & ${-}43.2\%$ \\
\midrule
\multirow{4}{*}{\textds{llm\_written}} & \CBM{} & ${-}7.8\%$ & ${-}7.8\%$ \\
 & \CEM{} & ${-}5.1\%$ & ${-}3.7\%$ \\
 & \ProbCBM{} & ${-}17.1\%$ & ${-}15.3\%$ \\
 & \ECBM{} & ${-}42.3\%$ & ${-}42.9\%$ \\
\midrule
\multirow{4}{*}{\textds{clip\_words}} & \CBM{} & ${-}5.2\%$ & ${-}5.2\%$ \\
 & \CEM{} & ${-}5.3\%$ & ${-}4.1\%$ \\
 & \ProbCBM{} & ${-}17.0\%$ & ${-}15.2\%$ \\
 & \ECBM{} & ${-}40.2\%$ & ${-}41.7\%$ \\
\bottomrule
\end{tabular}

}
\caption{We report the change in accuracy on the robots intervened on, averaged over correcting up to $1$, $3$, and all concepts, when interventions supply the model's own concept predictions or the ground-truth concept labels. With automated concepts, the model's own predictions lower accuracy by a similar amount as the ground-truth labels.}
\label{Table::OwnAnswers}
\end{table}

\paragraph{Interventions with the model's own answers}
An intervention replaces the output of the concept detector for the selected concepts. The \CBM{} sets the concept value, the label-free \CBM{} sets the concept's score to the 5th or 95th percentile of its training scores, the \CEM{} and \ProbCBM{} replace the concept's embedding with the embedding for ``present'' or ``absent'', and \ECBM{} switches to predicting the label from the concept values. To separate the effect of this replacement from the effect of the supplied value, we repeat the interventions and supply the value that the model already predicts, i.e., its predicted probability thresholded at $0.5$. In this case, no concept value changes, and the intervention policy selects the same robots and concepts. We report the change in accuracy on the robots intervened on in \cref{Table::OwnAnswers}.

\paragraph{Exchanging the answers of the LLM and the expert}
For the \CBM{} trained on \textds{oracle} and on \textds{discovered} concepts, we repeat the interventions of the LLM and of the expert, and exchange their answers on \textcp{HasKnees} while keeping the selected robots and concepts fixed. We report the results in \cref{Table::KneeSwap}. The LLM is correct on \textcp{HasKnees} for $50\%$ of robots, whereas the expert is correct for $80\%$. On \textds{oracle} concepts, exchanging their answers on any other single concept changes accuracy by at most $1.2$ points.

\begin{table}[h!]
\centering
\small
\resizebox{\textwidth}{!}{%
\begin{tabular}{@{}lrrrrrr@{}}
& \multicolumn{6}{c}{\textbf{Interventions}} \\
\cmidrule(lr){2-7}
& \multicolumn{3}{c}{\textds{oracle}} & \multicolumn{3}{c}{\textds{discovered}} \\
\cmidrule(lr){2-4} \cmidrule(lr){5-7}
\textheader{Answers} & \multicolumn{1}{c}{$\leq 1$ concept} & \multicolumn{1}{c}{$\leq 3$ concepts} & \multicolumn{1}{c}{all concepts} & \multicolumn{1}{c}{$\leq 1$ concept} & \multicolumn{1}{c}{$\leq 3$ concepts} & \multicolumn{1}{c}{all concepts} \\
\midrule
Expert on every concept & 88.6\% & 86.6\% & 86.0\% & 81.8\% & 81.8\% & 80.5\% \\
LLM on every concept & 83.9\% & 83.5\% & 83.6\% & 80.5\% & 80.7\% & 80.3\% \\
\midrule
LLM, with the expert's answers on \textcp{HasKnees} & 88.5\% & 86.5\% & 85.7\% & 81.7\% & 80.8\% & 80.5\% \\
Expert, with the LLM's answers on \textcp{HasKnees} & 84.0\% & 83.6\% & 84.0\% & 80.6\% & 80.9\% & 80.2\% \\
\bottomrule
\end{tabular}

}
\caption{We report the accuracy of the \CBM{} when the LLM and the expert exchange their answers on \captioncp{HasKnees}. Before interventions, the \CBM{} is correct on $84.5\%$ of robots with \captionds{oracle} concepts and on $77.6\%$ with \captionds{discovered} concepts. Exchanging the answers on this one concept exchanges the results of the two interveners.}
\label{Table::KneeSwap}
\end{table}

\subsection{Real Datasets}
\label{Appendix::RealData}

We test whether our result on automated concepts (\cref{Sec::Noise}) holds on two real datasets with concept annotations. On each dataset, we train a \CBM{} on the human-annotated concepts and a label-free \CBM{}~\citep{oikarinen2023label}, which annotates concepts through CLIP similarity. We then correct the predicted concepts with their true values, using the intervention policy of the \textds{robots} experiments (see \cref{Appendix::InterventionPolicy}). We report the results in \cref{Table::CUB}.

\paragraph{Derm7pt}
\textds{Derm7pt}~\citep{kawahara2018derm} is a dermoscopy dataset with seven concepts annotated by clinicians. With the clinicians' concepts, interventions raise accuracy; with label-free concepts, they lower it at every budget (see \cref{Table::CUB}).

\paragraph{CUB}
We use the first 25 species (${\approx}1{,}400$ images) of \textds{CUB-200}~\citep{wah2011cub} with the raw per-image annotations of all $312$ attributes, i.e., without the class-level denoising of \citet{koh2020concept}, which limits the accuracy of the \CBM{} on the annotated concepts. Correcting every concept raises the accuracy of this \CBM{} and lowers that of the label-free \CBM{} by a larger amount; with a few interventions, both change little (see \cref{Table::CUB}).

\begin{table}[h!]
\centering
\small
\resizebox{\textwidth}{!}{%
\begin{tabular}{@{}lrrrrrr@{}}
& \multicolumn{5}{c}{\textbf{Interventions}} & \\
\cmidrule(lr){2-6}
\textds{Derm7pt} & \multicolumn{1}{c}{none} & \multicolumn{1}{c}{$\leq 1$ concept} & \multicolumn{1}{c}{$\leq 2$ concepts} & \multicolumn{1}{c}{$\leq 5$ concepts} & \multicolumn{1}{c}{all concepts} & \textheader{Change} \\
\midrule
human-annotated & 69.0\% & 72.5\% & 73.6\% & 83.8\% & 83.8\% & $+14.8\%$ \\
label-free & 69.7\% & 62.4\% & 60.0\% & 60.2\% & 68.8\% & $-0.9\%$ \\
\bottomrule
\end{tabular}
\quad
\begin{tabular}{@{}lrrrrr@{}}
& \multicolumn{4}{c}{\textbf{Interventions}} & \\
\cmidrule(lr){2-5}
\textds{CUB} & \multicolumn{1}{c}{none} & \multicolumn{1}{c}{$\leq 1$ concept} & \multicolumn{1}{c}{$\leq 10$ concepts} & \multicolumn{1}{c}{all concepts} & \textheader{Change} \\
\midrule
human-annotated & 39.6\% & 43.6\% & 42.5\% & 70.6\% & $+31.0\%$ \\
label-free & 84.8\% & 83.6\% & 83.3\% & 44.5\% & $-40.3\%$ \\
\bottomrule
\end{tabular}

}
\caption{We report the accuracy of a \CBM{} with human-annotated concepts and of a label-free \CBM{} for each intervention budget on two real datasets, and the change in accuracy with all concepts corrected. Results on \captionds{Derm7pt} are means over five seeds (human-annotated) and three seeds (label-free); results on \captionds{CUB} are from one seed. Interventions help with human-annotated concepts and do not help with label-free concepts.}
\label{Table::CUB}
\end{table}

\subsection{When the Class of Interest Is Rare}
\label{Appendix::RareClass}
\label{Sec::RareClass}

In many decision support tasks, the class of interest is rare, e.g., in medicine~\citep{rotemberg2021patient,mckinney2020international}. A model can then reach high accuracy by predicting the common class, and it is the expert's interventions that must catch the rare cases. To study this setting, we repeat our decision-support experiments under the sparse rule, under which $87.5\%$ of robots are Glorps (see \cref{Appendix::RobotsSetupParams}), and report the results in \cref{Table::BigTableSparse}. As in \cref{Sec::MisspecifiedConcepts}, every architecture is less accurate after interventions on \textds{discovered} concepts than on \textds{oracle} concepts. To isolate the effect of the architecture, we focus on perfect interventions on \textds{oracle} concepts (see \cref{Fig::ArchitectureResponseSparse}). These interventions supply exactly the values that the rule depends on, so any remaining error comes from the architecture.

\begin{table*}[t]
\centering
\small
\resizebox{1.0\textwidth}{!}{\renewcommand{\arraystretch}{1.1}
\begin{tabular}{@{}ll r rrr rrr rrr@{}}
& & \multicolumn{10}{c}{\textbf{Interventions}} \\
\cmidrule(lr){3-12}
& & \multicolumn{1}{c}{none} & \multicolumn{3}{c}{$\leq 1$ concept} & \multicolumn{3}{c}{$\leq 3$ concepts} & \multicolumn{3}{c}{all concepts} \\
\cmidrule(lr){3-3} \cmidrule(lr){4-6} \cmidrule(lr){7-9} \cmidrule(lr){10-12}
\textbf{Dataset} & \textbf{Model} & & \textds{perfect} & \textds{expert} & \textds{llm} & \textds{perfect} & \textds{expert} & \textds{llm} & \textds{perfect} & \textds{expert} & \textds{llm} \\
\midrule
\multirow{4}{*}{\cell{l}{\textds{oracle}\\$m{=}7$}}
 & \CBM{}     & 83.9\%{\scriptsize$\pm$1.3} & 95.2\%{\scriptsize$\pm$1.2} & 89.8\%{\scriptsize$\pm$0.8} & 84.7\%{\scriptsize$\pm$0.3} & 96.6\%{\scriptsize$\pm$0.8} & 89.0\%{\scriptsize$\pm$0.3} & 85.1\%{\scriptsize$\pm$0.4} & \textbf{97.3\%}{\scriptsize$\boldsymbol{\pm}$\textbf{0.7}} & 90.6\%{\scriptsize$\pm$0.2} & \textbf{87.5\%}{\scriptsize$\boldsymbol{\pm}$\textbf{0.1}} \\
 & \CEM{}     & 86.4\%{\scriptsize$\pm$0.3} & 93.2\%{\scriptsize$\pm$0.7} & 89.0\%{\scriptsize$\pm$0.5} & 87.3\%{\scriptsize$\pm$0.2} & 93.3\%{\scriptsize$\pm$0.7} & 89.0\%{\scriptsize$\pm$0.4} & 87.3\%{\scriptsize$\pm$0.2} & 95.9\%{\scriptsize$\pm$0.6} & \textbf{90.9\%}{\scriptsize$\boldsymbol{\pm}$\textbf{0.2}} & \textbf{87.5\%}{\scriptsize$\boldsymbol{\pm}$\textbf{0.1}} \\
 & \ProbCBM{} & \textbf{87.5\%}{\scriptsize$\boldsymbol{\pm}$\textbf{0.1}} & 87.5\%{\scriptsize$\pm$0.1} & 87.5\%{\scriptsize$\pm$0.1} & 87.5\%{\scriptsize$\pm$0.1} & 87.2\%{\scriptsize$\pm$1.8} & 83.2\%{\scriptsize$\pm$2.1} & 85.7\%{\scriptsize$\pm$1.6} & 91.6\%{\scriptsize$\pm$2.9} & 84.9\%{\scriptsize$\pm$1.9} & 87.1\%{\scriptsize$\pm$0.4} \\
 & \ECBM{}    & 86.9\%{\scriptsize$\pm$0.6} & 83.4\%{\scriptsize$\pm$0.7} & 82.8\%{\scriptsize$\pm$0.8} & 82.5\%{\scriptsize$\pm$0.9} & 84.0\%{\scriptsize$\pm$0.8} & 83.2\%{\scriptsize$\pm$0.6} & 82.7\%{\scriptsize$\pm$0.8} & 93.7\%{\scriptsize$\pm$0.6} & 89.3\%{\scriptsize$\pm$0.2} & 87.3\%{\scriptsize$\pm$0.2} \\
\midrule
\multirow{4}{*}{\cell{l}{\textds{discovered}\\$m{=}12$}}
 & \CBM{}     & 80.8\%{\scriptsize$\pm$2.3} & 92.0\%{\scriptsize$\pm$1.5} & 88.4\%{\scriptsize$\pm$1.4} & 87.3\%{\scriptsize$\pm$0.2} & 93.0\%{\scriptsize$\pm$1.0} & 88.4\%{\scriptsize$\pm$0.8} & 87.2\%{\scriptsize$\pm$0.1} & \textbf{94.1\%}{\scriptsize$\boldsymbol{\pm}$\textbf{0.5}} & \textbf{88.8\%}{\scriptsize$\boldsymbol{\pm}$\textbf{0.5}} & \textbf{87.5\%}{\scriptsize$\boldsymbol{\pm}$\textbf{0.1}} \\
 & \CEM{}     & 86.7\%{\scriptsize$\pm$0.8} & 88.4\%{\scriptsize$\pm$2.3} & 86.2\%{\scriptsize$\pm$2.3} & 87.3\%{\scriptsize$\pm$0.2} & 87.8\%{\scriptsize$\pm$2.3} & 83.1\%{\scriptsize$\pm$2.8} & 87.3\%{\scriptsize$\pm$0.3} & 92.6\%{\scriptsize$\pm$1.9} & 88.7\%{\scriptsize$\pm$0.9} & \textbf{87.5\%}{\scriptsize$\boldsymbol{\pm}$\textbf{0.1}} \\
 & \ProbCBM{} & \textbf{87.5\%}{\scriptsize$\boldsymbol{\pm}$\textbf{0.1}} & 87.5\%{\scriptsize$\pm$0.1} & 87.5\%{\scriptsize$\pm$0.1} & 87.5\%{\scriptsize$\pm$0.1} & 82.0\%{\scriptsize$\pm$4.7} & 81.0\%{\scriptsize$\pm$4.0} & 81.0\%{\scriptsize$\pm$4.2} & 83.8\%{\scriptsize$\pm$3.0} & 77.7\%{\scriptsize$\pm$4.6} & 83.2\%{\scriptsize$\pm$2.5} \\
 & \ECBM{}    & \textbf{87.5\%}{\scriptsize$\boldsymbol{\pm}$\textbf{0.1}} & 87.2\%{\scriptsize$\pm$1.0} & 86.6\%{\scriptsize$\pm$0.8} & 86.5\%{\scriptsize$\pm$0.6} & 87.2\%{\scriptsize$\pm$1.0} & 86.4\%{\scriptsize$\pm$0.8} & 86.5\%{\scriptsize$\pm$0.6} & 88.8\%{\scriptsize$\pm$0.5} & 87.8\%{\scriptsize$\pm$0.1} & 87.5\%{\scriptsize$\pm$0.1} \\
\midrule
\multirow{4}{*}{\cell{l}{\textds{machine\_}\\\textds{labeled}\\$m{=}12$}}
 & \CBM{}     & \textbf{87.5\%}{\scriptsize$\boldsymbol{\pm}$\textbf{0.1}} & 62.6\%{\scriptsize$\pm$3.9} & 58.5\%{\scriptsize$\pm$2.6} & 61.4\%{\scriptsize$\pm$4.5} & 62.6\%{\scriptsize$\pm$3.9} & 54.4\%{\scriptsize$\pm$3.0} & 62.7\%{\scriptsize$\pm$4.5} & 72.1\%{\scriptsize$\pm$2.6} & 64.3\%{\scriptsize$\pm$1.6} & 73.1\%{\scriptsize$\pm$3.5} \\
 & \CEM{}     & 86.7\%{\scriptsize$\pm$0.4} & 86.6\%{\scriptsize$\pm$0.4} & 86.6\%{\scriptsize$\pm$0.4} & 86.7\%{\scriptsize$\pm$0.4} & 86.5\%{\scriptsize$\pm$0.4} & 86.5\%{\scriptsize$\pm$0.4} & 86.7\%{\scriptsize$\pm$0.3} & 86.7\%{\scriptsize$\pm$0.3} & 86.7\%{\scriptsize$\pm$0.3} & 86.8\%{\scriptsize$\pm$0.3} \\
 & \ProbCBM{} & 87.2\%{\scriptsize$\pm$0.1} & 87.2\%{\scriptsize$\pm$0.2} & 87.2\%{\scriptsize$\pm$0.2} & 87.3\%{\scriptsize$\pm$0.2} & 87.2\%{\scriptsize$\pm$0.2} & 87.2\%{\scriptsize$\pm$0.2} & 87.3\%{\scriptsize$\pm$0.1} & 87.3\%{\scriptsize$\pm$0.1} & 87.3\%{\scriptsize$\pm$0.1} & 87.3\%{\scriptsize$\pm$0.1} \\
 & \ECBM{}    & 87.2\%{\scriptsize$\pm$0.2} & \textbf{87.5\%}{\scriptsize$\boldsymbol{\pm}$\textbf{0.1}} & \textbf{87.5\%}{\scriptsize$\boldsymbol{\pm}$\textbf{0.1}} & \textbf{87.5\%}{\scriptsize$\boldsymbol{\pm}$\textbf{0.1}} & 87.5\%{\scriptsize$\pm$0.1} & 87.5\%{\scriptsize$\pm$0.1} & 87.5\%{\scriptsize$\pm$0.1} & 87.5\%{\scriptsize$\pm$0.1} & 87.5\%{\scriptsize$\pm$0.1} & 87.5\%{\scriptsize$\pm$0.1} \\
\midrule
\multirow{4}{*}{\cell{l}{\textds{llm\_written}\\$m{=}12$}}
 & \CBM{}     & \textbf{87.5\%}{\scriptsize$\boldsymbol{\pm}$\textbf{0.1}} & 57.6\%{\scriptsize$\pm$1.3} & 54.5\%{\scriptsize$\pm$0.6} & 38.0\%{\scriptsize$\pm$1.0} & 57.6\%{\scriptsize$\pm$1.3} & 52.3\%{\scriptsize$\pm$0.4} & 36.0\%{\scriptsize$\pm$1.0} & 57.6\%{\scriptsize$\pm$1.3} & 55.1\%{\scriptsize$\pm$0.9} & 36.2\%{\scriptsize$\pm$2.9} \\
 & \CEM{}     & 86.2\%{\scriptsize$\pm$0.7} & 86.0\%{\scriptsize$\pm$0.8} & 86.1\%{\scriptsize$\pm$0.7} & 86.1\%{\scriptsize$\pm$0.7} & 85.9\%{\scriptsize$\pm$0.8} & 86.0\%{\scriptsize$\pm$0.7} & 86.2\%{\scriptsize$\pm$0.6} & 85.9\%{\scriptsize$\pm$0.8} & 86.3\%{\scriptsize$\pm$0.6} & 86.4\%{\scriptsize$\pm$0.6} \\
 & \ProbCBM{} & 75.2\%{\scriptsize$\pm$5.3} & 81.9\%{\scriptsize$\pm$4.1} & 81.9\%{\scriptsize$\pm$4.1} & 82.0\%{\scriptsize$\pm$4.0} & 84.7\%{\scriptsize$\pm$1.4} & 84.6\%{\scriptsize$\pm$1.4} & 84.6\%{\scriptsize$\pm$1.2} & 86.5\%{\scriptsize$\pm$0.4} & 86.5\%{\scriptsize$\pm$0.4} & 86.4\%{\scriptsize$\pm$0.4} \\
 & \ECBM{}    & 87.3\%{\scriptsize$\pm$0.2} & \textbf{87.5\%}{\scriptsize$\boldsymbol{\pm}$\textbf{0.1}} & \textbf{87.5\%}{\scriptsize$\boldsymbol{\pm}$\textbf{0.1}} & \textbf{87.5\%}{\scriptsize$\boldsymbol{\pm}$\textbf{0.1}} & 87.5\%{\scriptsize$\pm$0.1} & 87.5\%{\scriptsize$\pm$0.1} & 87.5\%{\scriptsize$\pm$0.1} & 87.5\%{\scriptsize$\pm$0.1} & 87.5\%{\scriptsize$\pm$0.1} & 87.5\%{\scriptsize$\pm$0.1} \\
\midrule
\multirow{4}{*}{\cell{l}{\textds{clip\_words}\\$m{=}12$}}
 & \CBM{}     & \textbf{87.5\%}{\scriptsize$\boldsymbol{\pm}$\textbf{0.1}} & 62.5\%{\scriptsize$\pm$1.3} & 58.5\%{\scriptsize$\pm$0.9} & 42.6\%{\scriptsize$\pm$2.5} & 62.5\%{\scriptsize$\pm$1.3} & 55.7\%{\scriptsize$\pm$0.5} & 38.4\%{\scriptsize$\pm$1.9} & 62.5\%{\scriptsize$\pm$1.3} & 57.3\%{\scriptsize$\pm$0.7} & 42.2\%{\scriptsize$\pm$9.8} \\
 & \CEM{}     & 86.6\%{\scriptsize$\pm$0.4} & 86.5\%{\scriptsize$\pm$0.4} & 86.6\%{\scriptsize$\pm$0.4} & 86.6\%{\scriptsize$\pm$0.4} & 86.5\%{\scriptsize$\pm$0.4} & 86.6\%{\scriptsize$\pm$0.4} & 86.6\%{\scriptsize$\pm$0.4} & 86.6\%{\scriptsize$\pm$0.4} & 86.6\%{\scriptsize$\pm$0.4} & 86.7\%{\scriptsize$\pm$0.4} \\
 & \ProbCBM{} & 85.8\%{\scriptsize$\pm$0.9} & 86.7\%{\scriptsize$\pm$0.3} & 86.7\%{\scriptsize$\pm$0.3} & 86.8\%{\scriptsize$\pm$0.3} & 86.9\%{\scriptsize$\pm$0.1} & 86.9\%{\scriptsize$\pm$0.2} & 86.8\%{\scriptsize$\pm$0.3} & 87.2\%{\scriptsize$\pm$0.2} & 87.2\%{\scriptsize$\pm$0.2} & 87.2\%{\scriptsize$\pm$0.2} \\
 & \ECBM{}    & \textbf{87.5\%}{\scriptsize$\boldsymbol{\pm}$\textbf{0.1}} & 87.5\%{\scriptsize$\pm$0.1} & 87.5\%{\scriptsize$\pm$0.1} & 87.5\%{\scriptsize$\pm$0.1} & 87.5\%{\scriptsize$\pm$0.1} & 87.5\%{\scriptsize$\pm$0.1} & 87.5\%{\scriptsize$\pm$0.1} & 87.5\%{\scriptsize$\pm$0.1} & 87.5\%{\scriptsize$\pm$0.1} & 87.5\%{\scriptsize$\pm$0.1} \\
\bottomrule
\end{tabular}%
}
\caption{Accuracy of concept-based models in the decision-support benchmark under each concept set and intervention regime after correcting up to $1$, $3$, or all concepts per robot under the sparse rule (mean $\pm$ SE, $n{=}4$ seeds). With \captionds{discovered} concepts, every architecture is less accurate after all interventions than with \captionds{oracle} concepts.}
\label{Table::BigTableSparse}
\end{table*}

\begin{wrapfigure}{R}{0.5\linewidth}
    \centering
    \vspace{-1em}
    \resizebox{\linewidth}{!}{\begin{tikzpicture}
\newcommand{\dsCM}[2]{\fontencoding{OT1}\fontfamily{#1}\fontsize{#2}{#2}\selectfont}
\definecolor{dsBlue}{HTML}{3B6FB6}
\definecolor{dsCEM}{HTML}{3E8E5E}
\definecolor{dsECBM}{HTML}{8E5BA8}
\definecolor{dsProb}{HTML}{D9822B}
\definecolor{dsLabel}{HTML}{4D4D4D}
\definecolor{dsGrid}{HTML}{DFE4E9}
\begin{axis}[
  set layers, width=5.6cm, height=4.8cm, scale only axis, clip=false,
  xmin=-0.35, xmax=3.1, ymin=80, ymax=100,
  axis x line*=bottom, axis y line*=left,
  axis line style={-, black!55, line width=0.5pt},
  every tick/.style={black!55, line width=0.4pt}, tick align=inside, tick pos=left,
  xtick={0,1,2,3}, xticklabels={none,{$\leq$\,1},{$\leq$\,3},all},
  ytick={80,85,90,95,100}, yticklabel={\pgfmathprintnumber[assume math mode=true]{\tick}\%},
  ticklabel style={font=\dsCM{phv}{8}, text=black!75},
  label style={font=\dsCM{phv}{8}, text=dsLabel}, xlabel style={yshift=2pt}, ylabel style={yshift=-3.36pt},
  xlabel={Interventions}, ylabel={Accuracy},
  ymajorgrids, grid style={dsGrid, line width=0.4pt},
  mark size=1.8pt,
  dsLine/.style={#1, line width=1.1pt, mark options={solid, fill=#1}},
  dsBand/.style={draw=none, fill=#1!18, forget plot},
]
\addplot[dsBand=dsBlue] coordinates {(0,82.62)(1,94.03)(2,95.86)(3,96.68)(3,97.98)(2,97.38)(1,96.37)(0,85.17)} -- cycle;
\addplot[dsBand=dsCEM] coordinates {(0,86.09)(1,92.47)(2,92.6)(3,95.34)(3,96.5)(2,93.97)(1,93.84)(0,86.71)} -- cycle;
\addplot[dsBand=dsECBM] coordinates {(0,86.25)(1,82.69)(2,83.25)(3,93.05)(3,94.3)(2,84.76)(1,84.12)(0,87.45)} -- cycle;
\addplot[dsBand=dsProb] coordinates {(0,87.48)(1,87.49)(2,85.43)(3,88.66)(3,94.55)(2,89.0)(1,87.61)(0,87.61)} -- cycle;
\addplot[dsLine=dsBlue, mark=*] coordinates {(0,83.9)(1,95.2)(2,96.62)(3,97.33)};
\addplot[dsLine=dsCEM, mark=square*] coordinates {(0,86.4)(1,93.16)(2,93.28)(3,95.92)};
\addplot[dsLine=dsECBM, mark=diamond*] coordinates {(0,86.85)(1,83.41)(2,84.0)(3,93.68)};
\addplot[dsLine=dsProb, mark=triangle*] coordinates {(0,87.54)(1,87.55)(2,87.22)(3,91.6)};
\addplot[black, line width=1.1pt, dash pattern=on 3pt off 2.5pt, forget plot, on layer=axis foreground] coordinates {(-0.35,87.5) (3.1,87.5)};
\tikzset{dsEnd/.style={anchor=west, inner sep=0pt, font=\fontsize{8}{9}\fontfamily{phv}\selectfont}}
\node[dsEnd, text=dsBlue] at ([xshift=6pt]axis cs:3,97.33) {CBM};
\node[dsEnd, text=dsCEM] at ([xshift=6pt]axis cs:3,95.92) {CEM};
\node[dsEnd, text=dsECBM] at ([xshift=6pt]axis cs:3,93.68) {ECBM};
\node[dsEnd, text=dsProb] at ([xshift=6pt]axis cs:3,91.6) {ProbCBM};
\node[dsEnd, text=black!85] at ([xshift=3pt]axis cs:3.1,87.5) {majority class};
\end{axis}
\end{tikzpicture}%
}
    \caption{Accuracy of each architecture as the intervention budget grows, with perfect interventions on \captionds{oracle} concepts under the sparse rule (mean $\pm$ SE over $n{=}4$ seeds). The dashed line marks the majority-class rate. The \CBM{} and \CEM{} turn interventions into correct predictions, whereas \ProbCBM{} stays near the majority-class rate until every concept is corrected.}
    \label{Fig::ArchitectureResponseSparse}
\end{wrapfigure}

\textbf{Interventions work when the representation of an intervened concept is trained to predict the label.} The \CBM{} and \CEM{} start below the majority rate and reach high accuracy as interventions grow. The \CBM{} trains its front-end model on true concept values, so an intervened value is exactly the input it learned from. \CEM{} represents each concept by learned embeddings for ``present'' and ``absent'', but is trained with interventions on random concepts, so the label loss shapes these embeddings to separate the classes. Both architectures therefore benefit from interventions even on concepts their detectors cannot recognize. \ECBM{} is an exception. It predicts the label from the image until an intervention and from its concept predictions after, so its first interventions lower accuracy until enough concepts are corrected.

\textbf{Interventions do not help when that representation is trained only to detect the concept.} \ProbCBM{} predicts Glorp for every robot, and interventions rarely change this prediction. This is because \ProbCBM{} replaces an intervened concept with a vector for ``present'' or ``absent'' that is learned to match its concept detector. The detector, however, is at chance on two of the three concepts the label depends on, since both are difficult to recognize in the image. Their two vectors therefore end up nearly identical, and an intervention barely changes the input to the label predictor, or its prediction. An architecture can therefore predict the common class well and still give the expert no reliable way to catch the rare cases.

\section{Supporting Material for Automation}
\label{Appendix::AutomationResults}

In what follows, we report the quality of concept detection and the budget sweep behind our automation results (\cref{Sec::AutomationValidation}), and repeat the automation experiments under other settings. Every selective classifier is Platt-scaled on the validation split. We fit a logistic regression of the validation labels on the logit of the model's label probability and apply it to the test boards, so that a model abstains on a calibrated probability~\citep{joren2024classification}; the abstention threshold is fitted once on the same split, without checks, and kept for every budget.

\subsection{Quality of Concept Detection}
\label{Appendix::AutomationDetection}

\begin{table}[h!]
\centering
\small
\resizebox{0.85\textwidth}{!}{
\begin{tabular}{@{}l rr rr@{}}
& \multicolumn{2}{c}{\textbf{Digits (81 cells)}} & \multicolumn{2}{c}{\textbf{Concepts (27 constraints)}} \\
\cmidrule(lr){2-3} \cmidrule(lr){4-5}
\textbf{Resolution} & \textsf{Accuracy} & \textsf{Boards with an error} & \textsf{Accuracy} & \textsf{Boards with an error} \\
\midrule
\textds{50\,px} & 99.7\%{\scriptsize$\pm$0.0} & 19.1\%{\scriptsize$\pm$2.8} & 97.5\%{\scriptsize$\pm$0.4} & 25.4\%{\scriptsize$\pm$4.4} \\
\textds{10\,px} & 98.9\%{\scriptsize$\pm$0.1} & 57.5\%{\scriptsize$\pm$3.8} & 92.3\%{\scriptsize$\pm$1.0} & 58.7\%{\scriptsize$\pm$3.9} \\
\bottomrule
\end{tabular}

}
\caption{We report the accuracy with which the \CBM{} reads the digits in the $81$ cells of a test board and detects its $27$ constraint concepts, and the share of test boards with at least one wrong digit or concept. The detectors are accurate per cell and per concept, yet their errors reach a quarter of the boards at 50\,px and more than half at 10\,px.}
\label{Table::SudokuDetection}
\end{table}

For every seed and resolution, we score how well the \CBM{} reads the digits in the cells and detects the $27$ constraints of the test boards (see \cref{Table::SudokuDetection}). A wrong digit changes up to three constraints, so a digit error rate of $0.3\%$ at 50\,px and $1.1\%$ at 10\,px puts a wrong concept on a quarter and on more than half of the boards. A wrong concept can make a valid board appear invalid, which happens on $10\%$ of test boards at 50\,px and $28\%$ at 10\,px. The model then misclassifies the board unless it abstains. These errors are therefore a likely reason why no model in \cref{Table::sudoku_px} covers every board.

\subsection{Budget Sweep}
\label{Appendix::AutomationKSweep}

\begin{figure}[h!]
    \centering
    \resizebox{\linewidth}{!}{\begin{tikzpicture}
\def\ksFont#1{\fontencoding{T1}\fontfamily{DejaVuSans-TLF}\fontsize{#1}{#1}\selectfont}
\definecolor{ksVal}{rgb}{0.3019607843,0.3019607843,0.3019607843}
\definecolor{ksTest}{rgb}{0.231372549,0.4352941176,0.7137254902}
\definecolor{ksGrid}{rgb}{0.6901960784,0.6901960784,0.6901960784}
\definecolor{ksLegendEdge}{rgb}{0.8,0.8,0.8}
\def\ksAxisWidth{276.4504545455pt}\def\ksAxisHeight{175.96pt}\def\ksAxisBottom{41.96pt}
\def\ksTau{[scale=0.012] (362,100) .. controls (374,84) and (397,76) .. (433,76) -- (476,76) -- (476,0) -- (422,0) .. controls (363,0) and (320,15) .. (294,47) .. controls (267,79) and (254,129) .. (254,198) -- (254,457) -- (49,457) -- (49,547) -- (553,547) -- (553,457) -- (346,457) -- (346,193) .. controls (346,145) and (351,114) .. (362,100) -- cycle}
\pgfplotstableread{
k vm50   vlo50  vhi50  tm50   tlo50  thi50  vm10   vlo10  vhi10  tm10   tlo10  thi10
0 78.5   72.645 84.355 79.75  73.865 85.635 58.0   54.267 61.733 58.75  54.93  62.57
1 79.954 73.698 86.21  81.0   74.793 87.207 62.094 58.465 65.723 63.172 59.161 67.184
2 80.791 74.251 87.33  81.624 75.156 88.092 63.948 60.769 67.127 65.515 61.589 69.441
3 81.285 74.581 87.99  81.896 75.267 88.526 65.17  62.023 68.318 66.593 62.784 70.401
4 81.283 74.027 88.539 81.785 74.822 88.748 67.039 64.001 70.077 67.957 64.585 71.33
5 80.304 72.322 88.285 80.843 73.235 88.451 67.42  64.396 70.445 68.313 65.019 71.607
6 78.696 69.559 87.834 79.265 70.539 87.991 65.878 62.645 69.11  67.35  63.752 70.948
7 73.907 61.482 86.333 74.456 62.569 86.342 64.293 61.349 67.236 66.583 63.768 69.399
}\ksData
\tikzset{
  ksBand/.style={draw=#1, fill=#1, opacity=0.12, line width=1pt, line join=round},
  ksMean/.style={#1, line width=1.5pt, line join=round, mark=*, mark size=1.75pt,
    mark options={solid, fill=#1, line width=1pt}},
  ksValLine/.style={ksMean=ksVal, dash pattern=on 5.55pt off 2.4pt},
  ksTestLine/.style={ksMean=ksTest, line cap=rect},
}
\def\ksPanel#1#2#3#4{%
\begin{axis}[
    scale only axis, width=\ksAxisWidth, height=\ksAxisHeight, anchor=south west, at={(#1 pt,\ksAxisBottom)},
    xmin=-0.35, xmax=7.35, ymin=47.5, ymax=92,
    xtick={0,...,7}, xticklabels={none,{$\leq$\,1},{$\leq$\,2},{$\leq$\,3},{$\leq$\,5},{$\leq$\,8},{$\leq$\,13},all}, ytick={50,60,...,90}, #4,
    xtick pos=bottom, ytick pos=left, tick align=outside, major tick length=3.5pt,
    every tick/.style={black, line width=0.8pt},
    separate axis lines, axis line style={line width=0.8pt, line cap=rect},
    grid=major, fill between/on layer=axis background, major grid style={ksGrid, opacity=0.3, line width=0.8pt, line cap=rect},
    tick label style={font=\ksFont{10}},
    xticklabel style={anchor=base, yshift=-11.25pt},
    yticklabel style={anchor=base east, xshift=-0.17pt, yshift=-3.35pt},
  ]
  \pgfplotsinvokeforeach{vlo,vhi,tlo,thi}{\addplot[draw=none, name path=##1] table[x=k, y=##1#2]{\ksData};}
  \addplot[ksBand=ksVal] fill between[of=vlo and vhi];
  \addplot[ksBand=ksTest] fill between[of=tlo and thi];
  \addplot[ksValLine] table[x=k, y=vm#2]{\ksData};
  \addplot[ksTestLine] table[x=k, y=tm#2]{\ksData};
  \pgfplotstablegetelem{#3}{vm#2}\of\ksData \let\ksValStar\pgfplotsretval
  \pgfplotstablegetelem{#3}{tm#2}\of\ksData \let\ksTestStar\pgfplotsretval
  \edef\ksStarLabels{
    \noexpand\node[ksVal, anchor=base, inner sep=0, yshift=-14pt, font=\noexpand\ksFont{8}] at (axis cs:#3,\ksValStar) {best};
    \noexpand\node[ksTest, anchor=base, inner sep=0, yshift=9pt, font=\noexpand\ksFont{8}] at (axis cs:#3,\ksTestStar) {best};
  }
  \ksStarLabels
\end{axis}
\begin{scope}[x=1pt, y=1pt, xshift={(#1-54.02)*1pt}]
  \node[anchor=base west, inner sep=0, font=\ksFont{12}] at (128.4014772727,223.92) {CBM, #2{} px,};
  \node[anchor=base west, inner sep=0, font=\ksFont{12}] at (204.4268678977,223.92) {$\alpha_{\textrm{target}} = 0.99$};
  \node[anchor=base, inner sep=0, font=\ksFont{10}] at (192.2452272727,13.694375) {Interventions};
\end{scope}}
\ksPanel{54.02}{50}{3}{yticklabels={50\%,60\%,70\%,80\%,90\%}}
\ksPanel{345.56}{10}{5}{yticklabels={}}
\begin{scope}[x=1pt, y=1pt]
\node[anchor=base, inner sep=0, rotate=90, font=\ksFont{10}] at (18.723125,129.94) {WorkSaved};
\filldraw[fill=white, draw=ksLegendEdge, opacity=0.8, line width=1pt, rounded corners=2pt] (59.02,46.96) rectangle (144.27,79.30375);
\foreach \ksY/\ksStyle/\ksLabel in {71.21/ksValLine/validation, 56.538125/ksTestLine/test} {
  \draw[\ksStyle, mark phase=2, mark repeat=3] plot coordinates {(63.02,\ksY) (73.02,\ksY) (83.02,\ksY)};
  \node[anchor=base west, inner sep=0, font=\ksFont{10}] at (91.02,\ksY-3.5) {\ksLabel};
}
\path (0,0) rectangle (633.6,244.8);
\end{scope}
\end{tikzpicture}%
}
    \caption{We report \captionwrk{} of the \CBM{} at every intervention budget, on the validation and the test split, at $\alpha_{\textrm{target}}{=}0.99$ (mean $\pm$ SE over ten seeds; ``best'' marks the budget that maximizes the mean). At 50\,px, \captionwrk{} peaks at three interventions per deferred board and falls $5$ points below no interventions with every concept verified. At 10\,px, \captionwrk{} peaks at eight interventions and stays above no interventions. The validation curve follows the test curve, so the budget can be chosen on the validation split.}
    \label{Fig::SudokuKSweep}
\end{figure}

We evaluate the \CBM{} at every budget $b \in \{0, 1, 2, 3, 5, 8, 13, 27\}$ on the validation and the test split (see \cref{Fig::SudokuKSweep}). The \CBM{} settles the most boards with its first interventions. It settles $19\%$ of the deferred boards at 50\,px and $15\%$ at 10\,px with one intervention, $35\%$ and $29\%$ with three, and $61\%$ and $72\%$ with every concept verified. A practitioner who picks the budget that maximizes \wrk{} on the validation split loses on average $0.5$ points (50\,px) and $0.6$ points (10\,px) of \wrk{} on the test split relative to the best test budget. The gain of that budget over no interventions is $3.1$ points at 50\,px and $11.9$ points at 10\,px.

\subsection{Deferred Boards With Every Concept Verified}
\label{Appendix::AutomationCeiling}
To find why further interventions save little work, we set every concept of each test board to its true value and score the board with the same scaling and abstention threshold as in \cref{Table::sudoku_px}. A board that remains in the abstention band cannot be labeled at the target with any number of interventions. We report the share of deferred boards that remain deferred and the label probability of invalid boards by the number of violated constraints in \cref{Table::SudokuCeiling}. An invalid board violates at least two constraints. The front-end models of the \CBM{}, \ProbCBM{}, and \ECBM{} assign a probability of $0.43$ to $0.97$ of being valid to invalid boards with two to four violated constraints, so a third to a half of their deferred boards remain deferred. The front-end model of the \CEM{} assigns lower probabilities to these boards, and only $12\%$ of its deferred boards remain deferred at 50\,px.

\begin{table}[h!]
\centering
\small
\begin{tabular}{@{}l rr rrr@{}}
& \multicolumn{2}{c}{\textbf{Remain deferred}} & \multicolumn{3}{c}{\textbf{Probability of a valid board}} \\
\cmidrule(lr){2-3} \cmidrule(lr){4-6}
\textbf{Model} & \textds{50\,px} & \textds{10\,px} & 2 violated & 3 violated & 4 violated \\
\midrule
\CBM{}     & 41.6\%{\scriptsize$\pm$7.7} & 29.2\%{\scriptsize$\pm$1.5} & 0.89{\scriptsize$\pm$0.02} & 0.77{\scriptsize$\pm$0.04} & 0.54{\scriptsize$\pm$0.07} \\
\CEM{}     & 11.6\%{\scriptsize$\pm$4.0} & 31.5\%{\scriptsize$\pm$9.8} & 0.67{\scriptsize$\pm$0.07} & 0.44{\scriptsize$\pm$0.09} & 0.14{\scriptsize$\pm$0.05} \\
\ProbCBM{} & 48.6\%{\scriptsize$\pm$7.2} & 51.3\%{\scriptsize$\pm$8.4} & 0.66{\scriptsize$\pm$0.04} & 0.58{\scriptsize$\pm$0.05} & 0.43{\scriptsize$\pm$0.08} \\
\ECBM{}    & 38.2\%{\scriptsize$\pm$14.0} & 34.5\%{\scriptsize$\pm$12.9} & 0.96{\scriptsize$\pm$0.04} & 0.97{\scriptsize$\pm$0.03} & 0.89{\scriptsize$\pm$0.05} \\
\bottomrule
\end{tabular}

\caption{We report the share of deferred test boards that remain deferred with every concept set to its true value (left) and the label probability that each front-end model assigns to invalid boards with two, three, or four violated constraints (right), at $\alpha_{\textrm{target}}{=}0.99$ (mean $\pm$ SE over seeds with a threshold). Every front-end model except the \CEM{}'s remains uncertain about invalid boards with a few violated constraints.}
\label{Table::SudokuCeiling}
\end{table}

\subsection{Sensitivity to the Accuracy Target and the Cost of an Intervention}
\label{Appendix::AutomationSensitivity}

We repeat the automation experiments at other selective-accuracy targets $\alpha_{\textrm{target}}$ and with other costs of an intervention.

\paragraph{Accuracy target}
We report the results in \cref{Table::TauSweep}. At every target and every budget, every model automates less work at 10\,px than at 50\,px, and the gap grows with the target. Without interventions, the \CEM{} automates significantly less work at 10\,px at every target, and the \CBM{} and the \ProbCBM{} at $0.95$ and $0.99$. The loss of the \ECBM{} is not significant at any target. At $\alpha_{\textrm{target}}{=}0.90$ and $0.95$, every model automates at least as much work with every concept verified as with three. Only at $0.99$ do some models automate less work with more interventions (see \cref{Sec::AutomationValidation}).

\begin{table}[h!]
\centering
\small
\resizebox{\textwidth}{!}{
\begin{tabular}{@{}lll rrrr rrrr@{}}
& & & \multicolumn{8}{c}{\textbf{Interventions}} \\
\cmidrule(lr){4-11}
& & & \multicolumn{4}{c}{$\wrk$} & \multicolumn{4}{c}{$\cov$} \\
\cmidrule(lr){4-7} \cmidrule(lr){8-11}
\textbf{$\alpha_{\textrm{target}}$} & \textbf{Resolution} & \textbf{Model} & \multicolumn{1}{c}{none} & \multicolumn{1}{c}{$\leq 1$ concept} & \multicolumn{1}{c}{$\leq 3$ concepts} & \multicolumn{1}{c}{all concepts} & \multicolumn{1}{c}{none} & \multicolumn{1}{c}{$\leq 1$ concept} & \multicolumn{1}{c}{$\leq 3$ concepts} & \multicolumn{1}{c}{all concepts} \\
\midrule
\multirow{8}{*}{$0.90$} & \multirow{4}{*}{\textds{50\,px}} & \CBM{}     & 99.4\%{\scriptsize$\pm$0.2} & 99.9\%{\scriptsize$\pm$0.1} & 99.9\%{\scriptsize$\pm$0.1} & 99.9\%{\scriptsize$\pm$0.0} & 99.4\%{\scriptsize$\pm$0.2} & 99.9\%{\scriptsize$\pm$0.1} & 100.0\%{\scriptsize$\pm$0.0} & 100.0\%{\scriptsize$\pm$0.0} \\
 &  & \CEM{}     & 99.2\%{\scriptsize$\pm$0.3} & 100.0\%{\scriptsize$\pm$0.0} & 100.0\%{\scriptsize$\pm$0.0} & 100.0\%{\scriptsize$\pm$0.0} & 99.2\%{\scriptsize$\pm$0.3} & 100.0\%{\scriptsize$\pm$0.0} & 100.0\%{\scriptsize$\pm$0.0} & 100.0\%{\scriptsize$\pm$0.0} \\
 &  & \ProbCBM{} & 98.2\%{\scriptsize$\pm$1.0} & 99.0\%{\scriptsize$\pm$0.8} & 99.4\%{\scriptsize$\pm$0.6} & 99.4\%{\scriptsize$\pm$0.6} & 98.2\%{\scriptsize$\pm$1.0} & 99.0\%{\scriptsize$\pm$0.7} & 99.5\%{\scriptsize$\pm$0.5} & 99.8\%{\scriptsize$\pm$0.2} \\
 &  & \ECBM{}    & 98.8\%{\scriptsize$\pm$0.6} & 99.8\%{\scriptsize$\pm$0.2} & 99.8\%{\scriptsize$\pm$0.2} & 99.8\%{\scriptsize$\pm$0.2} & 98.8\%{\scriptsize$\pm$0.6} & 99.8\%{\scriptsize$\pm$0.2} & 99.8\%{\scriptsize$\pm$0.2} & 99.8\%{\scriptsize$\pm$0.2} \\
 & \multirow{4}{*}{\textds{10\,px}} & \CBM{}     & 94.5\%{\scriptsize$\pm$3.6} & 97.0\%{\scriptsize$\pm$2.3} & 98.5\%{\scriptsize$\pm$1.3} & 98.7\%{\scriptsize$\pm$1.0} & 94.5\%{\scriptsize$\pm$3.6} & 97.2\%{\scriptsize$\pm$2.2} & 98.8\%{\scriptsize$\pm$1.0} & 99.5\%{\scriptsize$\pm$0.4} \\
 &  & \CEM{}     & 89.7\%{\scriptsize$\pm$1.9} & 93.2\%{\scriptsize$\pm$1.5} & 96.5\%{\scriptsize$\pm$1.2} & 98.9\%{\scriptsize$\pm$0.3} & 89.7\%{\scriptsize$\pm$1.9} & 93.5\%{\scriptsize$\pm$1.4} & 97.3\%{\scriptsize$\pm$1.1} & 100.0\%{\scriptsize$\pm$0.0} \\
 &  & \ProbCBM{} & 82.2\%{\scriptsize$\pm$8.1} & 84.5\%{\scriptsize$\pm$8.4} & 86.5\%{\scriptsize$\pm$8.7} & 86.9\%{\scriptsize$\pm$10.3} & 82.2\%{\scriptsize$\pm$8.1} & 85.2\%{\scriptsize$\pm$8.1} & 88.2\%{\scriptsize$\pm$7.9} & 95.0\%{\scriptsize$\pm$4.3} \\
 &  & \ECBM{}    & 88.2\%{\scriptsize$\pm$7.1} & 97.3\%{\scriptsize$\pm$1.5} & 98.2\%{\scriptsize$\pm$0.8} & 98.4\%{\scriptsize$\pm$0.7} & 88.2\%{\scriptsize$\pm$7.1} & 97.7\%{\scriptsize$\pm$1.3} & 98.7\%{\scriptsize$\pm$0.7} & 99.0\%{\scriptsize$\pm$0.7} \\
\midrule
\multirow{8}{*}{$0.95$} & \multirow{4}{*}{\textds{50\,px}} & \CBM{}     & 94.8\%{\scriptsize$\pm$1.6} & 97.1\%{\scriptsize$\pm$1.4} & 97.3\%{\scriptsize$\pm$1.5} & 97.6\%{\scriptsize$\pm$1.2} & 94.8\%{\scriptsize$\pm$1.6} & 97.2\%{\scriptsize$\pm$1.3} & 97.7\%{\scriptsize$\pm$1.3} & 99.7\%{\scriptsize$\pm$0.1} \\
 &  & \CEM{}     & 91.8\%{\scriptsize$\pm$1.4} & 94.9\%{\scriptsize$\pm$1.3} & 96.5\%{\scriptsize$\pm$1.2} & 98.6\%{\scriptsize$\pm$0.6} & 91.8\%{\scriptsize$\pm$1.4} & 95.2\%{\scriptsize$\pm$1.3} & 97.2\%{\scriptsize$\pm$1.1} & 99.8\%{\scriptsize$\pm$0.1} \\
 &  & \ProbCBM{} & 91.8\%{\scriptsize$\pm$1.7} & 93.7\%{\scriptsize$\pm$1.8} & 95.0\%{\scriptsize$\pm$1.7} & 96.0\%{\scriptsize$\pm$1.9} & 91.8\%{\scriptsize$\pm$1.7} & 94.0\%{\scriptsize$\pm$1.7} & 95.7\%{\scriptsize$\pm$1.6} & 98.7\%{\scriptsize$\pm$0.8} \\
 &  & \ECBM{}    & 97.5\%{\scriptsize$\pm$1.0} & 99.7\%{\scriptsize$\pm$0.2} & 99.7\%{\scriptsize$\pm$0.2} & 99.7\%{\scriptsize$\pm$0.2} & 97.5\%{\scriptsize$\pm$1.0} & 99.8\%{\scriptsize$\pm$0.2} & 99.8\%{\scriptsize$\pm$0.2} & 99.8\%{\scriptsize$\pm$0.2} \\
 & \multirow{4}{*}{\textds{10\,px}} & \CBM{}     & 77.4\%{\scriptsize$\pm$4.7} & 81.0\%{\scriptsize$\pm$4.4} & 86.1\%{\scriptsize$\pm$3.3} & 86.2\%{\scriptsize$\pm$2.6} & 77.4\%{\scriptsize$\pm$4.7} & 81.8\%{\scriptsize$\pm$4.2} & 88.2\%{\scriptsize$\pm$2.9} & 95.2\%{\scriptsize$\pm$0.7} \\
 &  & \CEM{}     & 58.1\%{\scriptsize$\pm$10.0} & 61.0\%{\scriptsize$\pm$10.2} & 66.9\%{\scriptsize$\pm$10.3} & 74.4\%{\scriptsize$\pm$11.0} & 58.1\%{\scriptsize$\pm$10.0} & 61.9\%{\scriptsize$\pm$10.2} & 69.2\%{\scriptsize$\pm$10.3} & 80.7\%{\scriptsize$\pm$10.7} \\
 &  & \ProbCBM{} & 64.0\%{\scriptsize$\pm$7.7} & 66.7\%{\scriptsize$\pm$8.2} & 69.2\%{\scriptsize$\pm$8.8} & 72.1\%{\scriptsize$\pm$11.0} & 64.0\%{\scriptsize$\pm$7.7} & 67.9\%{\scriptsize$\pm$7.9} & 72.5\%{\scriptsize$\pm$8.0} & 87.8\%{\scriptsize$\pm$5.5} \\
 &  & \ECBM{}    & 84.3\%{\scriptsize$\pm$7.5} & 95.9\%{\scriptsize$\pm$2.0} & 96.9\%{\scriptsize$\pm$1.5} & 97.1\%{\scriptsize$\pm$1.4} & 84.3\%{\scriptsize$\pm$7.5} & 96.4\%{\scriptsize$\pm$1.8} & 97.5\%{\scriptsize$\pm$1.4} & 97.8\%{\scriptsize$\pm$1.4} \\
\midrule
\multirow{8}{*}{$0.99$} & \multirow{4}{*}{\textds{50\,px}} & \CBM{}     & 79.8\%{\scriptsize$\pm$5.9} & 81.0\%{\scriptsize$\pm$6.2} & 81.9\%{\scriptsize$\pm$6.6} & 74.5\%{\scriptsize$\pm$11.9} & 79.8\%{\scriptsize$\pm$5.9} & 81.8\%{\scriptsize$\pm$6.0} & 84.0\%{\scriptsize$\pm$6.0} & 88.5\%{\scriptsize$\pm$6.0} \\
 &  & \CEM{}     & 72.5\%{\scriptsize$\pm$9.5} & 73.5\%{\scriptsize$\pm$9.7} & 80.5\%{\scriptsize$\pm$9.3} & 86.1\%{\scriptsize$\pm$9.6} & 72.5\%{\scriptsize$\pm$9.5} & 74.2\%{\scriptsize$\pm$9.7} & 82.0\%{\scriptsize$\pm$9.3} & 89.5\%{\scriptsize$\pm$10.0} \\
 &  & \ProbCBM{} & 79.5\%{\scriptsize$\pm$2.6} & 80.2\%{\scriptsize$\pm$2.3} & 80.3\%{\scriptsize$\pm$2.4} & 80.5\%{\scriptsize$\pm$2.7} & 79.5\%{\scriptsize$\pm$2.6} & 81.0\%{\scriptsize$\pm$2.2} & 82.5\%{\scriptsize$\pm$2.1} & 92.9\%{\scriptsize$\pm$1.1} \\
 &  & \ECBM{}    & 53.1\%{\scriptsize$\pm$11.8} & 66.7\%{\scriptsize$\pm$12.3} & 67.1\%{\scriptsize$\pm$12.5} & 58.0\%{\scriptsize$\pm$13.8} & 53.1\%{\scriptsize$\pm$11.8} & 67.7\%{\scriptsize$\pm$12.4} & 68.9\%{\scriptsize$\pm$12.5} & 69.3\%{\scriptsize$\pm$12.5} \\
 & \multirow{4}{*}{\textds{10\,px}} & \CBM{}     & 58.8\%{\scriptsize$\pm$3.8} & 63.2\%{\scriptsize$\pm$4.0} & 66.6\%{\scriptsize$\pm$3.8} & 66.6\%{\scriptsize$\pm$2.8} & 58.8\%{\scriptsize$\pm$3.8} & 64.7\%{\scriptsize$\pm$3.9} & 70.6\%{\scriptsize$\pm$3.4} & 88.5\%{\scriptsize$\pm$1.0} \\
 &  & \CEM{}     & 36.4\%{\scriptsize$\pm$8.6} & 38.3\%{\scriptsize$\pm$9.4} & 41.6\%{\scriptsize$\pm$10.6} & 53.4\%{\scriptsize$\pm$11.8} & 36.4\%{\scriptsize$\pm$8.6} & 39.5\%{\scriptsize$\pm$9.5} & 44.9\%{\scriptsize$\pm$10.8} & 64.7\%{\scriptsize$\pm$12.6} \\
 &  & \ProbCBM{} & 39.6\%{\scriptsize$\pm$6.9} & 40.7\%{\scriptsize$\pm$7.6} & 40.5\%{\scriptsize$\pm$8.4} & 26.5\%{\scriptsize$\pm$14.7} & 39.6\%{\scriptsize$\pm$6.9} & 42.8\%{\scriptsize$\pm$7.4} & 46.5\%{\scriptsize$\pm$7.6} & 66.3\%{\scriptsize$\pm$7.4} \\
 &  & \ECBM{}    & 34.5\%{\scriptsize$\pm$11.1} & 52.7\%{\scriptsize$\pm$11.1} & 54.0\%{\scriptsize$\pm$11.9} & 43.4\%{\scriptsize$\pm$15.0} & 34.5\%{\scriptsize$\pm$11.1} & 54.0\%{\scriptsize$\pm$11.2} & 56.6\%{\scriptsize$\pm$11.9} & 58.5\%{\scriptsize$\pm$12.3} \\
\bottomrule
\end{tabular}

}
\caption{We report \captionwrk{} and \captioncov{} at three selective-accuracy targets $\alpha_{\textrm{target}}$. When no threshold reaches the target on the validation split, the model defers every board and we count its coverage as zero (at $\alpha_{\textrm{target}}{=}0.95$, \CEM{} on one seed at 10\,px; at $0.99$, \CEM{} on one seed at 50\,px and two at 10\,px, and \ECBM{} on two seeds at each resolution). Every model automates less work at 10\,px than at 50\,px, at every target.}
\label{Table::TauSweep}
\end{table}

\paragraph{Cost of an intervention}
We consider three costs of an intervention: every intervention costs the same (equal), verifying a block costs twice as much as verifying a row or column ($2\times$ block), and verifying a row, column, and block costs $1.34$, $1.26$, and $0.92$, which we draw at random (random). The cost of an intervention changes which concept a model verifies first (see \cref{Appendix::InterventionPolicy}). We report the results in \cref{Table::CostModels}. With up to three concepts verified per deferred board, every model automates about as much work under every cost, within $3$ points of the equal cost. With every concept verified, every model automates less work under the $2\times$ block cost than under the equal cost, because \wrk{} subtracts the cost of every verified concept. Coverage differs by at most $2$ points across costs, and the findings of \cref{Sec::AutomationValidation} hold under every cost.

\begin{table}[h!]
\centering
\small
\resizebox{\textwidth}{!}{
\begin{tabular}{@{}lll rrrr rrrr@{}}
& & & \multicolumn{8}{c}{\textbf{Interventions}} \\
\cmidrule(lr){4-11}
& & & \multicolumn{4}{c}{$\wrk$} & \multicolumn{4}{c}{$\cov$} \\
\cmidrule(lr){4-7} \cmidrule(lr){8-11}
\textbf{Cost} & \textbf{Resolution} & \textbf{Model} & \multicolumn{1}{c}{none} & \multicolumn{1}{c}{$\leq 1$ concept} & \multicolumn{1}{c}{$\leq 3$ concepts} & \multicolumn{1}{c}{all concepts} & \multicolumn{1}{c}{none} & \multicolumn{1}{c}{$\leq 1$ concept} & \multicolumn{1}{c}{$\leq 3$ concepts} & \multicolumn{1}{c}{all concepts} \\
\midrule
\multirow{8}{*}{equal} & \multirow{4}{*}{\textds{50\,px}} & \CBM{}     & 79.8\%{\scriptsize$\pm$5.9} & 81.0\%{\scriptsize$\pm$6.2} & 81.9\%{\scriptsize$\pm$6.6} & 74.5\%{\scriptsize$\pm$11.9} & 79.8\%{\scriptsize$\pm$5.9} & 81.8\%{\scriptsize$\pm$6.0} & 84.0\%{\scriptsize$\pm$6.0} & 88.5\%{\scriptsize$\pm$6.0} \\
 &  & \CEM{}     & 72.5\%{\scriptsize$\pm$9.5} & 73.5\%{\scriptsize$\pm$9.7} & 80.5\%{\scriptsize$\pm$9.3} & 86.1\%{\scriptsize$\pm$9.6} & 72.5\%{\scriptsize$\pm$9.5} & 74.2\%{\scriptsize$\pm$9.7} & 82.0\%{\scriptsize$\pm$9.3} & 89.5\%{\scriptsize$\pm$10.0} \\
 &  & \ProbCBM{} & 79.5\%{\scriptsize$\pm$2.6} & 80.2\%{\scriptsize$\pm$2.3} & 80.3\%{\scriptsize$\pm$2.4} & 80.5\%{\scriptsize$\pm$2.7} & 79.5\%{\scriptsize$\pm$2.6} & 81.0\%{\scriptsize$\pm$2.2} & 82.5\%{\scriptsize$\pm$2.1} & 92.9\%{\scriptsize$\pm$1.1} \\
 &  & \ECBM{}    & 53.1\%{\scriptsize$\pm$11.8} & 66.7\%{\scriptsize$\pm$12.3} & 67.1\%{\scriptsize$\pm$12.5} & 58.0\%{\scriptsize$\pm$13.8} & 53.1\%{\scriptsize$\pm$11.8} & 67.7\%{\scriptsize$\pm$12.4} & 68.9\%{\scriptsize$\pm$12.5} & 69.3\%{\scriptsize$\pm$12.5} \\
 & \multirow{4}{*}{\textds{10\,px}} & \CBM{}     & 58.8\%{\scriptsize$\pm$3.8} & 63.2\%{\scriptsize$\pm$4.0} & 66.6\%{\scriptsize$\pm$3.8} & 66.6\%{\scriptsize$\pm$2.8} & 58.8\%{\scriptsize$\pm$3.8} & 64.7\%{\scriptsize$\pm$3.9} & 70.6\%{\scriptsize$\pm$3.4} & 88.5\%{\scriptsize$\pm$1.0} \\
 &  & \CEM{}     & 36.4\%{\scriptsize$\pm$8.6} & 38.3\%{\scriptsize$\pm$9.4} & 41.6\%{\scriptsize$\pm$10.6} & 53.4\%{\scriptsize$\pm$11.8} & 36.4\%{\scriptsize$\pm$8.6} & 39.5\%{\scriptsize$\pm$9.5} & 44.9\%{\scriptsize$\pm$10.8} & 64.7\%{\scriptsize$\pm$12.6} \\
 &  & \ProbCBM{} & 39.6\%{\scriptsize$\pm$6.9} & 40.7\%{\scriptsize$\pm$7.6} & 40.5\%{\scriptsize$\pm$8.4} & 26.5\%{\scriptsize$\pm$14.7} & 39.6\%{\scriptsize$\pm$6.9} & 42.8\%{\scriptsize$\pm$7.4} & 46.5\%{\scriptsize$\pm$7.6} & 66.3\%{\scriptsize$\pm$7.4} \\
 &  & \ECBM{}    & 34.5\%{\scriptsize$\pm$11.1} & 52.7\%{\scriptsize$\pm$11.1} & 54.0\%{\scriptsize$\pm$11.9} & 43.4\%{\scriptsize$\pm$15.0} & 34.5\%{\scriptsize$\pm$11.1} & 54.0\%{\scriptsize$\pm$11.2} & 56.6\%{\scriptsize$\pm$11.9} & 58.5\%{\scriptsize$\pm$12.3} \\
\midrule
\multirow{8}{*}{2$\times$ block} & \multirow{4}{*}{\textds{50\,px}} & \CBM{}     & 79.8\%{\scriptsize$\pm$5.9} & 81.0\%{\scriptsize$\pm$6.2} & 81.2\%{\scriptsize$\pm$6.6} & 69.9\%{\scriptsize$\pm$13.8} & 79.8\%{\scriptsize$\pm$5.9} & 81.8\%{\scriptsize$\pm$6.0} & 83.3\%{\scriptsize$\pm$5.9} & 88.5\%{\scriptsize$\pm$6.0} \\
 &  & \CEM{}     & 72.5\%{\scriptsize$\pm$9.5} & 73.9\%{\scriptsize$\pm$9.7} & 81.3\%{\scriptsize$\pm$9.3} & 86.0\%{\scriptsize$\pm$9.6} & 72.5\%{\scriptsize$\pm$9.5} & 74.6\%{\scriptsize$\pm$9.7} & 82.9\%{\scriptsize$\pm$9.4} & 89.5\%{\scriptsize$\pm$10.0} \\
 &  & \ProbCBM{} & 79.5\%{\scriptsize$\pm$2.6} & 79.3\%{\scriptsize$\pm$2.1} & 78.6\%{\scriptsize$\pm$2.3} & 73.3\%{\scriptsize$\pm$3.4} & 79.5\%{\scriptsize$\pm$2.6} & 80.3\%{\scriptsize$\pm$2.0} & 81.5\%{\scriptsize$\pm$2.0} & 91.1\%{\scriptsize$\pm$1.4} \\
 &  & \ECBM{}    & 53.1\%{\scriptsize$\pm$11.8} & 66.6\%{\scriptsize$\pm$12.3} & 66.7\%{\scriptsize$\pm$12.5} & 54.4\%{\scriptsize$\pm$14.7} & 53.1\%{\scriptsize$\pm$11.8} & 67.8\%{\scriptsize$\pm$12.4} & 68.9\%{\scriptsize$\pm$12.5} & 69.2\%{\scriptsize$\pm$12.5} \\
 & \multirow{4}{*}{\textds{10\,px}} & \CBM{}     & 58.8\%{\scriptsize$\pm$3.8} & 63.1\%{\scriptsize$\pm$4.0} & 65.2\%{\scriptsize$\pm$3.9} & 60.0\%{\scriptsize$\pm$3.4} & 58.8\%{\scriptsize$\pm$3.8} & 64.7\%{\scriptsize$\pm$3.9} & 69.3\%{\scriptsize$\pm$3.5} & 88.3\%{\scriptsize$\pm$1.0} \\
 &  & \CEM{}     & 36.4\%{\scriptsize$\pm$8.6} & 37.4\%{\scriptsize$\pm$9.3} & 40.7\%{\scriptsize$\pm$10.4} & 51.1\%{\scriptsize$\pm$11.9} & 36.4\%{\scriptsize$\pm$8.6} & 38.9\%{\scriptsize$\pm$9.3} & 44.4\%{\scriptsize$\pm$10.6} & 64.5\%{\scriptsize$\pm$12.6} \\
 &  & \ProbCBM{} & 39.6\%{\scriptsize$\pm$6.9} & 39.6\%{\scriptsize$\pm$7.3} & 38.4\%{\scriptsize$\pm$8.2} & 13.9\%{\scriptsize$\pm$16.4} & 39.6\%{\scriptsize$\pm$6.9} & 42.4\%{\scriptsize$\pm$7.0} & 46.2\%{\scriptsize$\pm$7.3} & 66.3\%{\scriptsize$\pm$7.1} \\
 &  & \ECBM{}    & 34.5\%{\scriptsize$\pm$11.1} & 52.5\%{\scriptsize$\pm$11.1} & 53.5\%{\scriptsize$\pm$11.8} & 38.5\%{\scriptsize$\pm$16.4} & 34.5\%{\scriptsize$\pm$11.1} & 54.1\%{\scriptsize$\pm$11.2} & 56.6\%{\scriptsize$\pm$11.8} & 58.5\%{\scriptsize$\pm$12.2} \\
\midrule
\multirow{8}{*}{random} & \multirow{4}{*}{\textds{50\,px}} & \CBM{}     & 79.8\%{\scriptsize$\pm$5.9} & 80.7\%{\scriptsize$\pm$6.2} & 81.6\%{\scriptsize$\pm$6.8} & 71.9\%{\scriptsize$\pm$12.9} & 79.8\%{\scriptsize$\pm$5.9} & 81.5\%{\scriptsize$\pm$5.9} & 84.0\%{\scriptsize$\pm$6.0} & 88.5\%{\scriptsize$\pm$6.0} \\
 &  & \CEM{}     & 72.5\%{\scriptsize$\pm$9.5} & 73.4\%{\scriptsize$\pm$9.8} & 80.2\%{\scriptsize$\pm$9.3} & 85.8\%{\scriptsize$\pm$9.6} & 72.5\%{\scriptsize$\pm$9.5} & 74.1\%{\scriptsize$\pm$9.7} & 81.8\%{\scriptsize$\pm$9.3} & 89.7\%{\scriptsize$\pm$10.0} \\
 &  & \ProbCBM{} & 79.5\%{\scriptsize$\pm$2.6} & 79.2\%{\scriptsize$\pm$2.1} & 79.3\%{\scriptsize$\pm$2.2} & 76.2\%{\scriptsize$\pm$2.8} & 79.5\%{\scriptsize$\pm$2.6} & 80.1\%{\scriptsize$\pm$2.0} & 81.8\%{\scriptsize$\pm$1.9} & 91.8\%{\scriptsize$\pm$1.2} \\
 &  & \ECBM{}    & 53.1\%{\scriptsize$\pm$11.8} & 66.4\%{\scriptsize$\pm$12.3} & 66.8\%{\scriptsize$\pm$12.5} & 55.9\%{\scriptsize$\pm$14.2} & 53.1\%{\scriptsize$\pm$11.8} & 67.6\%{\scriptsize$\pm$12.4} & 68.8\%{\scriptsize$\pm$12.5} & 69.2\%{\scriptsize$\pm$12.5} \\
 & \multirow{4}{*}{\textds{10\,px}} & \CBM{}     & 58.8\%{\scriptsize$\pm$3.8} & 62.5\%{\scriptsize$\pm$4.0} & 65.9\%{\scriptsize$\pm$3.9} & 62.9\%{\scriptsize$\pm$3.2} & 58.8\%{\scriptsize$\pm$3.8} & 64.3\%{\scriptsize$\pm$3.8} & 70.7\%{\scriptsize$\pm$3.4} & 88.5\%{\scriptsize$\pm$1.0} \\
 &  & \CEM{}     & 36.4\%{\scriptsize$\pm$8.6} & 38.3\%{\scriptsize$\pm$9.4} & 42.0\%{\scriptsize$\pm$10.7} & 52.2\%{\scriptsize$\pm$11.7} & 36.4\%{\scriptsize$\pm$8.6} & 39.5\%{\scriptsize$\pm$9.5} & 45.4\%{\scriptsize$\pm$10.9} & 65.0\%{\scriptsize$\pm$12.6} \\
 &  & \ProbCBM{} & 39.6\%{\scriptsize$\pm$6.9} & 38.7\%{\scriptsize$\pm$6.8} & 38.0\%{\scriptsize$\pm$7.6} & 17.6\%{\scriptsize$\pm$15.2} & 39.6\%{\scriptsize$\pm$6.9} & 41.1\%{\scriptsize$\pm$6.6} & 45.1\%{\scriptsize$\pm$6.8} & 66.0\%{\scriptsize$\pm$7.2} \\
 &  & \ECBM{}    & 34.5\%{\scriptsize$\pm$11.1} & 52.4\%{\scriptsize$\pm$11.1} & 54.4\%{\scriptsize$\pm$12.0} & 41.1\%{\scriptsize$\pm$15.8} & 34.5\%{\scriptsize$\pm$11.1} & 54.0\%{\scriptsize$\pm$11.2} & 57.4\%{\scriptsize$\pm$12.0} & 58.5\%{\scriptsize$\pm$12.2} \\
\bottomrule
\end{tabular}

}
\caption{We report \captionwrk{} and \captioncov{} at $\alpha_{\textrm{target}}{=}0.99$ under three costs of an intervention. Under both unequal costs, every model automates less work with every concept verified than under the equal cost.}
\label{Table::CostModels}
\end{table}

\clearpage
\mlchecklist
\end{document}